\documentclass[screen,sigconf]{acmart}

 \setcopyright{none}
 \setcopyright{none}
 \renewcommand\footnotetextcopyrightpermission[1]{}

\usepackage{enumitem}
\usepackage{url}
\usepackage{xurl}
\usepackage{graphicx}  %
	\setkeys{Gin}{width=\textwidth, height=\textheight, keepaspectratio}
\usepackage[skip=2.5pt]{subcaption}
\newif\ifcomment
 \commenttrue
\ifcomment
	\newcommand{\edc}[1]{\textbf{\em\color{red}EDC: #1}}
	\newcommand{\gvg}[1]{\textbf{\em\color{blue}GG: #1}}
	\newcommand{\arxiv}[1]{#1}
	\newcommand{\ccsver}[1]{}
\else
	\newcommand\edc[1]{}
	\newcommand\gvg[1]{}
	\newcommand\arxiv[1]{}
	\newcommand{\ccsver}[1]{#1}
\fi
\newcommand{\descr}[1]{\smallskip\noindent\textbf{#1}}
\newcommand{\descrfirst}[1]{\noindent\textbf{#1}}

\definecolor{darkred}{RGB}{153,0,0}
\definecolor{darkblue}{RGB}{0,0,99}
\hypersetup{colorlinks=true, linkcolor=darkred, citecolor=darkred, urlcolor=darkblue}

\newcommand{\reduce}{\vspace{-0.05cm}}

\begin{document}
 \pagestyle{plain}

\sloppy
\title{Graphical vs. Deep Generative Models: Measuring the Impact of Differentially Private Mechanisms and Budgets on Utility$^*$}

\author{Georgi Ganev$^{1}$, Kai Xu$^{2}$, Emiliano De Cristofaro$^{3}$\\[0.5ex]
\large $^1$University College London and Hazy, $^2$MIT-IBM Watson AI Lab, $^3$University of California, Riverside\\[3ex]
}

\thanks{$^*$A shorter version of this paper appears in the Proceedings of the 31st ACM Conference on Computer and Communications Security (ACM CCS 2024). This is the full version.}

\renewcommand{\shortauthors}{Ganev et al.}
\renewcommand{\shorttitle}{Measuring the Impact of Differentially Private Mechanisms and Budgets on Utility}

\begin{abstract}
Generative models trained with Differential Privacy (DP) can produce synthetic data while reducing privacy risks.
However, navigating their privacy-utility tradeoffs makes finding the best models for specific settings/tasks challenging.
This paper bridges this gap by profiling how DP generative models for tabular data distribute privacy budgets across rows and columns, which is one of the primary sources of utility degradation.
We compare {\em graphical} and {\em deep generative} models, focusing on the key factors contributing to how privacy budgets are spent, i.e., underlying modeling techniques, DP mechanisms, and data dimensionality.

Through our measurement study, we shed light on the characteristics that make {\em different models suitable for various settings and tasks}.
For instance, we find that graphical models distribute privacy budgets horizontally and thus cannot handle relatively wide datasets for a fixed training time; also, the performance on the task they were optimized for monotonically increases with more data but could also overfit.
Deep generative models spend their budgets per iteration, so their behavior is less predictable with varying dataset dimensions, but are more flexible as they could perform better if trained on more features.
Moreover, low levels of privacy ($\epsilon\geq100$) could help some models generalize, achieving better results than without applying DP.
We believe our work will aid the deployment of DP synthetic data techniques by navigating through the best candidate models vis-à-vis the dataset features, desired privacy levels, and downstream tasks.
\end{abstract}

\maketitle

\section{Introduction}
\label{sec:introduction}
Generative machine learning models are increasingly used to create {\em synthetic data} that statistically resembles sensitive datasets without, at least in theory, exposing the real data.
The idea is that the synthetic data could then be freely shared with reduced privacy and regulatory concerns.
For instance, Microsoft recently worked with the International Organization for Migration to release a synthetic dataset to help counter human trafficking~\cite{microsoft2022iom}.
Beyond academic papers (see Sec.~\ref{sec:rw}), deployments and applications exist both in the public sector~\cite{benedetto2018creation, nist2018differential, nist2020differential, nhs2021ae, ico2022privacy, ons2023synthesising, hod2024differentially} and industry, where synthetic data vendors~\cite{crunchbase2022synthetic, techcrunch2022the, forbes2022synthetic} have offered solutions for test data generation, sensitive data sharing, augmentation, de-biasing, etc., often in highly regulated sectors like finance and health~\cite{ico2023synthetic, fca2024using, microsoft2024south,nhs2021ae}.

Alas, training generative models without privacy guarantees can lead to overfitting and memorization of training records~\cite{carlini2019secret, webster2019detecting}.
This, in turn, could enable attacks like membership and attribute inference~\cite{hayes2019logan, hilprecht2019monte, chen2020gan, stadler2022synthetic}, allowing an adversary to learn sensitive information about the real data, e.g., the inclusion of a particular individual in the real dataset or private features.
Thus, models should be trained while satisfying Differential Privacy (DP)~\cite{dwork2014algorithmic}, the established framework to provide rigorous privacy guarantees.
DP is typically achieved by applying calibrated noisy/random mechanisms during training so that any single data record's contribution, and thus its exposure, is provably bounded.

\descr{Motivation.}
The more complex DP algorithms become, the harder it is to predict their real-world performance and grasp for which applications/settings they could work well, which prompts the need for empirical evaluations~\cite{hay2016principled}.
While previous measurement papers have studied DP queries and classifiers~\cite{hay2016principled, jayaraman2019evaluating, liu2022ml, wei2023dpmlbench}, we focus on generative models for tabular data as predicting their complex behavior on downstream tasks with varying data dimensions and privacy budgets proves to be quite difficult.
The conventional wisdom is that DP methods become more accurate with more training data and less with stricter privacy guarantees.
However, in some cases, increasing the dataset size or the number of training iterations might make deep classifiers perform worse when optimized with DP-SGD~\cite{near2021programming}.
Moreover, a small degree of randomness introduced by DP algorithms for image data, even if not enough to provide meaningful privacy guarantees, can sometimes improve the performance of Convolutional Neural Networks (CNNs) on limited data~\cite{pearce22can} and Generative Adversarial Networks (GANs) for imbalanced data~\cite{ganev2022dp}.
This prompts the need to evaluate the performance of DP generative models case by case, as also discussed in~\cite{jordon2022synthetic}.

\descr{Technical Roadmap.}
In this paper, we measure the {\em effects} on the privacy-utility tradeoffs of {\em i)} different generative techniques, {\em ii)} various DP mechanisms, and {\em iii)} the dimensionality of the training data.
We profile DP generative models based on how they distribute their privacy budget across rows and columns while varying their number, as ``spending the budget'' through noisy/random mechanisms is where utility degradation mainly comes from.
This lets us evaluate how the choice of model and DP mechanism affects the quality of the synthetic data for downstream tasks, e.g., capturing distributions, maintaining high similarity, clustering, classification.

Our evaluation involves both graphical and deep generative state-of-the-art models; we experiment with PrivBayes~\cite{zhang2017privbayes} and MST~\cite{mckenna2021winning} for the former and DP-WGAN~\cite{alzantot2019differential} and PATE-GAN~\cite{jordon2018pate} for the latter.
In essence, graphical models use directed acyclic or undirected graphs to break down the joint distribution of the training data into lower-dimensional marginals, while the deep generative models rely on the GAN architecture, which consists of two competing neural networks.

The models we study have won various NIST competitions~\cite{nist2018differential, nist2020differential}, have been used by the UK and Israel governments to release census data~\cite{ons2023synthesising, hod2024differentially}, and are part of product offerings by leading companies in the space~\cite{tonic2021what, gretel2023models, hazy2023models}.
However, previous benchmark studies~\cite{tao2022benchmarking, mckenna2022aim, liu2022utility} evaluate these models on small datasets (i.e., involving at most a dozen features) and do not go beyond a single downstream task or metric.
These studies argue that graphical models are superior to deep generative models, with MST~\cite{mckenna2021winning} highlighted as the best overall~\cite{mckenna2022asimple}.
To verify whether this holds in a scalable way, we train around 21,000 models on realistic datasets with dimensionality at least ten times larger than related work~\cite{tao2022benchmarking, mckenna2022aim} and test their performance on a variety of downstream tasks.

\descr{Main Findings.}
We study the effects of model and DP instantiations, as well as dataset dimensions, by measuring how {\em scalable} DP generative models are in terms of dataset dimensions and whether DP generative models distribute their privacy budgets in a similar way.
We show that the graphical models distribute their privacy budget per column and cannot scale to many features within practical time constraints (256 for PrivBayes and 128 for MST at most) as they suffer from the ``curse of dimensionality''~\cite{bellman1957dynamic}.
Also, increasing the number of rows barely affects the training time.
Whereas, deep generative models spend their budget per training iteration and can handle much wider datasets but become slower with more data.

Overall, our measurements yield a few interesting findings:
\begin{enumerate}[leftmargin=*]
	\item Graphical generative models are better suited for datasets with limited features and simple tasks, while deep generative models for larger datasets and more complex problems. \smallskip
	\item %
	PrivBayes~\cite{zhang2017privbayes} is the only model with fairly consistent behavior; its performance monotonically degrades with stricter privacy budgets or more columns, while more data counters these effects.
	Also, it is the only model that can successfully separate signal from noise in the clustering task. \smallskip
	\item MST~\cite{mckenna2021winning} has the best privacy-utility tradeoff in capturing simple statistics.
	With limited data, it benefits from DP noise with high bounds ($\epsilon\geq100$).
	However, excessively increasing the number of rows can cause it to overfit and degrade its performance on more complex tasks.
	Moreover, it does not scale well and often underperforms compared to other models, thus contradicting previous research %
	\cite{nist2018differential, tao2022benchmarking, mckenna2022asimple}. \smallskip
	\item Deep generative models exhibit more flexible and variable behaviors with different dataset dimensions.
	While they are not as competitive on simple tasks, PATE-GAN~\cite{jordon2018pate} is, in fact, better suited to more complex tasks and often outperforms both graphical models, which refutes arguments that GANs are unsuitable for tabular data~\cite{tao2022benchmarking, mckenna2022aim, liu2022utility}.\smallskip
	\item Perhaps unexpectedly, for all models, adding more data or relaxing the privacy constraints (increasing $\epsilon$) can, in some cases, hurt performance, %
	thus showcasing the difficulty of deploying stable and consistent DP synthetic data models.
\end{enumerate}

\arxiv{\smallskip
We conclude by discussing potential first-cut improvements as well as future research directions.}
Overall, our work can assist researchers and practitioners deploying DP synthetic data techniques in understanding the tradeoffs and navigating through the best candidate models vis-\`a-vis the dataset features, desired privacy levels, and downstream tasks.

\section{Background}
\label{sec:back}

\descr{Differential Privacy (DP).}
A randomized algorithm $\mathcal{A}$ satisfies ($\epsilon, \delta$)-DP if, for all sets of its possible outputs $S$, and all neighboring datasets $D$ and $D^{\prime}$ ($D$ and $D^{\prime}$ are identical except for a single data row), the following holds~\cite{dwork2006calibrating, dwork2014algorithmic}:\reduce
\begin{equation}
\label{eq:dp}
P[{\mathcal{A}}(D)\in S]\leq \exp \left(\epsilon \right)\cdot P[{\mathcal{A}}(D^{\prime})\in S] + \delta\reduce
\end{equation}
where $\epsilon$ is a positive real number (also referred to as the {\em privacy budget}) that bounds the information leakage, while $\delta$, usually a very small real number, allows for a probability of failure.
Put simply, looking at the output of $\mathcal{A}$ (e.g., a trained model), it is impossible to distinguish whether any individual's data was included in the input dataset (e.g., the training data).

\descr{DP Generative Models}.
In this paper, we study techniques to create synthetic data through generative machine learning models.
A sample dataset $D^n$, consisting of $n$ iid drawn records with $d$ features from the population $D^n{\sim}P(\mathbb{D})$, is fed as input to the generative model training algorithm $GM(D^n)$ during the fitting step.
In turn, $GM(D^n)$ updates its internal parameters $\theta$ to learn $P_{\overline{\theta}}(D^n)$, a lower-dimensional representation of the probability of the sample dataset $P(D^n)$, and outputs the trained model $\overline{GM}(D^n)$.
Then, $\overline{GM}(D^n)$ is sampled to generate a synthetic data of size $n^{\prime}$, $S^{n^{\prime}}{\sim}P_{\overline{\theta}}(D^n)$.
Since the fitting and generation steps are stochastic, one can train the generative model $m$ times and sample $s$ synthetic datasets for each trained model to get confidence intervals.

We focus on generative models for {\em tabular} synthetic data generation trained while satisfying DP.
Prior work has proposed a number of algorithsm, including copulas~\cite{li2014differentially, asghar2021differentially, gambs2021growing}, graphical models~\cite{zhang2017privbayes, zhang2019differentially, mckenna2021winning, cai2021data, mahiou2022dpart}, workload/query-based~\cite{vietri2020new, aydore2021differentially, liu2021iterative, vietri2022private, mckenna2022aim}, Variational Autoencoders (VAEs)~\cite{acs2018differentially, abay2018privacy, takagi2021p3gm}, Generative Adversarial Networks (GANs)~\cite{xie2018differentially, zhang2018differentially, jordon2018pate, alzantot2019differential, frigerio2019differentially, tantipongpipat2021differentially, long2021gpate}, etc.~\cite{chanyaswad2019ron, zhang2021privsyn, ge2021kamino}.

Finally, we leverage two DP properties: composition~\cite{kairouz2015composition} and post-processing~\cite{dwork2014algorithmic}.
The former implies that different DP mechanisms can be combined while easily tracking the overall privacy budget.
The latter guarantees that, once a DP model is trained, it can be re-used arbitrarily many times, including to generate fresh synthetic datasets, without further privacy leakage.

\section{Experimental Setup}
\label{sec:preliminaries}
\arxiv{In this section, we present the DP generative models, datasets, and downstream tasks used in our evaluation.}

\begin{table}[t!]
	\small
	\centering
	\begin{tabular}{lccr}
		\toprule
		\bf DP Model & \bf Model Type & \bf DP Mechanism & \bf Max $d$ \\
		\midrule
		Independent & Marginal & Laplace & 1,024 \\
		PrivBayes & Graphical & Exponential + Laplace & 256 \\
		MST & Graphical &	Exponential + Gaussian & 128 \\
		DP-WGAN & Deep Learning	& DP-SGD & 1,024 \\
		PATE-GAN & Deep Learning & PATE & 1,024 \\
		\bottomrule
	\end{tabular}
	\caption{DP generative models studied in this paper ($d$ denotes the number of data features a model can handle within practical time constraints; more details in Table~\ref{tab:scal_train_cols}).}
	\label{tab:dp_models}
\reduce
\end{table}

\subsection{DP Generative Models}
\label{subsec:dp_models}
Table~\ref{tab:dp_models} lists the DP generative models used in our empirical evaluation.
In addition to a baseline marginal model that treats each feature as separate and uncorrelated to the others (``Independent''), we focus on two types of approaches: graphical models (PrivBayes and MST) and deep generative models (DP-WGAN and PATE-GAN).
The former model the joint distribution by breaking it down to explicit lower-dimensional marginals.
The latter approximate the distribution implicitly by iteratively optimizing two competing neural networks: %
a generator creates synthetic data from noise, and a discriminator separates real from synthetic points.
(Since we study two GAN models from the latter category, we use the terms GANs and deep generative models interchangeably.)

\descr{Why These Models.}
Arguably, these models are the most popular, well-studied, and highly cited DP generative models for tabular data.
They also have reliable and thoroughly tested open-source implementations and have been widely deployed in the wild.
For instance, MST and DP-WGAN won, respectively, the NIST DP Synthetic Data~\cite{nist2018differential} and Unlinkable Data~\cite{nist2018the} open challenges, while MST and PrivBayes have been used by the UK Office Of National Statistics~\cite{ons2023synthesising} and the Israel Ministry of Health~\cite{hod2024differentially} to release synthetic data to the public.
Also, these models are widely used by the leading synthetic data providers~\cite{tonic2021what, gretel2023models, hazy2023models}.
Finally, the four models rely on different modeling techniques and DP mechanisms, thus covering a wide problem space and broader downstream tasks, thus allowing us to draw comparisons over different dimensions.

The specific implementations we use are listed in footnotes.
For all experiments, we use the default hyperparameters used by the models' authors unless stated otherwise.

\descr{Independent.}\footnote{\label{hazy}\url{https://github.com/hazy/dpart}}
As a baseline, we use a simple model that, for all columns, independently captures noisy marginal counts through the Laplace mechanism and then samples from them to generate synthetic data.
It ignores all pairwise and higher-order correlations in the data.
Although very simple, it has proven to perform better than more sophisticated models in certain scenarios~\cite{tao2022benchmarking}.

\descr{PrivBayes~\cite{zhang2017privbayes}}\textsuperscript{\ref{hazy}}
first constructs a directed acyclic graph, referred to as a Bayesian network, to break down the high-dimensional joint distribution into low-dimensional conditional ones.
In the network, every node represents a random variable corresponding to a column in the dataset.
To build the network, the model starts by randomly picking up a node to serve as the root.
It then iteratively adds one node at a time until every node is linked -- at each step, it noisily adds directed edges from (a subset of) already connected ``parent'' nodes to a non-connected ``child'' node with the highest mutual information (the mutual information scores of all combinations between (subsets of) high-dimensional ``parent'' distributions and one-dimensional ``child'' distributions are calculated).
Half of $\epsilon$ is spent at this step using the Exponential mechanism, while the network degree argument determines the maximum number of parents a given node can have (an example of a fitted PrivBayes network is given in Figure~\ref{fig:census_network_privbayes}).
As a second step, PrivBayes uses the resulting low-dimensional marginals to compute noisy contingency tables (utilizing the Laplace mechanism) before normalizing and converting them to conditional distributions.
Since all conditional distributions are calculated iteratively based on the captured directed network, they are also consistent, i.e., they are valid distributions (sum up to 1), and there are no contradictions with the joint distributions.

In PrivBayes, two hyperparameters might affect how the privacy is distributed, namely the network degree (by default, set to 3) and the number of bins (numerical columns need to be discretized; by default, set to 20).
As previous work~\cite{tao2022benchmarking, mckenna2022aim} confirms that both PrivBayes and MST are expected to work well out of the box with the default hyperparameters, we also use those.

We run PrivBayes for datasets with columns up to 256, setting the network degree to 3 for datasets with fewer than 100 columns and to 2 otherwise to improve performance.

\descr{MST~\cite{mckenna2021winning}\footnote{\url{https://github.com/ryan112358/private-pgm}}}
follows a similar procedure.
For selection, it starts with all 1-way marginals and finds attribute pairs (2-way marginals) that form an undirected graph.
More precisely, the $L_1$ distance between real and estimated 2-way marginals serve as edge weights, and a maximum spanning tree, optimizing for maximum weight, is iteratively constructed.
This is achieved by first estimating all 2-way marginals using Private-PGM~\cite{mckenna2019graphical} (a post-processing method that infers a data distribution given noisy measurements) and, then, one by one, noisily adding a highly weighted edge to the graph until all nodes are connected (an example of a fitted MST network is shown in Figure~\ref{fig:census_network_mst}).
All selected marginals are measured privately using the Gaussian mechanism.
MST allocates $1/3$ of the privacy budget to selection and the remaining $2/3$ to measurement.

As the model only takes discrete data as input, the number of bins (default is 20) could be a factor in how the privacy budget is spent.
We train MST on datasets with up to 128 columns and set $\delta=10^{-5}$ (to be consistent with DP-WGAN and PATE-GAN).

\descr{DP-WGAN~\cite{alzantot2019differential}\footnote{\url{https://github.com/nesl/nist_differential_privacy_synthetic_data_challenge}}}
utilizes the Wasserstein GAN (WGAN) architecture~\cite{arjovsky2017wasserstein}, which achieves better learning stability and improves mode collapse issues compared to the original GAN by using Wasserstein distance rather than Jensen-Shannon divergence.
The model relies on DP-SGD to ensure the privacy of the discriminator during training, which in turn guarantees the privacy of the generator since it is never exposed to real data.

In DP-WGAN, the hyperparameters that could influence the privacy distribution are the learning rate (default: 0.001), batch size (default: 64), number of iterations/epochs (default number of epochs is 20), clipping bound (default: 1), and noise multiplier (default: 1); we use the default hyperparameters from the winning solution of the NIST Unlinkable Data challenge~\cite{nist2018the}.

\descr{PATE-GAN~\cite{jordon2018pate}\footnote{\url{https://bitbucket.org/mvdschaar/mlforhealthlabpub}}}
adapts PATE and combines it with a GAN architecture (which removes the need for public data).
The architecture consists of a single generator, $t$ teacher-discriminators (by default, set to 10), and a student-discriminator.
The teacher-discriminators are only presented with disjoint subsets of the training data and are trained to improve their loss with respect to the generator (i.e., classifying samples as real or fake).
In contrast, the student-discriminator is trained on noisily aggregated labels provided by the teachers, and its loss gradients are sent to train the generator.

Besides the number of teacher-discriminators, the learning rate (default: 0.0001), batch size (default: 128), and number of iterations/epochs (default number of epochs is 20) are additional parameters that could also have an effect on the privacy budget.

\descr{\em Note on Hyperparameters:} As mentioned, we use the default hyperparameters, emulating real-world situations where optimization procedures might not always be accessible and following previous related work~\cite{tao2022benchmarking, mckenna2022aim, stadler2022synthetic, ganev2022robin}.
We note that optimizing hyperparameters for every setting would consume additional privacy budget and add yet another dimension to our measurement study, and thus, we consider it to be out of the scope of this work.

\begin{table*}[t!]
	\centering
	\small
	\setlength{\tabcolsep}{2.5pt}
	\begin{tabular}{lrrrp{13.4cm}}
		\toprule
		\bf Dataset & \bf Max $n$& \bf Max $d$ \bf && \bf Downstream Tasks (Relevant Plots)\\
		\midrule
		\emph{Eye~Gauss} & 128k & 1,024 && T1: Statistics -- Mean\arxiv{ (Fig.~\ref{fig:eye_mean})} and Correlation\arxiv{ (Fig.~\ref{fig:eye_other_corr})}, T3: Clustering -- PCA\arxiv{ (Fig.~\ref{fig:eye_pca_rows},~\ref{fig:eye_pca_cols})}\\[0.5ex]
		\emph{Corr~Gauss} & 128k & 1,024 && M1: Scalability -- Fitting (Tab.~\ref{tab:scal_train_rows},~\ref{tab:scal_train_cols}) and Generation \arxiv{(Tab.~\ref{tab:scal_gen_rows},~\ref{tab:scal_gen_cols}) }runtime,\\
		&&&&T1: Statistics -- Mean\arxiv{ (Fig.~\ref{fig:corr_mean})} and Correlation (Fig.~\ref{fig:corr_off_diag_corr}\arxiv{,~\ref{fig:corr_other_corr}}), T3: Clustering -- PCA \arxiv{(Fig.~\ref{fig:corr_pca_rows},~\ref{fig:corr_pca_cols})}\\[0.5ex]
		\emph{Mix Gauss Unsup}& 128k & 1,024 && T3: Clustering -- PCA (Fig.~\ref{fig:mix_pca_rows},~\ref{fig:mix_pca_cols}) and Silhouette \arxiv{(Fig.~\ref{fig:mix_sil})}\\[0.5ex]
		\emph{Mix Gauss Sup} & 128k & 1,024 && T3: Clustering -- PCA\arxiv{ (Fig.~\ref{fig:mix_target_pca_rows},~\ref{fig:mix_target_pca_cols})}, T4: Classification -- Accuracy \arxiv{(Fig.~\ref{fig:mix_target_acc})}\\[0.5ex]
		\emph{Plants} & 28k & 70 && T3: Clustering -- UMAP\arxiv{ (Fig.~\ref{fig:plants_umap})} and Silhouette \arxiv{(Fig.~\ref{fig:plants_sil})}\\[0.5ex]
		\emph{Diabetes} & 57k & 37 && T2: Similarity -- Marginal and Mutual information \arxiv{(Fig.~\ref{fig:diabetes_sim_mi})}\\[0.5ex]
		\emph{Covertype} & 465k & 55 && T2: Similarity -- Marginal and Mutual information \arxiv{(Fig.~\ref{fig:covertype_sim_mi})}\\[0.5ex]
		\emph{Census} & 199k & 41 && T2: Similarity -- Marginal and Mutual information (Fig.~\ref{fig:census_sim_mi},~\ref{fig:census_sim_mi_mst_pategan},~\ref{fig:census_network_privbayes_mst},~\ref{fig:census_network_mi_privbayes_mst}),\ T4: Classification -- Accuracy and F1-score (Fig.~\ref{fig:census_acc_f1},~\ref{fig:census_pretrained_acc_f1})\\[0.5ex]
		\emph{Connect~4} & 54k & 43 && T4: Classification -- Accuracy and F1-score \arxiv{(Fig.~\ref{fig:connect_4_acc_f1})}\\[0.5ex]
		\emph{MNIST} & 60k & 784 && T4: Classification -- Accuracy (Fig.~\ref{fig:mnist_acc},~\ref{fig:upscaled_mnist_acc})\\
		\bottomrule
	\end{tabular}
	\caption{Datasets and downstream tasks used in our empirical experiments; $n$ and $d$ denote the number of data records and the number of features, respectively.\ccsver{ Additional tables and figures are presented in the full version of the paper~\cite{ganev2023understanding}.}}
	\label{tab:datasets}
\reduce
\end{table*}

\reduce\subsection{Datasets}
\label{subsec:datasets}

As mentioned, we focus on tabular data as it is among the most popular data modalities in real-world applications~\cite{benedetto2018creation, nist2018differential, nist2018the, nist2020differential, nhs2021ae, ico2022privacy, ons2023synthesising, hod2024differentially, tonic2021what, gretel2023models, hazy2023models}.
Also, all generative models we study were originally proposed and tested for tabular data.

Our experiments use two common, relatively simple datasets (i.e., \emph{Census} and \emph{MNIST}) to be consistent with prior DP evaluations~\cite{ganev2022dp, pearce22can, tao2022benchmarking}, as well as four high-dimensional, mixed-type datasets (i.e., \emph{Plants}, \emph{Diabetes}, \emph{Covertype} and \emph{Connect~4}) to ensure the generalizability of our results.
We also create four controlled datasets based on the normal distribution (\emph{Eye~Gauss}, \emph{Corr~Gauss}, \emph{Mix~Gauss~Unsup}, \emph{Mix~Gauss~Sup}), with progressively more difficult tasks to test the basic properties of the generative models and their ability to replicate datasets appropriately.
Compared to previous benchmarking studies~\cite{tao2022benchmarking, mckenna2022aim, liu2022utility}, these datasets are more realistic and have higher dimensionality, covering a comprehensive and diverse set of domains and downstream tasks.
A summary of our ten datasets is provided in Table~\ref{tab:datasets}; details follow below.

\descr{Gaussians.}
We create four progressively more complex datasets based on the Gaussian distribution:
\begin{itemize}[leftmargin=0.5cm]
	\item \emph{Eye~Gauss} consists of columns that are independently distributed standard normals.\vspace{0.05cm}
	\item \emph{Corr~Gauss}~(inspired by~\cite{belghazi2018mutual, poole2019variational, rhodes2020telescoping}) is a multivariate normal with mean 0 and covariance matrix with 1s on the diagonal (unit variance), 0.5s on the off-diagonal (all neighboring columns have correlation 0.5), and 0s everywhere else (not neighboring columns are uncorrelated).
	The idea is to see if the model can capture the correlation correctly.\vspace{0.05cm}
	\item The first two columns of \emph{Mix~Gauss~Unsup}~(inspired by~\cite{frigerio2019differentially}) are a mixture of six Gaussians distributed in a ring with center 0, while the remaining columns represent noise in the form of uncorrelated gaussians with mean 0.
	The model should be able to separate signal from noise and reproduce the six clusters.\vspace{0.05cm}
	\item The \emph{Mix~Gauss~Sup} dataset is the same as the previous one but with an added target column, labeling the six Gaussians in a non-linearly separable way.
	Classifiers trained on the real and on the synthetic data should have similar performance.
\end{itemize}

The number of columns for all datasets varies in \{8, 16, 32, 64, 128, 256, 512, 1,024\} while we keep the rows to 16k.
We also vary the number of rows in the range \{250, 500, 1k, 2k, 4k, 8k, 16k, 32k, 64k, 128k\} while fixing the columns to 32.
When applicable, we create test datasets with sizes equal to 20\% of the training.

\descr{Plants.}
The Plants dataset~\cite{dua2017uci} includes all plants (both species and genera) from the USDA database, along with the states in the USA and Canada where they are found.
It features 70 binary variables, each indicating the presence of the plants in a specific state.
The dataset is intended for a clustering task.
Although there are a total of 34,781 plants, we set aside 20\% (6,957) for testing and vary the training data in the range of \{1k, 2k, 4k, 8k, 16k, 27,824\}.

\descr{Diabetes.}
The Diabetes dataset~\cite{dua2017uci} spans a decade (1999-2008) of clinical data from 130 US healthcare facilities.
It centers on the hospital records of diabetic patients, encompassing their laboratory tests, medications, and stays lasting up to 14 days.
From the original 47 features, we retain 37 (5 numerical, 23 categorical), discarding those that are highly imbalanced (>99\%).
We set aside 14,304 (20\%) records for testing and adjust the training data within the range of \{1k, 2k, 4k, 8k, 16k, 32k, 57,214\}.

\descr{Covertype.}
This dataset~\cite{dua2017uci} contains cartographic variables such as wilderness areas and soil types.
There are 55 variables (10 numerical and 45 categorical).
While there are 581,012 data points, we separate 20\% (116,203) for testing purposes and vary the training points in the range \{1k, 2k, 4k, 8k, 16k, 32k, 64k, 128k, 256k, 464,809\}.

\descr{Census.}
The Census dataset~\cite{dua2017uci} is extracted from the 1994 and 1995 Current Population Surveys conducted by the US Census Bureau.
It contains 41 (six numerical and 35 categorical) demographic and employment-related variables.
The target column indicates whether the individual's income exceeds \$50k/year.
The dataset consists of 199,523 training and 99,762 testing instances.
We vary the training points in the range \{1k, 2k, 4k, 8k, 16k, 32k, 64k, 128k, 199,523\} (the latter being the original size).

\descr{Connect~4.}
The Connect~4 dataset~\cite{dua2017uci} contains all legal positions in the game of connect-4 (vertically, horizontally, or diagonally) in which neither player has won yet and the next move is not forced.
The outcome class is the game-theoretical value for the first player.
The dataset consists of 67,557 data instances and 43 categorical features, including the target class.
We set aside 20\% (13,512) of the data for testing and vary the training size in the range \{1k, 2k, 4k, 8k, 16k, 32k, 54,045\}.

\descr{MNIST.}
The MNIST dataset~\cite{lecun2010mnist} is a collection of greyscale handwritten digits.
There are 60k training and 10k testing samples.
The task is to classify the digit.
We experiment with a varying number of features.
On top of the original 28x28 pixels, we also rescale the images to 10x10, 16x16, and 22x22, which allows us to test different data dimensionalities without really compromising quality.

\subsection{Measurements and Downstream Tasks}
\label{subsec:tasks}
Table~\ref{tab:datasets} lists the associated measurements and downstream tasks on which we evaluate the models: scalability (M1), statistics (T1), similarity (T2), clustering (T3), and classification (T4).
For clarity, we also use the bracketed codes to denote these tasks.

We aim to test the DP generative models on diverse and realistic downstream tasks of increasing complexity, and we measure success through specific metrics.
For the scalability measurement (M1), we report metrics based solely on the generative models, focusing only on the shape of the training data.
For the statistics task (T1), we directly report the statistics calculated on the synthetic data, as we know what the target values are.
For similarity (T2), we directly compare the synthetic dataset to the real one.
For clustering (T3) and classification (T4), we adopt the evaluation approach outlined by~\cite{esteban2017real, jordon2018pate}.
Specifically, we fit two predictive models: one on the real training data and another on the synthetic data.
We then compare their performance on a set-aside real test dataset, which comes from the same distribution as the training but remains unseen by both the predictive and generative models.

We generate datasets with the same dimensions as the training data for all measurements and tasks.
To capture variability in our results, each evaluation metric is run once for every synthetic data sample, totaling 25 runs (comprising 5 training runs times 5 generation runs per training run).
We report the average results along with confidence intervals.

\descr{M1: Scalability.}
We use a simple and standard performance metric by measuring the runtime in minutes of the two main steps of the generative models -- fitting and generation.

\descr{T1: Statistics.}
For this task, we test whether the DP generative models can capture and reproduce simple distributions based on the standard normal distribution.
We report their success at modeling the two main parameters, mean and covariance matrix.
For the mean, we report the average across all columns; for the pairwise correlations, depending on the dataset, we report two or three types of averages: across the diagonal, across the off-diagonal, and across all non-diagonal elements.

\descr{T2: Similarity.}
To measure the similarity between two tabular datasets, in line with~\cite{patki2016the, ping2017datasynthesizer, tao2022benchmarking, qian2023synthcity}, we use marginal similarity and pairwise mutual information similarity.
To accommodate the mixed-type nature of tabular data, both metrics first discretize the datasets (this is also a necessary pre-processing step of Independent, PrivBayes, and MST; we again use 20 bins).
The former measures the average similarity of the distributions between corresponding sets of columns using the Jaccard score.
The latter calculates the average element-wise score (again Jaccard) between the two $d*d$ mutual information matrices (in which the $(i,j)$-th entry corresponds to the mutual information score between the $i$-th and $j$-th columns), excluding the diagonal since its value is always one.

\descr{T3: Clustering.}
To visually assess whether the models capture the overall distributions of the real high-dimensional datasets, we employ standard PCA and UMAP dimensionality reduction techniques and then run KDE on the first two components.
For quantitative measurements of how well the DP generative models separate different data clusters, we fit Mixture of Gaussians models on the reduced datasets, test them on set-aside test data, and compare their silhouette scores.

\descr{T4: Classification.}
We chose logistic regression as our predictive model to minimize additional sources of randomness.
While this approach might not yield the highest possible accuracy, our primary focus is on comparing the performance between real and synthetic data.
To this end, we train logistic regression models on both datasets.
We then evaluate and contrast their accuracy and, when dealing with imbalanced training data, their F1-scores using an unseen test dataset.

\section{Privacy Profiling}
In this section, we focus on the steps during the training phase in which the DP generative models distribute their privacy budgets.
The two types of approaches (i.e., graphical and deep generative models) substantially differ from a DP perspective, e.g., what DP mechanisms they use, how they distribute their budgets, what factors cause more considerable expenditures, etc.

More precisely, the graphical models rely on the ``select-measure-generate'' paradigm, i.e., they start with 1) selecting a collection of low-dimensional marginals and 2) measuring them with a noise addition mechanism.
Naturally, as $d$ increases, the privacy budget must be distributed among more marginals, thus more noisy measurements.
However, increasing $n$ could decrease the per-measurement sensitivity, yielding more accurate estimations.

As for GANs, one of the most widely used approaches is to train them iteratively using mini-batches based on DP-SGD.
For fixed-networks architectures, increasing $d$ should not be a major factor as that only affects the discriminator's input and the generator's output layers.
Analyzing the effect of increasing $n$ is more complicated.
On the one hand, more (clean and diverse) training data is better for the model as that helps it generalize.
Also, with more data, one can theoretically achieve better utility for the same privacy or more privacy for the same utility through privacy amplification by sampling~\cite{balle2018privacy, mironov2019renyi}.
On the other hand, in practice, DP training typically requires larger datasets (or more iterations) to converge and achieve good utility.
But more iterations could also make the model worse as a lower privacy budget is used per training step, which increases the scale of the noise~\cite{near2021programming}.

\descr{Independent.}
Unlike the other four models, Independent is the only one that distributes its DP budget {\em independently} of the data.
It always selects the same marginals (all columns) and uses the same amount of budget to get every noisy marginal, $\epsilon/d$.

\descr{PrivBayes \& MST.}
In total, PrivBayes measures approximately $d$ 4-dimensional (if the network degree is three) noisy distributions, each of which has allocated $\epsilon/2d$.
The MST measurement step has a few advantages over PrivBayes: i) it devotes more budget to it (2/3 vs.~1/2); ii) it models structural zeros in the distribution (areas with negligible probability) to which it does not add noise; iii) it uses the Gaussian mechanism, which has better bounds than Laplace under some conditions (i.e., number of measurements is greater than $\log(1/\delta) + \epsilon$~\cite{mckenna2022aim}); and iv) even though in total it measures more marginals (approx.~$2d$ 1 or 2-way vs. $d$ 4-way), their dimensionality is lower, which means that the noise could be distributed more efficiently.
However, it requires more computations, potentially leading to slower runtimes as Private-PGM is called twice, in the selection and generation steps.

\descr{DP-WGAN \& PATE-GAN.}
In DP-WGAN, the privacy budget is not directly computed but estimated using the moments accountant method~\cite{abadi2016deep}.
Unlike graphical models, DP-WGAN spends its privacy budget iteratively through DP-SGD, which, unfortunately, tends to overestimate the sensitivity of many data points~\cite{thudi2023gradients}.
PATE-GAN also relies on the moments accountant.
An advantage over DP-WGAN is that noise is not added directly to the gradients but to the vote of the teacher-discriminators~\cite{papernot2017semi, papernot2018scalable}.
Furthermore, the accountant in PATE-GAN would attribute a lower privacy cost to accessing noisy aggregations (from the teacher-discriminators) with stronger consensus as a single teacher/data point would have a lower influence on the final vote.

\begin{table}[t!]
	\centering
 	\small
	\setlength{\tabcolsep}{4pt}
	\begin{tabular}{l@{}rrrrrrrr}
		\toprule
		\bf DP Model$\downarrow$ \bf $n\hspace{-0.05cm}\rightarrow$ & \bf 250 & \bf 500 & \bf 1k & \bf 4k & \bf 16k & \bf 32k & \bf 64k & \bf 128k \\
		\midrule
		Independent & 0.01 & 0.01 & 0.01 & 0.01 & 0.01 & 0.02 & 0.03 & 0.05 \\
		PrivBayes & 0.02 & 0.03 & 0.03 & 0.05 & 0.06 & 0.07 & 0.10 & 0.13 \\
		MST & 3.23 & 3.27 & 3.27 & 3.28 & 3.23 & 3.23 & 3.32 & 3.30 \\
		DP-WGAN & 0.11 & 0.11 & 0.13 & 0.24 & 0.76 & 1.42 & 2.73 & 5.37 \\
		PATE-GAN & 0.02 & 0.03 & 0.05 & 0.18 & 0.76 & 1.75 & 4.55 & 13.40 \\
		\bottomrule
	\end{tabular}
	\caption{M1: Runtime (in mins) of the model's fitting step for DP generative models, on \emph{Corr~Gauss}, varying $n$ and $d=32$.}
	\label{tab:scal_train_rows}
\reduce
\end{table}

\section{Experimental Evaluation}
\label{sec:experiments}
In this section, we present a comprehensive experimental evaluation involving the four DP generative models (along with the baseline model, Independent) and the ten datasets introduced above on several measurements/downstream tasks -- scalability (M1), statistics (T1), similarity (T2), clustering (T3), and classification (T4).

For all generative models and all privacy budgets, we train the model $m=5$ times and generate $s=5$ synthetic datasets; this yields 25 synthetic datasets for each reported point in the plots (we report the average score and confidence intervals).
Besides varying the dataset dimensions\arxiv{ (as discussed in Section~\ref{subsec:datasets})}, we also vary the privacy budget $\epsilon$ in \arxiv{the range} \{0.01, 0.1, 1, 10, 100, 1k, 10k, $\infty$\} to be consistent with previous studies~\cite{pearce22can, balle2022reconstructing, guo2023analyzing, hayes2023bounding}.
While not claiming that high $\epsilon$ values ($\geq100$) are universally save to use, these papers argue they might be enough to protect vs. specific threat models such as multiple hypothesis testing and reconstruction attacks with strong adversaries.
Therefore, we include these $\epsilon$ values in our evaluation as well.

In total, we train 21k generative models and generate 105k synthetic datasets.
All experiments are run on an AWS instance (m4.16xlarge) with a 2.4GHz Intel Xeon E5-2676 v3 (Haswell) processor, 64 vCPUs, and 256GB RAM.
As mentioned, a summary of all our experiments is reported in Table~\ref{tab:datasets}.

\subsection{M1: Scalability}
\label{subsec:scalability}

\descr{Setup.}
We run all models on \emph{Corr~Gauss} while varying $d$ and $n$, and report runtime (in mins) of the fitting and generation steps averaged across all $\epsilon$ values.
In Tables~\ref{tab:scal_train_rows} and~\ref{tab:scal_train_cols}, we report the fitting step runtime while varying $n$ and $d$, respectively.
In Appendix~\arxiv{\ref{app:scalability}}\ccsver{A.1 of the paper's full version~\cite{ganev2023understanding}}, we also report the generation runtimes (for a fitted model) across different dimensions\arxiv{; see Tables~\ref{tab:scal_gen_rows} and \ref{tab:scal_gen_cols} in Appendix~\ref{app:scalability}}.

\begin{table}[t!]
	\centering
	\small
	\setlength{\tabcolsep}{4pt}
	\begin{tabular}{l@{}rrrrrrrr}
		\toprule
		\bf DP Model$\downarrow$ \bf $d\hspace{-0.05cm}\rightarrow$ & \bf 8 & \bf 16 & \bf 32 & \bf 64 & \bf 128 & \bf 256 & \bf 512 & \bf 1,024 \\
		\midrule
		Independent & 0.00 & 0.01 &	0.01 & 0.03 & 0.08 & 0.22 & 0.89 & 3.03 \\
		PrivBayes & 0.01 & 0.02 & 0.06 & 0.26 & 2.29 & 34.15 & & \\
		MST & 0.75 & 1.55 & 3.23 & 7.71 & 35.90 & & & \\
		DP-WGAN & 0.50 & 0.58 & 0.76 & 1.24 & 3.53 & 6.47 & 12.52 & 23.97 \\
		PATE-GAN & 0.65 & 0.68 & 0.76 & 0.94 & 1.40 & 2.13 & 4.09 & 10.38 \\
		\bottomrule
	\end{tabular}
	\caption{M1: Runtime (in mins) of the model's fitting step for DP generative models, on \emph{Corr~Gauss}, varying $d$ and $n=16k$.}
	\label{tab:scal_train_cols}
\reduce
\end{table}

\begin{figure*}[t!]

	\centering
	\begin{subfigure}{0.53\linewidth}
		 \includegraphics[width=0.99\textwidth]{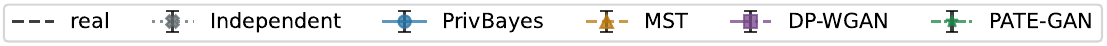}
	\end{subfigure}
	\begin{subfigure}{0.49\linewidth}
		 \includegraphics[width=0.99\textwidth]{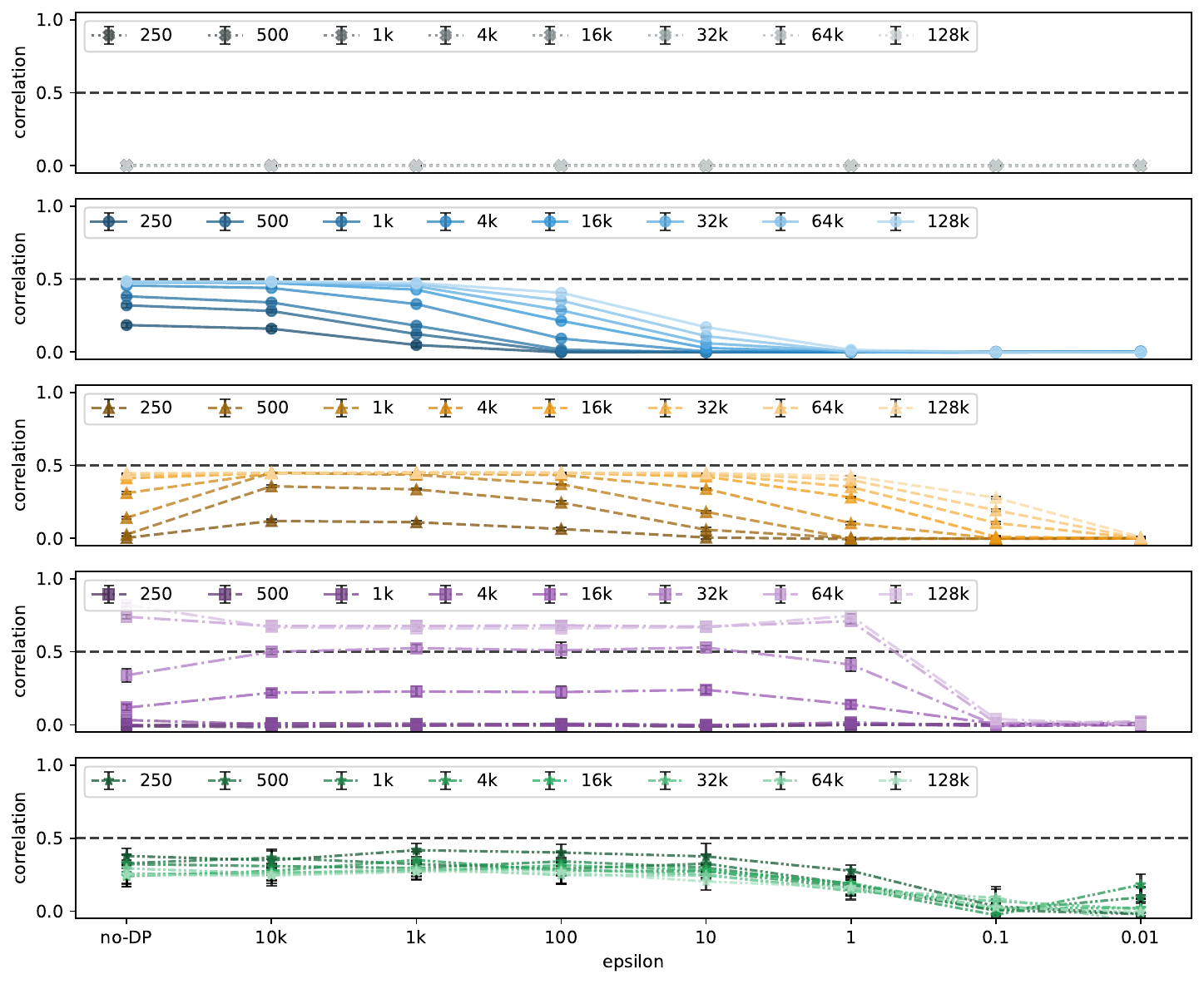}
		\caption{Varying $n$ and $d=32$}
	\end{subfigure}
	\begin{subfigure}{0.49\linewidth}
		\includegraphics[width=0.99\textwidth]{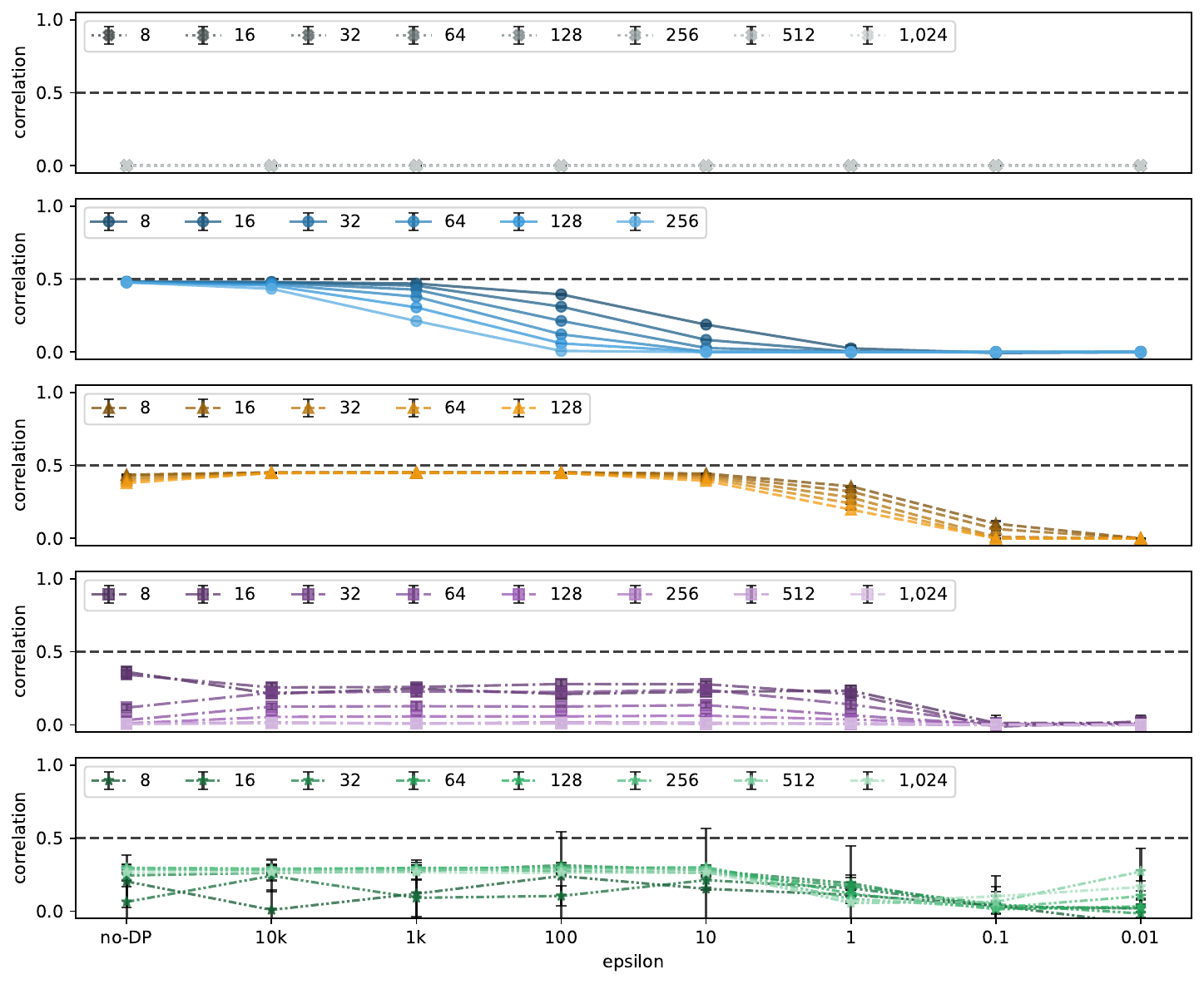}
		\caption{Varying $d$ and $n=16k$}
	\end{subfigure}
	\caption{T1: Off-diagonal pairwise correlation for different $\epsilon$ levels, on \emph{Corr~Gauss}, varying $n$ and $d$.}
	\label{fig:corr_off_diag_corr}
\reduce
\end{figure*}

To stress-test scalability {\em claims} from leading commercial companies~\cite{accelario2023, datagen2023, syntho2023, gretel2023}--that synthetic data can be scaled and delivered in minutes even for very large training datasets--we set practical time constraints.
Specifically, we discard models that take longer than 60 minutes to train.
In all experiments, this only affects the graphical models, as discussed later.
While, in theory, the graphical models could scale beyond these limits (in terms of data size and time), doing so would result in increased computational costs, which might not be practical or desirable.
This approach also allows us to feasibly test a significant number of models, totaling 21k.

\descr{Analysis.}
The graphical models scale polynomially with $d$ and quickly approach the 60 minutes limit; for instance, it takes PrivBayes 35 minutes to fit on a 256-dimensional dataset while MST requires 36 minutes for 128 dimensions.
This finding contradicts previous work~\cite{mckenna2021winning, mckenna2022asimple}, which does not test either model on more than 100 columns but concludes that MST is more scalable than PrivBayes.
Training PrivBayes and MST on wider datasets exceeds the set limit, so we refrain from doing so for all tasks.
Increasing $n$ has a minimal impact on time.

On the other hand, increasing $d$ or $n$ slows the GANs, but they train on all data settings within the time limit.
In other words, throughout our experiments, all GAN models either train for the preset number of iterations or until there is available privacy budget.
The increase in $d$ impacts the size of the first layer of the discriminator and the last layer of the generator.
By contrast, increasing $n$ leads to more iterations since we fix the number of epochs.
PATE-GAN is more efficient than DP-WGAN because PATE only introduces noise to the teacher-discriminator votes, whereas DP-SGD modifies the per-instance gradients and adds noise to all discriminator layers.
In fact, applying DP only materially slows DP-WGAN.

For all models except Independent, the data generation step only takes a fraction of the training time\arxiv{ (as shown in Tables~\ref{tab:scal_gen_rows} and~\ref{tab:scal_gen_cols} in Appendix~\ref{app:scalability})}.

\descr{Take-Aways.}
Deep generative models are far more scalable than graphical models.
Indeed, training deep generative models on high-dimensional tabular data is accessible with relatively modest computational resources (only CPU).
Comparing PrivBayes and MST, however, shows that the former can handle datasets with more features given the same computational constraints (256 vs. 128).

\subsection{T1: Statistics}
\label{subsec:statistics}

\descrfirst{Setup.}
We generate synthetic datasets with all models on \emph{Eye~Gauss} and \emph{Corr~Gauss} with varying $d$, $n$, and $\epsilon$ and capture different statistics.
For \emph{Eye~Gauss}, we show the marginal mean and pairwise correlation (excluding the diagonal) averaged across all columns, while for \emph{Corr~Gauss}, in addition to the marginal mean, we break down the pairwise correlations into off-diagonal and other.

We plot the off-diagonal pairwise correlation for \emph{Corr~Gauss} in Figure~\ref{fig:corr_off_diag_corr}.
We visualize the remaining statistics for \emph{Eye~Gauss} and \emph{Corr~Gauss} in Appendix~\arxiv{\ref{app:statistics}}\ccsver{A.2 of the paper's full version~\cite{ganev2023understanding}.}\arxiv{ (see Figure~\ref{fig:eye_mean},~\ref{fig:corr_mean} for mean and Figure~\ref{fig:eye_other_corr},~\ref{fig:corr_other_corr} for other pairwise correlation).}

\descr{Analysis.}
Overall, MST captures the distributions best, especially the mean and other pairwise correlations (likely due to the explicit modeling of structural zeros), and it is the least sensitive to changes in $n$ and $d$.
The off-diagonal pairwise correlation in Figure~\ref{fig:corr_off_diag_corr} shows that, for $n\leq4k$, the correlation score {\em improves} when privacy is applied and exceeds the ``no-DP'' values for $\epsilon\geq10$.
In essence, DP acts as regularization when the model has insufficient data to capture the distribution.
To our knowledge, this is the first finding of this kind for non-deep generative DP models.
Previous studies focused on CNNs~\cite{pearce22can} and GANs~\cite{ganev2022dp}, which are neural networks relying on DP-SGD and PATE, respectively.
Note that MST does not scale beyond 128 features within the set time limit.

PrivBayes performs as expected, with the mean and other pairwise correlations remaining close to 0 for all $n$ and $d$ across both datasets.
However, for off-diagonal pairwise correlation (see Figure~\ref{fig:corr_off_diag_corr}), there is a monotonic improvement with increasing $n$ and deterioration with increasing $d$, reaching the baseline levels of Independent for different levels of $\epsilon$.
This demonstrates that both MST and PrivBayes suffer from the curse of dimensionality, which contradicts recent work by~\citet{li2023statistical}.

\begin{figure*}[t!]
	\centering
	\begin{subfigure}{0.51\linewidth}
		 \includegraphics[width=0.99\textwidth]{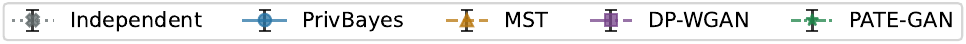}
	\end{subfigure}
	\centering
	\begin{subfigure}{0.495\linewidth}
		 \includegraphics[width=0.99\textwidth]{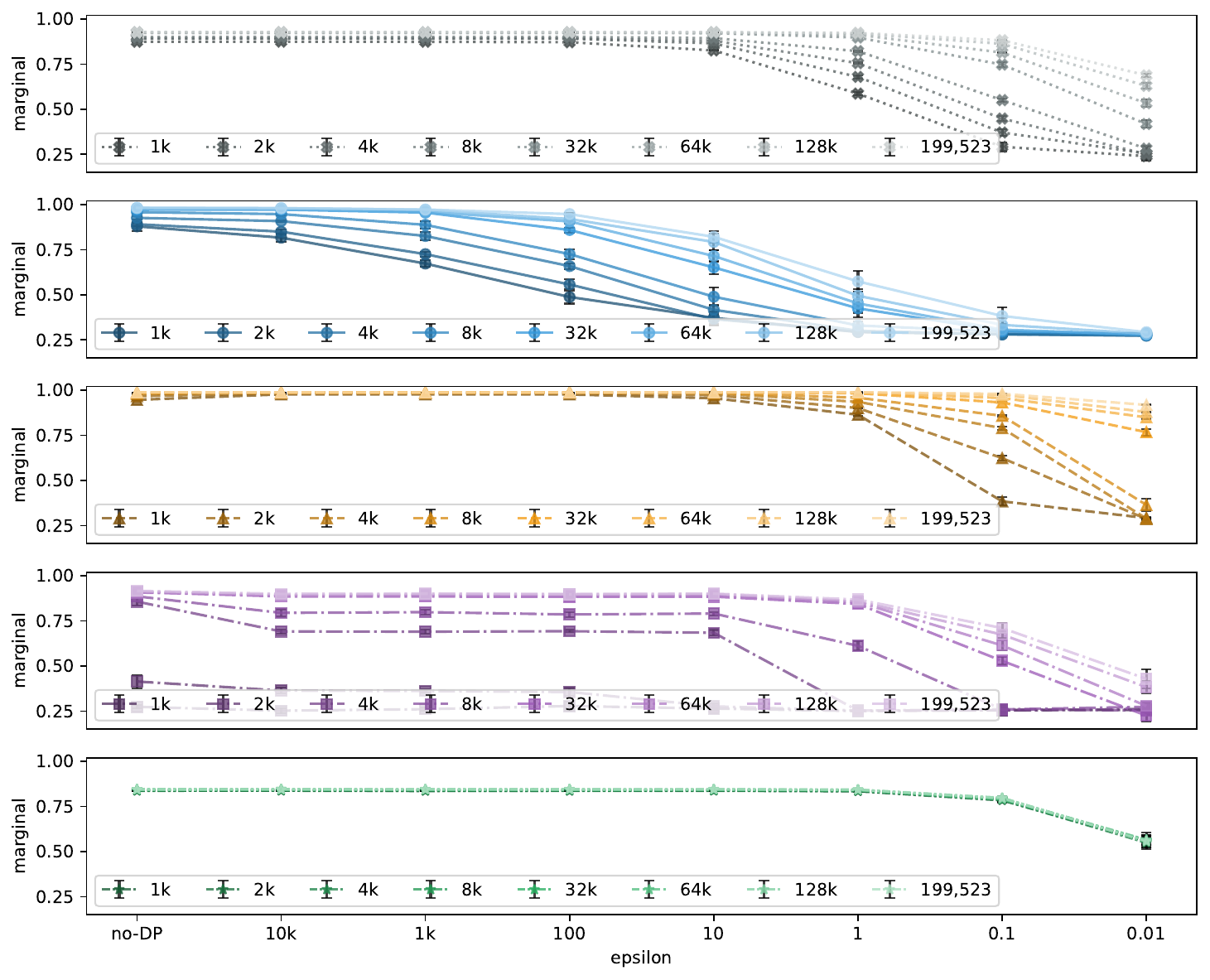}
		\caption{Marginal}
	\end{subfigure}
	\begin{subfigure}{0.495\linewidth}
		\includegraphics[width=0.99\textwidth]{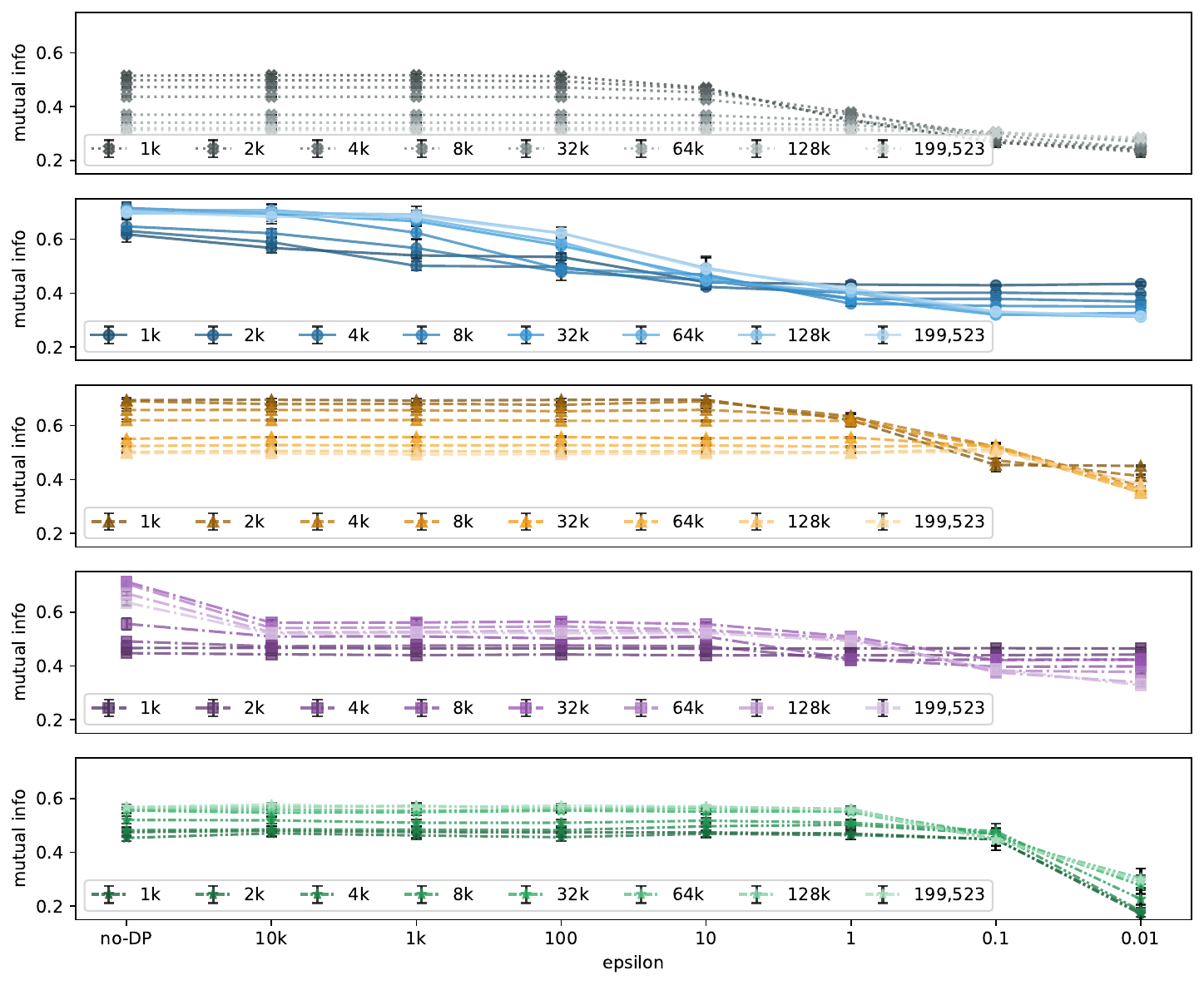}
		\caption{Mutual information}
		\label{fig:census_mi}
	\end{subfigure}
	\caption{T2: Marginal and pairwise mutual information similarity for different $\epsilon$ levels, on \emph{Census}, varying $n$.}
	\label{fig:census_sim_mi}
\end{figure*}

\begin{figure*}[t!]
	\centering
	\begin{subfigure}{0.59\linewidth}
		 \includegraphics[width=0.99\textwidth]{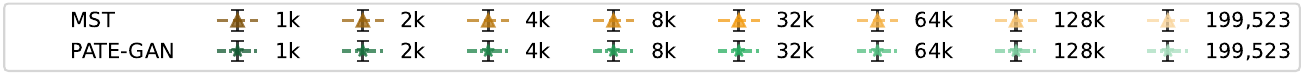}
	\end{subfigure}
	\centering
	\begin{subfigure}{0.495\linewidth}
		 \includegraphics[width=0.99\textwidth]{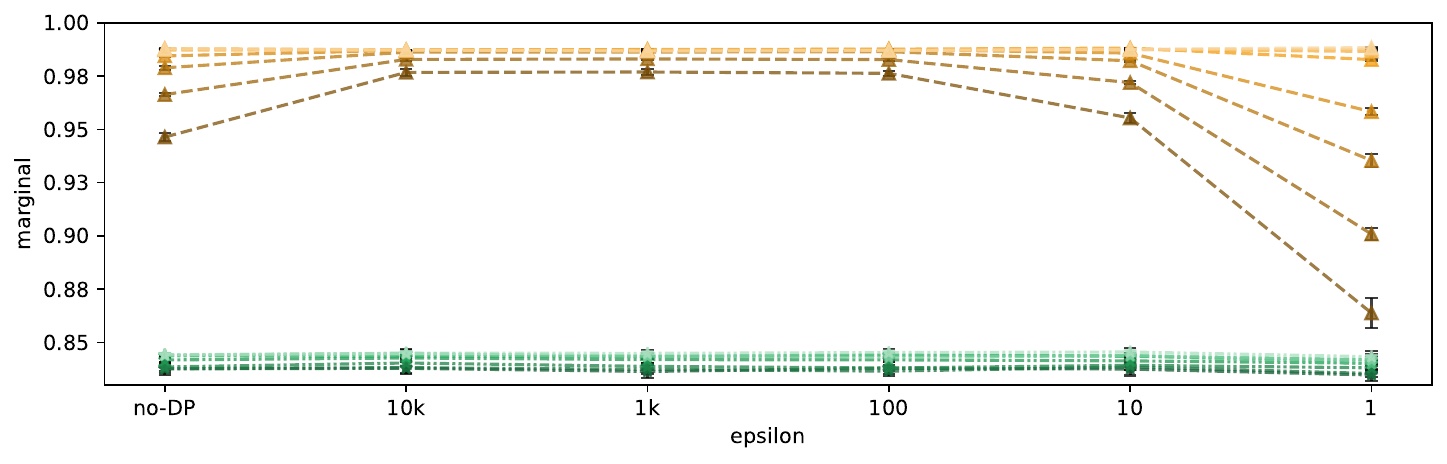}
		\caption{Marginal}
		\label{fig:census_sim_mst_pategan}
	\end{subfigure}
	\begin{subfigure}{0.495\linewidth}
		\includegraphics[width=0.99\textwidth]{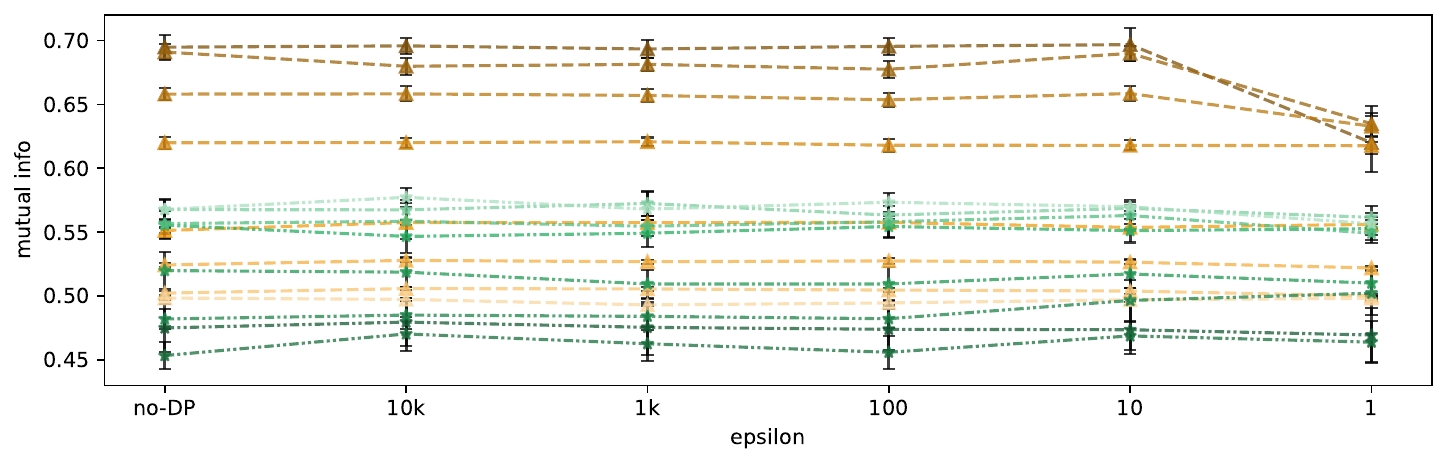}
		\caption{Mutual information}
		\label{fig:census_mi_mst_pategan}
	\end{subfigure}
	\caption{T2: Marginal and pairwise mutual information similarity zoomed-in for MST and PATE-GAN for different $\epsilon$ levels and $n$ (\emph{Census}).}
	\label{fig:census_sim_mi_mst_pategan}
\reduce
\end{figure*}

The GANs behave less predictably, and their performance is not monotonic when the dimensions are varied.
In all scenarios, PATE-GAN outperforms DP-WGAN, closely matching MST for $\epsilon>0.1$ for \emph{Eye~Gauss}.
For \emph{Corr~Gauss}, however, it cannot capture the off-diagonal correlation sufficiently well, creating data with correlation closer to 0.3 as opposed to 0.5 for the real one.
Interestingly, varying the dataset dimensions affects DP-WGAN differently –- for both datasets, increasing $n$ yields more correlated distributions with mean around 0, while, for \emph{Corr~Gauss}, increasing $d$ beyond 128 makes the model generate data with relatively uncorrelated columns ($\leq0.2$) but with mean further away from 0 ($>0.5$).
In fact, for \emph{Corr~Gauss} with $d=32$, the model fails to distinguish between off-diagonal and other correlations and creates data with uniform correlation.
At the very least, both PATE-GAN and DP-WGAN outperform Independent (which expectedly yields 0) in capturing the off-diagonal correlations for \emph{Corr~Gauss}.

\descr{Take-Aways.}
Graphical models outperform GANs at simple tasks like capturing statistics, making them more suitable to applications involving aggregated data.
MST is the best model overall, and its performance benefits from some degree of noise when training data is limited, while PrivBayes displays the most consistent behavior.

\subsection{T2: Similarity}
\label{subsec:similarity}

\descrfirst{Setup.}
We train all models on \emph{Diabetes}, \emph{Covertype}, and \emph{Census} with different values of $n$ and report marginal and pairwise mutual information similarities between the real and synthetic datasets.

\begin{figure*}[t!]
	\centering
	\begin{subfigure}[b]{0.35\linewidth}
		\centering
		\includegraphics[width=1\textwidth]{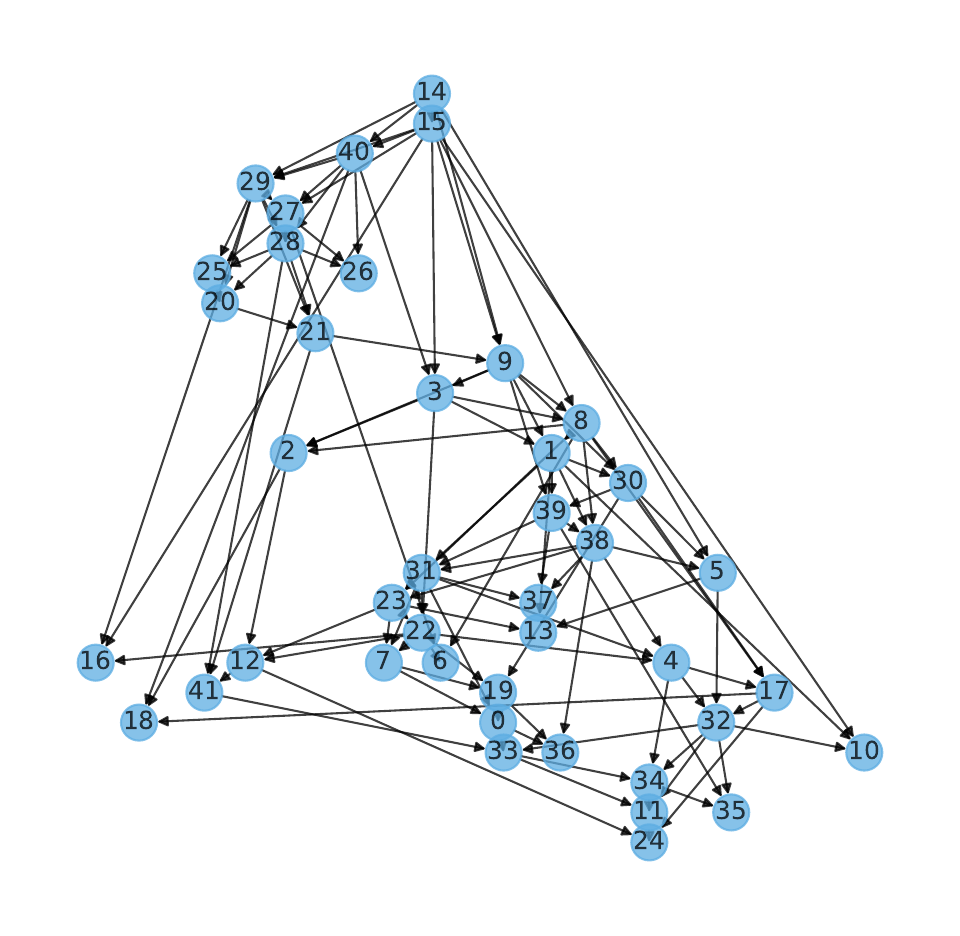}
		\caption{PrivBayes (directed acyclic graph)}
		\label{fig:census_network_privbayes}
	\end{subfigure}
	\begin{subfigure}[b]{0.35\linewidth}
		\centering
		\includegraphics[width=1\textwidth]{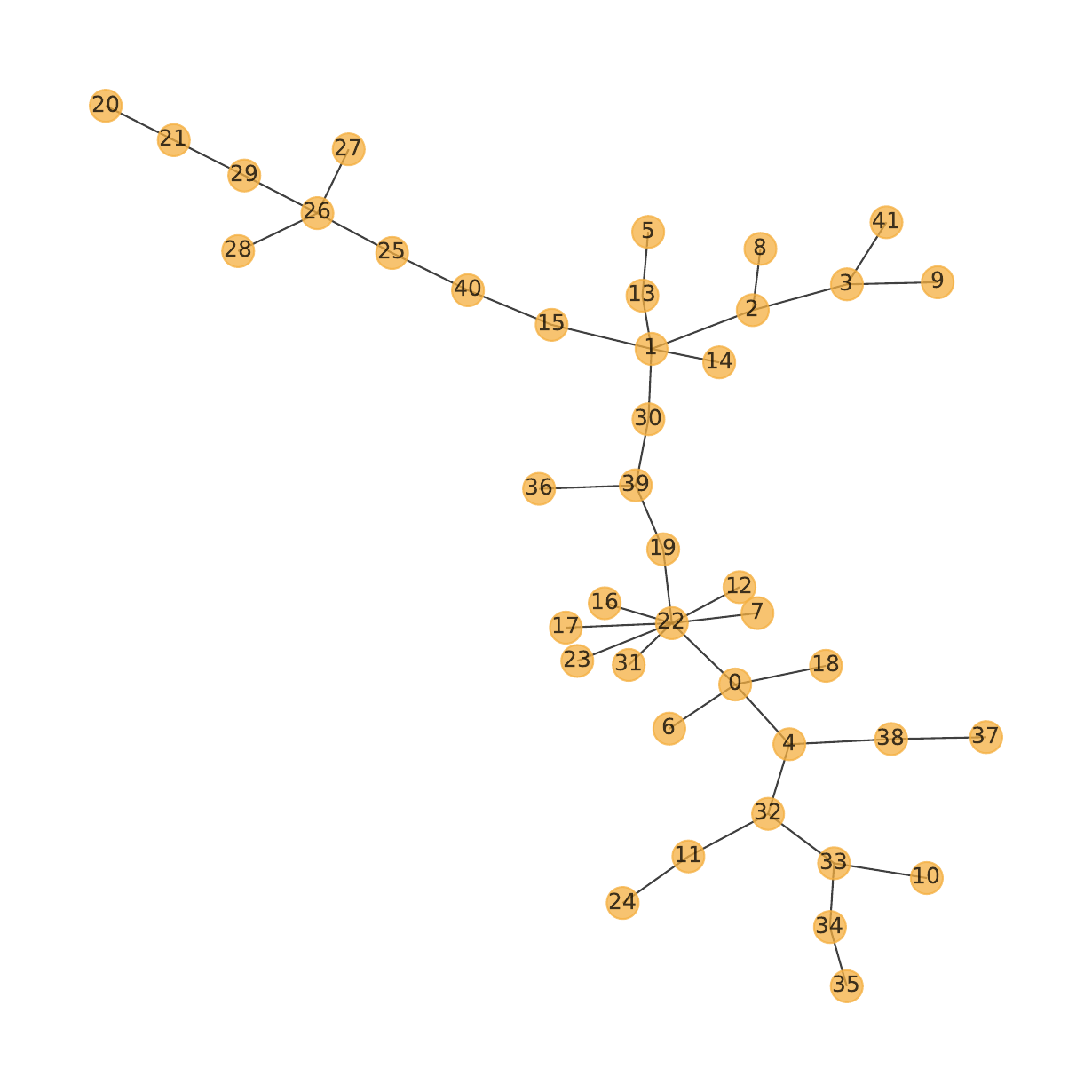}
		\caption{MST (undirected graph)}
		\label{fig:census_network_mst}
	\end{subfigure}
	\caption{T2: Example fitted networks for PrivBayes (with network degree 3) and MST with $\epsilon=1$, on \emph{Census}. The nodes correspond to the columns in the dataset, while the edges denote dependencies between them. For PrivBayes, the edges represent conditional distributions, for MST, they represent 2-way marginal counts; both are noisily measured to capture a collection of low-dimensional distributions and are used to generate synthetic data.}
	\label{fig:census_network_privbayes_mst}
	\reduce
\end{figure*}

In Figure~\ref{fig:census_sim_mi}, we plot the two metrics for \emph{Census}.
Furthermore, we show a zoomed-in plot for MST and PATE-GAN (Figure~\ref{fig:census_sim_mi_mst_pategan}), visualized PrivBayes and MST networks (Figure~\ref{fig:census_network_privbayes_mst}), and broken down mutual information for connected vs.~non-connected nodes (Figure~\ref{fig:census_network_mi_privbayes_mst}).
In Appendix~\arxiv{\ref{app:similarity}}\ccsver{A.3 of the paper's full version~\cite{ganev2023understanding}}, we also report the similarities for \emph{Diabetes} and \emph{Covertype}\arxiv{ in Figure~\ref{fig:diabetes_sim_mi} and~\ref{fig:covertype_sim_mi}}.

\descr{Analysis.}
Looking at the marginal similarity across all datasets, Independent, PrivBayes, MST, and DP-WGAN behave as expected: higher $n$ results in monotonically better scores.
An increase in $n$ also enhances the scores for PATE-GAN, though the sensitivity is almost negligible.
Additionally, DP-WGAN needs at least 4k or 8k training points to produce datasets with meaningful marginal similarity to the real one.
The fact that Independent is very competitive and sometimes outperforms PrivBayes and the GANs for $\epsilon\leq10$ should not come as a surprise as it only captures the marginal distributions and does not ``waste'' any privacy budget for other purposes.
On the one hand, even though MST is the best-performing model in the majority of settings, PATE-GAN becomes more accurate when there is little data ($n\leq4k$) and strict privacy constraints ($\epsilon<1$).
On the other hand, as depicted in Figure~\ref{fig:census_sim_mst_pategan}, when training data is scarce ($n<4k$), introducing a degree of privacy ($10k\leq\epsilon\leq10$), once again, {\em helps} MST.

\begin{figure*}[t!]
	\centering
	\begin{subfigure}{0.55\linewidth}
		 \includegraphics[width=0.99\textwidth]{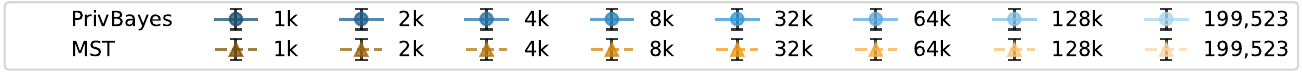}
	\end{subfigure}\\
	\centering
	\begin{subfigure}{0.4975\linewidth}
		 \includegraphics[width=1\columnwidth]{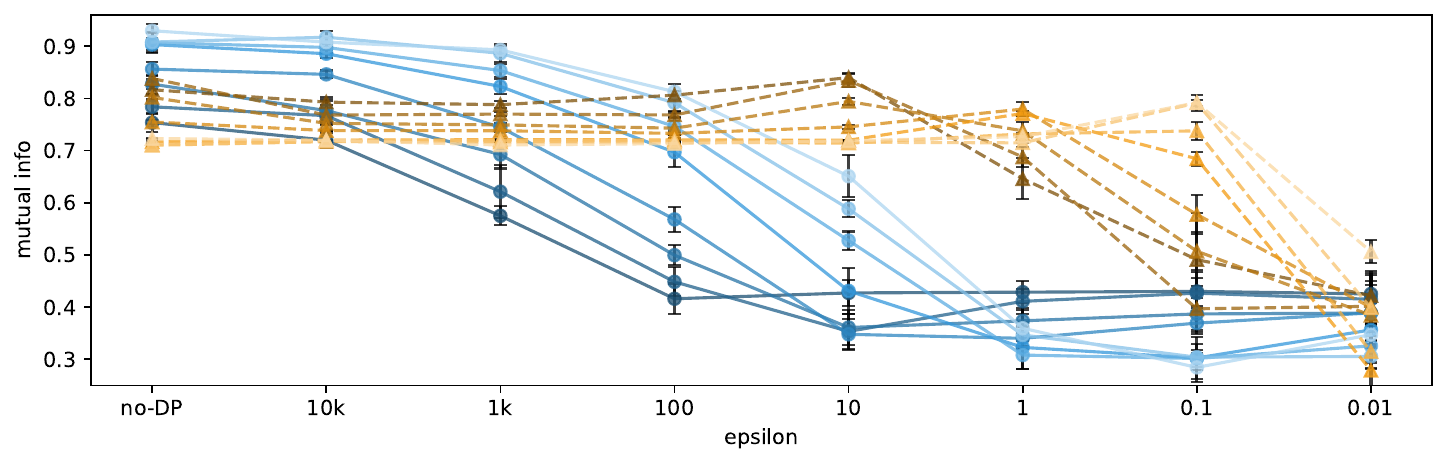}
		\caption{Connected nodes}
		\label{fig:census_network_mi_privbayes}
	\end{subfigure}
	\begin{subfigure}{0.4975\linewidth}
		\includegraphics[width=1\textwidth]{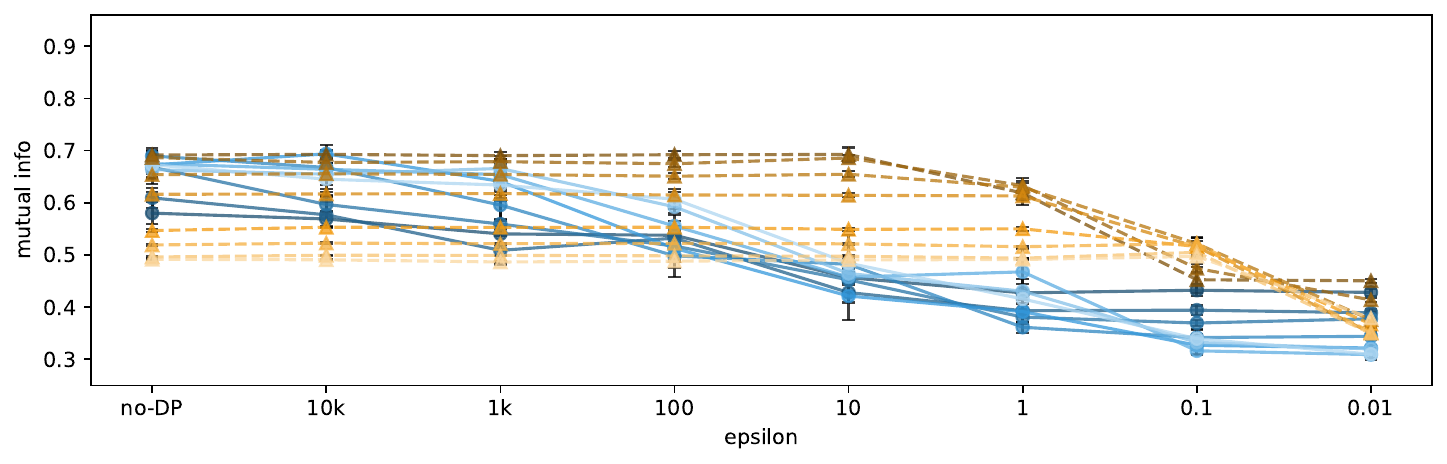}
		\caption{Non-connected nodes}
		\label{fig:census_network_mi_mst}
	\end{subfigure}
	\caption{T2: Pairwise mutual information (connected/non-connected nodes extracted from the fitted networks for PrivBayes and MST) similarity for different $\epsilon$ levels, on \emph{Census}, varying $n$.}
	\label{fig:census_network_mi_privbayes_mst}
\reduce
\end{figure*}

For mutual information similarity, see Figure~\ref{fig:census_mi} \arxiv{,~\ref{fig:diabetes_mi}, and~\ref{fig:covertype_mi}} for the \emph{Census}\arxiv{, \emph{Diabetes}, and \emph{Covertype}} dataset\arxiv{s, respectively (the latter appear in Appendix~\ref{app:similarity})}.
The fact that Independent does not score 0 across the pairwise relationships should not be entirely surprising, as maintaining high degrees of marginal similarity can also lead to preserving some degree of correlations between variables, as noted by~\cite{stadler2022synthetic}.
Quite unexpectedly, adding more data points to the training data {\em yields worse} scores for both Independent and MST.
This is a previously unobserved phenomenon for MST and could be due to the overfitting of the model to its objective function, specifically, targeting all 1-way and 41 2-way marginals (for \emph{Census}) as illustrated in Figure~\ref{fig:census_network_mst}.
This could lead to MST's inability to capture all 2-way marginals effectively.
The trend is further highlighted in Figure~\ref{fig:census_network_mi_privbayes_mst}.
Here, MST exhibits a more significant drop in scores of connected vs. non-connected nodes compared to PrivBayes (which models a much larger number of connections, 120, as shown in Figure~\ref{fig:census_network_privbayes}).

For $n>8k$, PATE-GAN also {\em outperforms} MST as shown in Figure~\ref{fig:census_mi_mst_pategan}; this could be explained by its training procedure, which prioritizes creating plausible synthetic data points (i.e., implicitly maintaining correlations between columns).
For \emph{Covertype}, however, PATE-GAN fails to capture the mutual information sufficiently well.
Most likely due to the nature of the dataset, which is heavily imbalanced and has a multi-class target column, all undesirable conditions for PATE-GAN as observed by~\cite{ganev2022dp}.

Finally, with PrivBayes and DP-WGAN for \emph{Census} and \emph{Covertype}, increasing $n$ helps the mutual information score when reducing the level of privacy, but only up to $\epsilon=1$.
Whereas for $\epsilon\leq1$, one would be better off training the models on {\em less} data.

\begin{figure*}[t!]
	\centering
	\begin{subfigure}{0.0763\linewidth}
		 \includegraphics[width=0.99\textwidth]{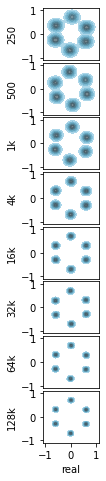}
		\caption{Real}
	\end{subfigure}
	\begin{subfigure}{0.175\linewidth}
		 \includegraphics[width=0.99\textwidth]{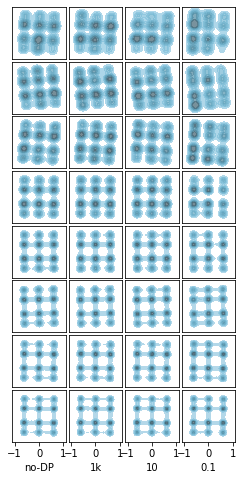}
		\caption{Independent}
	\end{subfigure}
	\begin{subfigure}{0.175\linewidth}
		 \includegraphics[width=0.99\textwidth]{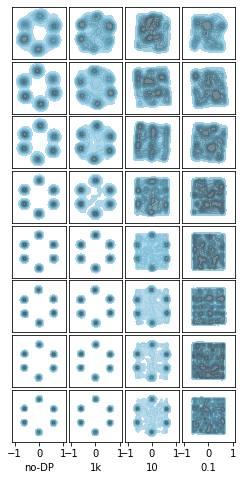}
		\caption{PrivBayes}
	\end{subfigure}
	\begin{subfigure}{0.175\linewidth}
		 \includegraphics[width=0.99\textwidth]{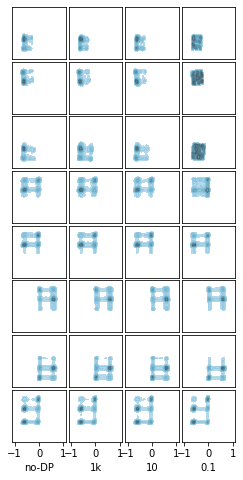}
		\caption{MST}
	\end{subfigure}
	\begin{subfigure}{0.175\linewidth}
		 \includegraphics[width=0.99\textwidth]{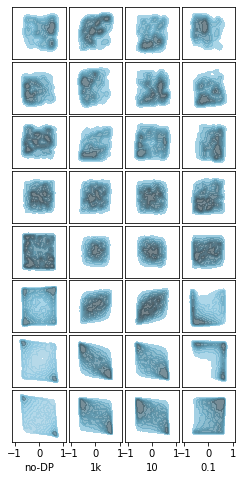}
		\caption{DP-WGAN}
	\end{subfigure}
	\begin{subfigure}{0.175\linewidth}
		 \includegraphics[width=0.99\textwidth]{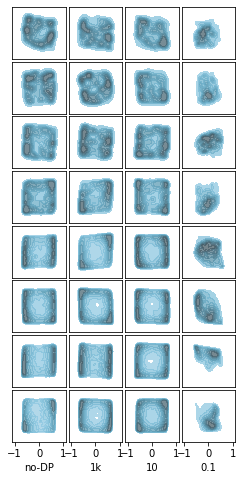}
		\caption{PATE-GAN}
	\end{subfigure}
	\caption{T3: KDE on the first 2 PCA principles for different $\epsilon$ levels, on \emph{Mix~Gauss~Unsup}, varying $n$.}
	\label{fig:mix_pca_rows}
\end{figure*}

\begin{figure*}[t!]
	\centering
	\begin{subfigure}{0.0763\linewidth}
		 \includegraphics[width=0.99\textwidth]{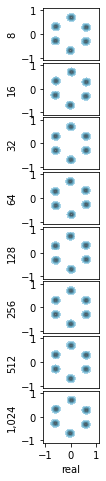}
		\caption{Real}
	\end{subfigure}
	\begin{subfigure}{0.175\linewidth}
		 \includegraphics[width=0.99\textwidth]{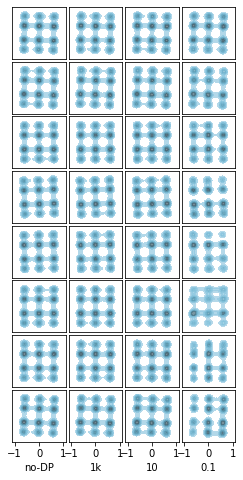}
		\caption{Independent}
	\end{subfigure}
	\begin{subfigure}{0.175\linewidth}
		 \includegraphics[width=0.99\textwidth]{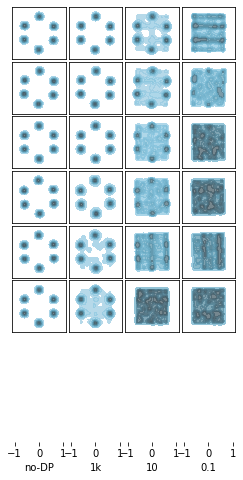}
		\caption{PrivBayes}
	\end{subfigure}
	\begin{subfigure}{0.175\linewidth}
		 \includegraphics[width=0.99\textwidth]{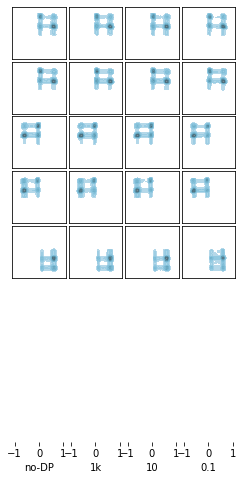}
		\caption{MST}
	\end{subfigure}
	\begin{subfigure}{0.175\linewidth}
		 \includegraphics[width=0.99\textwidth]{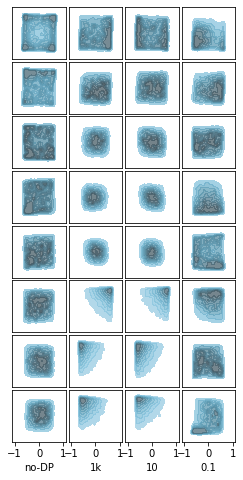}
		\caption{DP-WGAN}
	\end{subfigure}
	\begin{subfigure}{0.175\linewidth}
		 \includegraphics[width=0.99\textwidth]{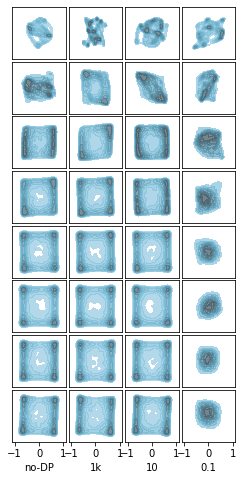}
		\caption{PATE-GAN}
	\end{subfigure}
	\caption{T3: KDE on the first 2 PCA principles for different $\epsilon$ levels, on \emph{Mix~Gauss~Unsup}, varying $d$.}
	\label{fig:mix_pca_cols}
\reduce
\end{figure*}

\descr{Take-Aways.}
While MST generally outperforms the other models in capturing marginal similarity, it can, in fact, overfit when trained on more data, leading to poorer performance in modeling pairwise mutual information.
The GANs, particularly PATE-GAN, become very competitive at capturing more complex correlations, except in cases of extreme imbalance.
This makes MST the better choice when one wants to preserve the similarity of the synthetic data.

\subsection{T3: Clustering}
\label{subsec:pca}

\descr{Setup.}
We evaluate all models using the four \emph{Gauss} datasets, where we vary $d$ and $n$, as well as the \emph{Plants} dataset, where we vary $n$.
We visualize the Kernel Density Estimation (KDE) of the first two PCA/UMAP components. %
Additionally, we apply Mixture of Gaussians models to the projected data and report the silhouette scores for the resulting clusters in \emph{Mix~Gauss~Unsup} and \emph{Plants}.

The 2d PCA for \emph{Mix~Gauss~Unsup} is depicted in Figure~\ref{fig:mix_pca_rows} and~\ref{fig:mix_pca_cols}.
In Appendix~\arxiv{\ref{app:pca}}\ccsver{A.4 of the full version~\cite{ganev2023understanding}}, we present the PCA plots for the remaining \emph{Gauss} datasets\arxiv{ in Figure~\ref{fig:eye_pca_rows}--\ref{fig:mix_target_pca_cols}}, and the UMAP for \emph{Plants}\arxiv{ is shown in Figure~\ref{fig:plants_umap}}.
The silhouette scores for \emph{Mix~Gauss~Unsup} and \emph{Plants} are \ccsver{also visualized.}\arxiv{plotted in Figure~\ref{fig:mix_sil} and Figure~\ref{fig:plants_sil}, respectively.}

\descr{Analysis.}
Analyzing the PCA plots of the \emph{Gauss} datasets, no model manages to replicate all of them to a satisfactory degree.
PrivBayes appears to come closest for $\epsilon\geq10$, and for the \emph{Mix~Gauss~Unsup} (as seen in Figure~\ref{fig:mix_pca_rows} and~\ref{fig:mix_pca_cols}) and \emph{Mix~Gauss~Sup} datasets, it is the only model that distinctly separates the first two columns, forming the six clusters, from the rest, containing noise.
When a tighter privacy budget is applied, or the dataset dimensions are increased, the variance of the synthetic data increases, too, for all datasets.
While MST excels with \emph{Eye~Gauss} and \emph{Corr~Gauss} (as already seen in Section~\ref{subsec:statistics}), it ultimately fails with the other two, producing some difficult-to-define structure, probably due to mode collapse.
As expected, Independent fails to capture \emph{Mix~Gauss~Unsup} distributions.
The GAN models, unfortunately, also underperform on these datasets, though they do manage to capture the data's overall range.
The influence of $n$ and $d$ is minimal.
For $d\geq32$ and $\epsilon>0.1$, PATE-GAN generates data that bears some resemblance to the original, but it is shaped more as a box rather than a ring.
Regarding the \emph{Plants} dataset\arxiv{ (see Figure~\ref{fig:plants_umap} in Appendix~\ref{app:pca})}, all models seem to replicate it well, with the potential exception at $\epsilon=0.1$.

The silhouette scores for \emph{Mix~Gauss~Unsup}\arxiv{ (see Figure~\ref{fig:mix_sil} in Appendix~\ref{app:pca})} appear quite variable and noisy.
This variability is likely because the clustering was based on the first two PCA components rather than the entire datasets, a decision made due to computational time constraints.
In general, these scores align with our observations from the PCA plots.
PrivBayes delivers strong performance for $n>1k$.
The scores for MST are notably affected by variations in $n$, resulting in unpredictable outcomes.
PATE-GAN consistently achieves comparatively good silhouette scores, equal or higher than 0.45 for all $n$.

Turning our attention to \emph{Plants}\arxiv{ (see Figure~\ref{fig:plants_sil} in Appendix~\ref{app:pca})}, we again observe variable and noisy trends, despite the UMAP plots appearing cohesive.
The scores fluctuate between 0 and 0.5, which is slightly higher than the score derived from the real data.
Notably, the MST scores frequently dip below 0, especially when $n\geq8k$.

\descr{Take-Aways.}
Clustering proves challenging for all models, as none seem to sufficiently capture the underlying distributions and separate signal from noise, perhaps except for PrivBayes.
Overall, our analysis highlights the need to exercise caution when performing clustering tasks, and prompts a challenging open research question.

\begin{figure*}[t!]

	\centering
	\begin{subfigure}{0.55\linewidth}
		 \includegraphics[width=0.99\textwidth]{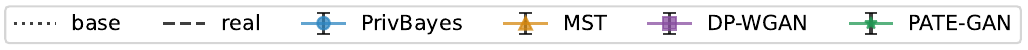}
	\end{subfigure}
	\centering
	\begin{subfigure}{0.495\linewidth}
		 \includegraphics[width=0.99\textwidth]{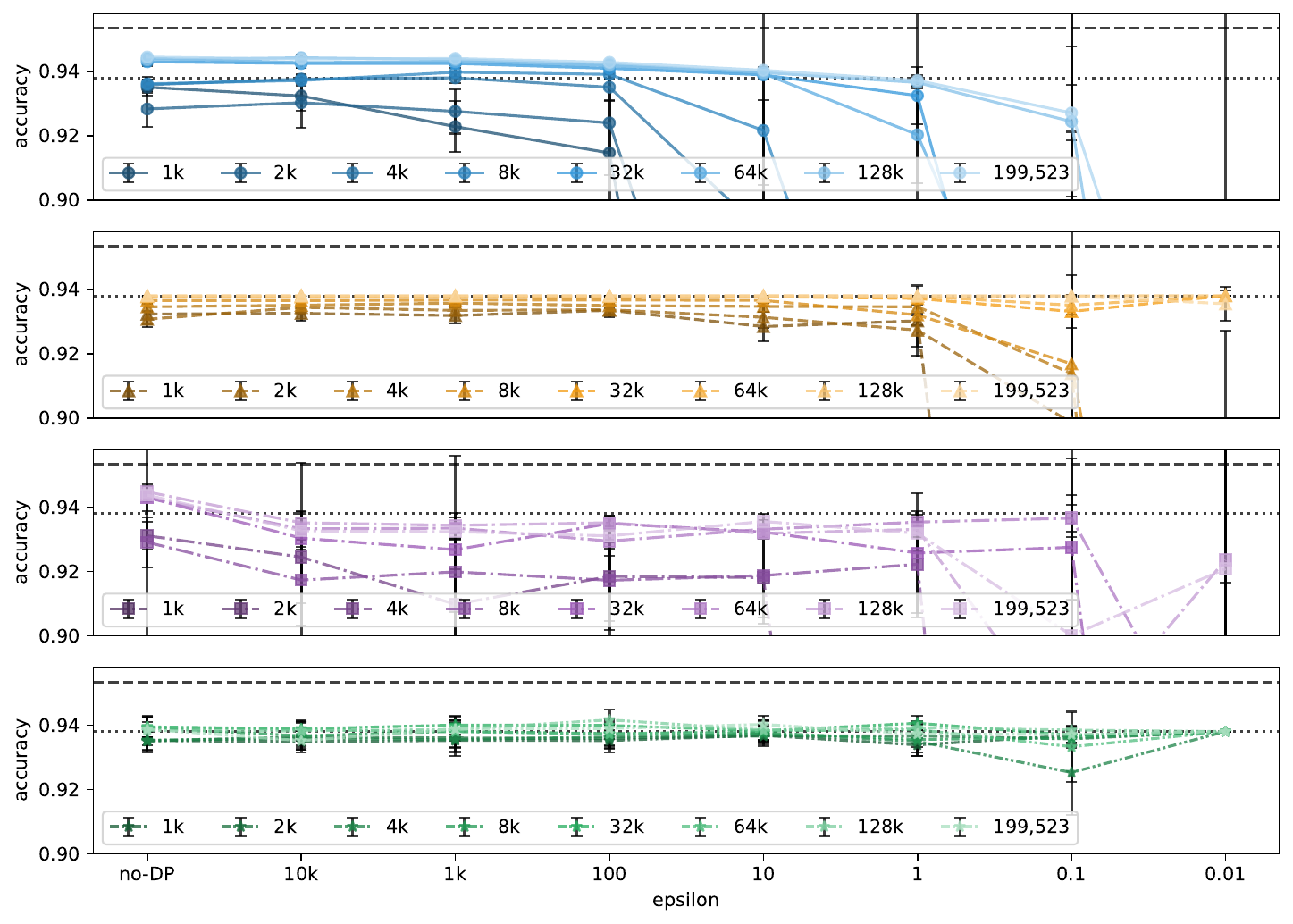}
		\caption{Accuracy}
	\end{subfigure}
	\begin{subfigure}{0.495\linewidth}
		\includegraphics[width=0.99\textwidth]{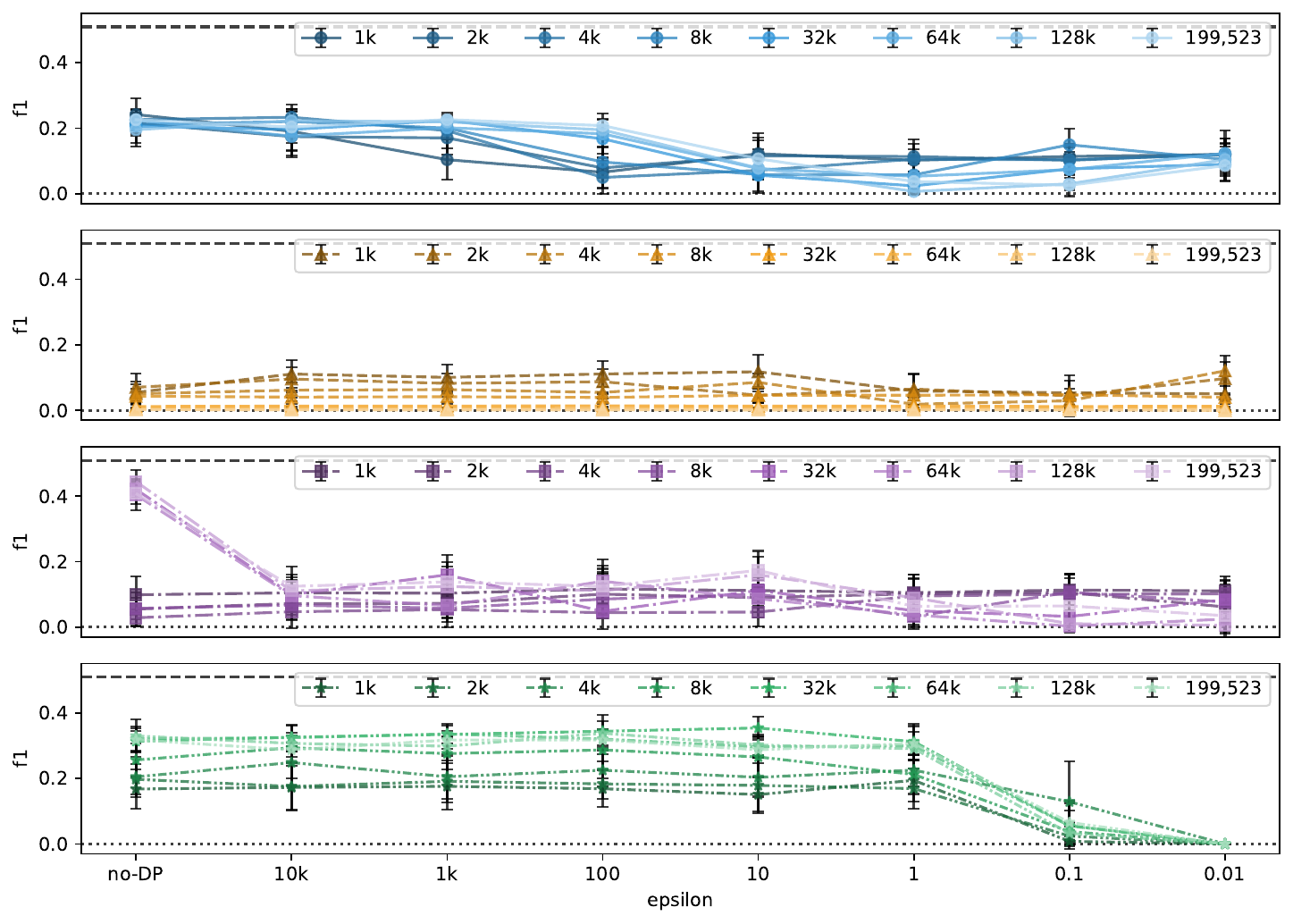}
		\caption{F1}
	\end{subfigure}
	\caption{T4: Accuracy and F1 for different $\epsilon$ levels, on \emph{Census}, varying $n$.}
	\label{fig:census_acc_f1}
\reduce
\end{figure*}
\begin{figure}[t!]
	\centering
	\begin{subfigure}{1\linewidth}
		 \includegraphics[width=0.999\textwidth]{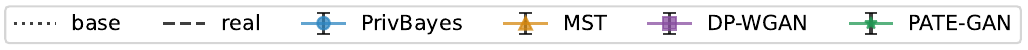}
	\end{subfigure}\\
	\begin{subfigure}{1\linewidth}
		\includegraphics[width=0.98\linewidth]{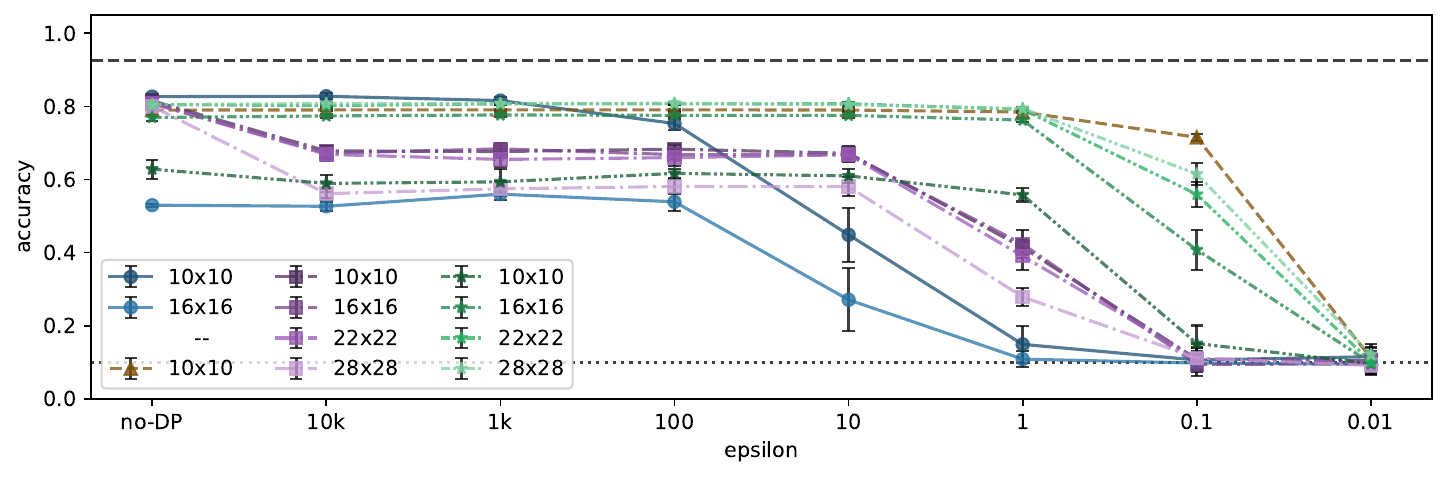}
	\end{subfigure}
	\caption{T4: Accuracy for varying $\epsilon$ and $d$, \emph{MNIST}.} %
	\label{fig:mnist_acc}
\reduce
\end{figure}

\subsection{T4: Classification}
\label{subsec:classification}

\descr{Setup.}
We run classification on synthetic data generated by all models for the datasets \emph{Mix~Gauss~Sup}, \emph{Census}, \emph{Connect~4}, and \emph{MNIST}.
For \emph{Mix~Gauss~Sup} we vary both $d$ and $n$, for \emph{Census}, \emph{Connect~4} we only vary $n$, while for \emph{MNIST}, we vary the resolution, $d$.
We report accuracy metrics for all datasets and include the F1-score for \emph{Census}, \emph{Connect~4} as they are slightly imbalanced.

We report the accuracy and F1-score for \emph{Census} and accuracy for \emph{MNIST} in Figure~\ref{fig:census_acc_f1} and~\ref{fig:mnist_acc}.
In Appendix~\arxiv{\ref{app:classification}}\ccsver{A.5 of the paper's full version~\cite{ganev2023understanding}}, we include accuracy results for \emph{Mix Gauss Sup}\arxiv{ (see Figure~\ref{fig:mix_target_acc})} and both accuracy and F1-score results for \emph{Connect~4}\arxiv{ (refer to Figure~\ref{fig:connect_4_acc_f1})}.

\descr{Analysis.}
Looking at \emph{Mix~Gauss~Sup}, PrivBayes, once again, behaves as expected for different $n$ and $d$.
There is a common trend for MST and DP-WGAN, as both models need at least 16k data points to achieve better than random accuracy.
Similarly, if there are too many features, 128 for MST and $\geq$256 for DP-WGAN, the accuracy approaches the random baseline.
Varying $n$ does not seem to be a significant factor for PATE-GAN as accuracy tends to be close to the real one for $\epsilon>0.1$.
Increasing $d$, however, has a negative effect, which contrasts with previous experiments.

As for \emph{Census}, we consider both accuracy and F1 as the dataset is imbalanced (93.8\% of the people make less than \$50k/year).
Somewhat unexpectedly, DP-WGAN comes closest to the real F1 baseline but only for the ``no-DP'' case.
Overall, PATE-GAN is the only model with an F1-score consistently close to 0.35, provided that it was trained on at least 8k points.
For MST, increasing $n$ helps accuracy (see Figure~\ref{fig:census_acc_f1}) but at the expense of F1, i.e., the classifiers trained on the synthetic data {\em overfit} to the majority class.
This behavior aligns with previous studies on DP generative models trained on imbalanced data~\cite{ganev2022robin} even though they have not tested MST explicitly.

We see similar trends for \emph{Connect~4}; PATE-GAN is the only model comfortably beating the random baseline and achieving the most consistent F1-score, MST has higher accuracy for {\em lower} F1 with increasing $n$, while DP-WGAN is the best model for ``no-DP.''

These trends are very close to the ones observed in the similarity experiments from Section~\ref{subsec:similarity}; the relative overperformance of the GAN models compared to MST is in direct contradiction with previous studies~\cite{tao2022benchmarking}.
Although we cannot say for sure, we believe that this might, in part, be due to not using the original GAN implementations but relying on third-party ones.
Unfortunately,~\citet{tao2022benchmarking} dismiss them as good candidates while arguably overstating the capabilities of other approaches.

As for the accuracy on \emph{MNIST} (Figure~\ref{fig:mnist_acc}), we see that more features (i.e., images with higher resolution/better quality) deteriorate the performance of PrivBayes and DP-WGAN but help PATE-GAN.
In principle, we could expect higher-resolution images to improve both GAN models, but this is the case only for PATE-GAN.
However, none of the models approach the real baseline (0.9 accuracy).
While this is expected for graphical models, it is somewhat surprising for the GANs.
We believe this is because both GANs rely only on feed-forward layers and not CNNs.
MST trained on 10x10 images performs very well, achieving 0.8 accuracy, on par with PATE-GAN ($d>$ 10x10) for $\epsilon>0.1$.
Again, this might be surprising, as MST does not explicitly model higher-level dependencies, which is important in the image domain but might not be necessary for a simple dataset like \emph{MNIST}.

\descr{Take-Aways.}
The GANs are very adept at more complex tasks like classification, with PATE-GAN notably outperforming the graphical models, achieving, on average, 133\% higher F1-score on \emph{Census}.
As observed previously, as the number of training records increases, MST may overfit, resulting in a tradeoff between higher accuracy and lower F1-score.

\section{Related Work}
\label{sec:rw}

\descr{DP Queries \& Data Dimensionality.}
\citet{hay2016principled} benchmark 15 DP algorithms for range queries over 1 and 2-dimensional datasets, showing that increasing values of $n$ reduce error.
For small $n$, data-dependent algorithms tend to perform better; for large $n$, data-independent algorithms dominate.
For (more complex) predicate counting queries and higher dimensional data, \citet{mckenna2021hdmm} propose a method with consistent utility improvements and show that increasing $d$ results in more significant errors.
However, they experiment with datasets with at most 15 features, and their model struggles to scale beyond 30-dimensional datasets.

\descr{DP Classifiers \& Data Dimensionality.}
For predictive models, more data and longer training usually lead to better performance. %
This also holds for DP logistic regression learned via empirical risk minimization and objective perturbation~\cite{chaudhuri2011differentially}: for large $n$, the cost function tends to the non-private one, and while the scale of the noise is independent of $d$, there is a linear shift to the objective function that does not affect the optimization if it is strongly peaked~\cite{antonova2016practical}.
However, this does not always hold; for instance, methods based on iterative training like DP-SGD trade off training steps and noise added per iteration~\cite{near2021programming}.

\descr{DP Generative Models \& Data Dimensionality.}
Unlike query answering and classification tasks, the output of generative models lies in a high-dimensional space.
Thus, it likely has much higher sensitivity, making its analysis much more complex.
Furthermore, private synthetic data generation is computationally challenging (exponential in $d$ in the worst-case scenario, i.e., all 2-way marginals are preserved~\cite{dwork2009complexity, ullman2011pcps}).
Nevertheless, worst-case complexities do not rule out practical algorithms (such as those introduced above); indeed, if {\em most}, rather than all, correlations are preserved, one can build computationally efficient algorithms~\cite{boedihardjo2021covariance}.

Since there is no ``one-size-fits-all'' DP synthetic data generation method, researchers have highlighted the need to empirically assess the privacy and fidelity of the data on a {\em per-case} basis~\cite{jordon2022synthetic}.
While \citet{hay2016principled} provide a set of standardized evaluation principles for DP query answering algorithms, including varying domain size, scale, and shape, no similar study focuses on synthetic data generation.
In fact, current frameworks for synthetic data evaluation~\cite{arnold2020really} do not consider varying $n$ and $d$ as essential factors, and benchmark studies~\cite{tao2022benchmarking, mckenna2022aim, liu2022utility} do not use datasets with more than 41 features.

\descr{DP Generative Models Benchmarks.}
Furthermore, these benchmark papers~\cite{tao2022benchmarking, mckenna2022aim, liu2022utility} claim that DP graphical models are superior to deep generative models.
Specifically, \citet{tao2022benchmarking} benchmark 12 DP generative models on similarity and classification tasks and conclude that MST is the best-performing model, while DPGAN~\cite{xie2018differentially} (which is similar to DP-WGAN) and PATE-GAN fail to beat simple baselines like Independent.
\citet{liu2022utility} claim that DP deep generative models are incapable of recovering utility and that PrivBayes performs far better than them.

To the best our knowledge, state-of-the-art graphical models have not been evaluated on datasets with more than 100 dimensions even though MST is presented as a generic and scalable solution~\cite{mckenna2022asimple}.
This might be problematic as, e.g., \citet{takagi2021p3gm} argue that PrivBayes only performs well for datasets with simple dependencies and a few features, while MST cannot reconstruct the essential information from limited information (i.e., 1 and 2-way marginals) required for more complex classification tasks.
Also, \citet{li2022optimizing} observe that the two models are usually tested on tabular datasets with dozens of dimensions and claim that they still suffer from the ``curse of dimensionality''~\cite{bellman1957dynamic}.

\descr{Empirical Evaluations of DP.}
Researchers have also conducted empirical privacy evaluations for DP models, aka auditing, whereby membership inference attacks~\cite{hayes2019logan, hilprecht2019monte, lu2019empirical, stadler2022synthetic} are used to establish empirical privacy guarantees and compared to the theoretical ones provided by the DP bounds.
Building on previous work on auditing discriminative DP models~\cite{nasr2021adversary, nasr2023tight}, emerging research has also begun to audit generative models~\cite{houssiau2022tapas, lokna2023group}.

\arxiv{\descr{\em Remarks.}
Our work bridges several gaps in the empirical understanding of how DP generative models behave, presenting an extensive and comprehensive evaluation, comparing graphical models and deep generative models on diverse dataset sizes and shapes, as well as a variety of downstream tasks with different complexity.}

\section{Discussion \& Conclusion}
This paper presented a comprehensive measurement of how various DP generative models distribute their privacy budget.
We experimented with different modeling approaches and DP mechanisms and focused on the challenges posed by datasets of expanding dimensions and varying privacy budgets, measuring the effects of these factors on the quality of the generated synthetic data on several tasks.
Overall, we are confident that our work will facilitate the understanding of which models work best for specific settings, datasets, and downstream tasks, thus helping practitioners integrate DP generative models for tabular data in real-world pipelines.

In the rest of this section, we summarize the lessons learned and discuss first-cut solutions, future research directions, and the limitations of our work.

\subsection{Lessons Learned and Recommendations}
Our experimental evaluation sheds light on the effects of different generative models, various DP mechanisms, and different dimensions of the training dataset on downstream tasks computed over the synthetic data.
Overall, these effects are mixed depending on the setting, and no single best-performing model exists.
In the process, we learned a few valuable lessons:
\begin{itemize}[leftmargin=0.5cm]
	\item PrivBayes exhibits the most predictable and monotonic behavior, possibly due to its relative modeling simplicity.
	\item Our experiments evidently refute claims that MST is scalable~\cite{nist2018differential, tao2022benchmarking, mckenna2022asimple} as it cannot actually handle more than 128 columns within practical time constraints.
	\item More training data helps MST on simple tasks but can unexpectedly cause overfitting, leading to worse performance on tasks requiring complex relationships (e.g., classification).
	\item In some instances, a small degree of DP noise can act as regularization and help when there is limited amount of data.
	\item The effects of the data dimensions are more unpredictable (more variable and usually not monotonic) for GANs.
	While they underperform at simple tasks on controllable datasets, we consistently observe that PATE-GAN could be quite competitive at more challenging tasks and improve with higher dimensions.
\end{itemize}

\descr{Recommendations for Practitioners.}
Our analysis paves the way for the following actionable recommendations for practitioners looking to use DP generative models to build synthetic datasets:
\begin{enumerate}[leftmargin=0.5cm]
	\item If the training data is small (e.g., the number of features $d$ is in the order of 100 or less) and the target downstream task is relatively simple (e.g., capturing statistics/marginals), one should be using graphical models.
	\item If the dataset is high-dimensional, there are enough records, and the downstream task is more complex (i.e., machine learning-related), deep generative models are likely a better choice.
	\item Regardless, with strict privacy constraints (i.e., low privacy budgets), dataset sampling and/or early stopping is likely to prove beneficial with respect to utility.
	\item In general, despite their disadvantages, graphical models are likely to see wider adoption due to their predictability and stronger performance on simple tasks, which carry less risk.
	\item Overall, our work confirms that practitioners should ensure that the synthetic data is not only of sufficient quality but also evaluated using appropriate metrics/privacy budgets.
\end{enumerate}

\descr{Implications for Researchers.}
Our measurement also sheds light on a few research gaps. %
For instance, we hope that the privacy engineering community will assist practitioners and stakeholders in identifying the use cases where synthetic data can be used safely, ideally even in a semi-automated way.
Moreover, we believe researchers could be incentivized -- including through public initiatives (e.g., similar to the NIST challenges~\cite{nist2018the, nist2018differential, nist2020differential}), joint industry-academia events, conference tracks, etc. -- to provide actionable guidelines to understand the distributions, types of data, tasks, and settings, where one could achieve reasonable privacy-utility tradeoffs via synthetic data, and through which model(s).
Finally, we call for researchers to extend our type of empirical measurement from tabular data to other kinds of data (e.g., images) to derive actionable guidelines regarding privacy-utility tradeoffs based on the datasets/tasks at hand.

\begin{figure}[t!]
	\centering
	\begin{subfigure}{0.99\linewidth}
		 \includegraphics[width=0.999\textwidth]{plots2/mnist/legend_base_real.pdf}
	\end{subfigure}
	\begin{subfigure}{1\linewidth}
		\includegraphics[width=0.99\textwidth]{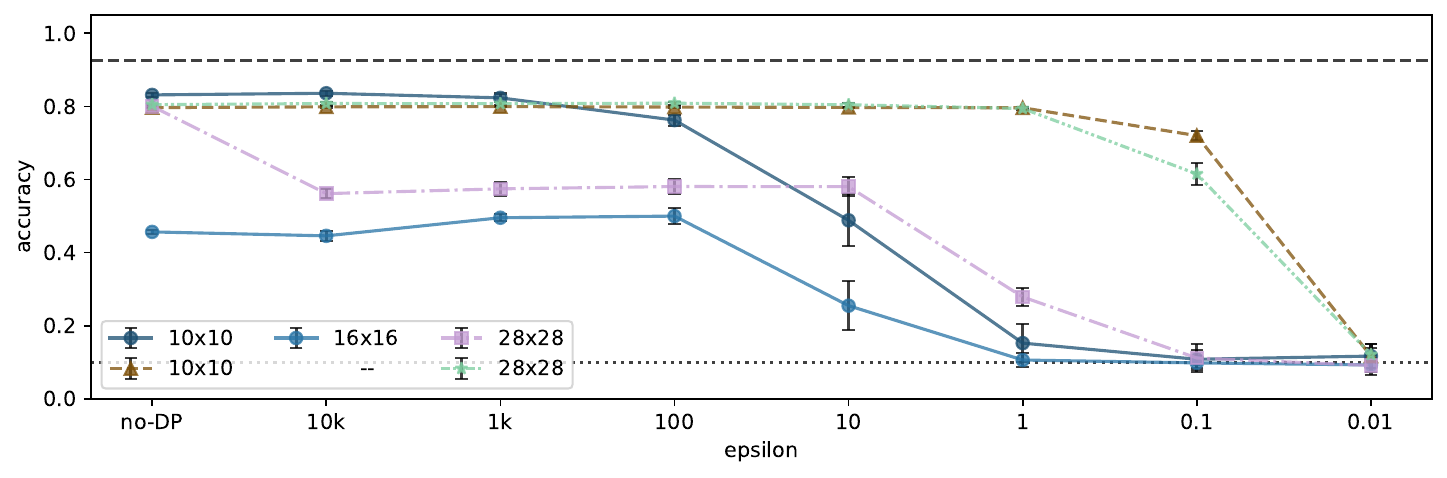}
	\end{subfigure}
	\caption{T4: Accuracy for different $\epsilon$ levels, \emph{MNIST}, upscaling to 28x28.}
	\label{fig:upscaled_mnist_acc}
\reduce
\end{figure}

\subsection{Possible Improvements}
Next, we build on the lessons learned and suggest some improvements to the models we studied.
In the process, we also discuss possible future research directions.

\descr{Increasing Number of Features.}
Our evaluation shows that increasing the data features results in synthetic data with progressively worse performance on the downstream task (except for PATE-GAN with \emph{MNIST}).
Furthermore, the graphical models (which perform better on lower-dimensional datasets) cannot scale beyond 128/256 dimensions within the set time constraints.
A logical step toward improvement would be to try to reduce the dataset's dimensionality, train/generate synthetic data in the lower space (this would also help with the DP budget) using the better-suited models, and, if necessary, upscale to the original space.
\citet{tantipongpipat2021differentially} propose a similar approach, combining VAE and GAN.

As a proof of concept, in Figure~\ref{fig:upscaled_mnist_acc}, we downscale \emph{MNIST} images using standard tools to 10x10/16x16, train MST (on 10x10) and PrivBayes (on 10x10 and 16x16), generate synthetic images, and then upscale them back to 28x28.
Knowing the attribute bounds is not an unrealistic expectation since all currently proposed DP generative models in the literature make this implicit assumption to start with.
Comparing classifiers trained on both real and synthetic images, we observe that: 1) classifiers trained on MST-generated data perform very well (on par with PATE-GAN trained on the original images), and 2) neither of MST/PrivBayes lose utility compared to when they were tested on downscaled real images (Figure~\ref{fig:mnist_acc}).
Another way to improve on a particular task would be to investigate the most relevant/important features and spend the privacy budget strategically~\cite{rosenblatt2022spending}.
Alternatively, we could choose the simplest ``good enough'' model, e.g., Independent preserved marginal similarity even in high dimensions.

\begin{figure}[t!]
	\centering
	\begin{subfigure}{0.92\columnwidth}
		 \includegraphics[width=0.99\textwidth]{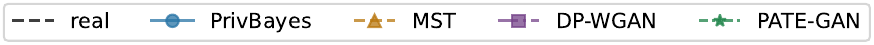}
	\end{subfigure}\\
	\centering
	\begin{subfigure}{0.99\columnwidth}
		 \includegraphics[width=0.99\textwidth]{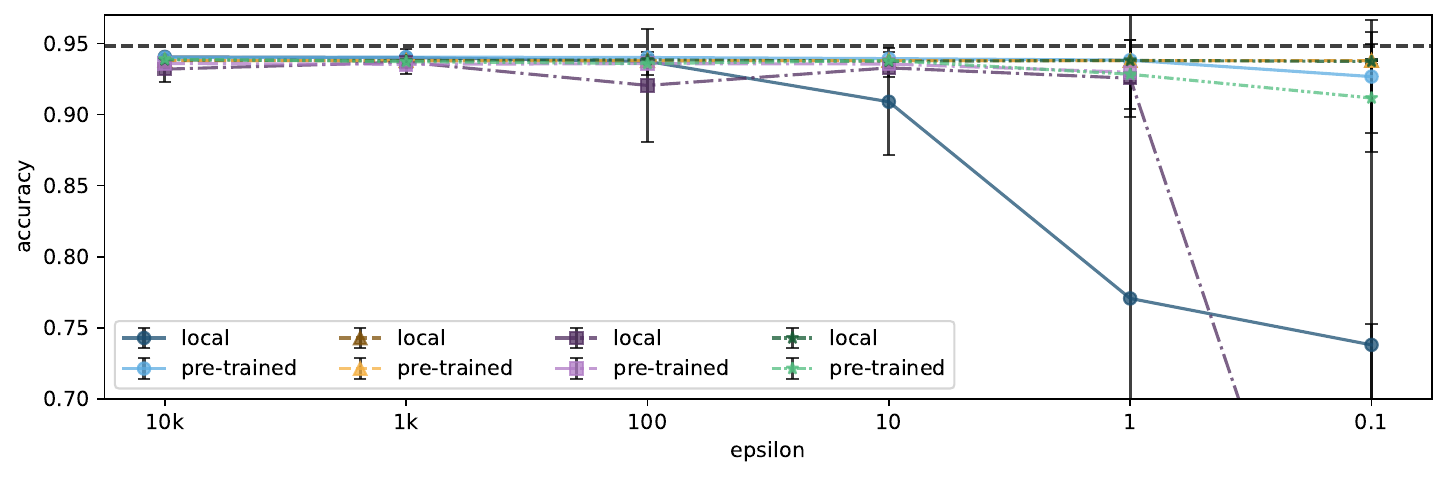}
		\caption{Accuracy}
	\end{subfigure}\\
	\begin{subfigure}{0.99\columnwidth}
		\includegraphics[width=0.99\textwidth]{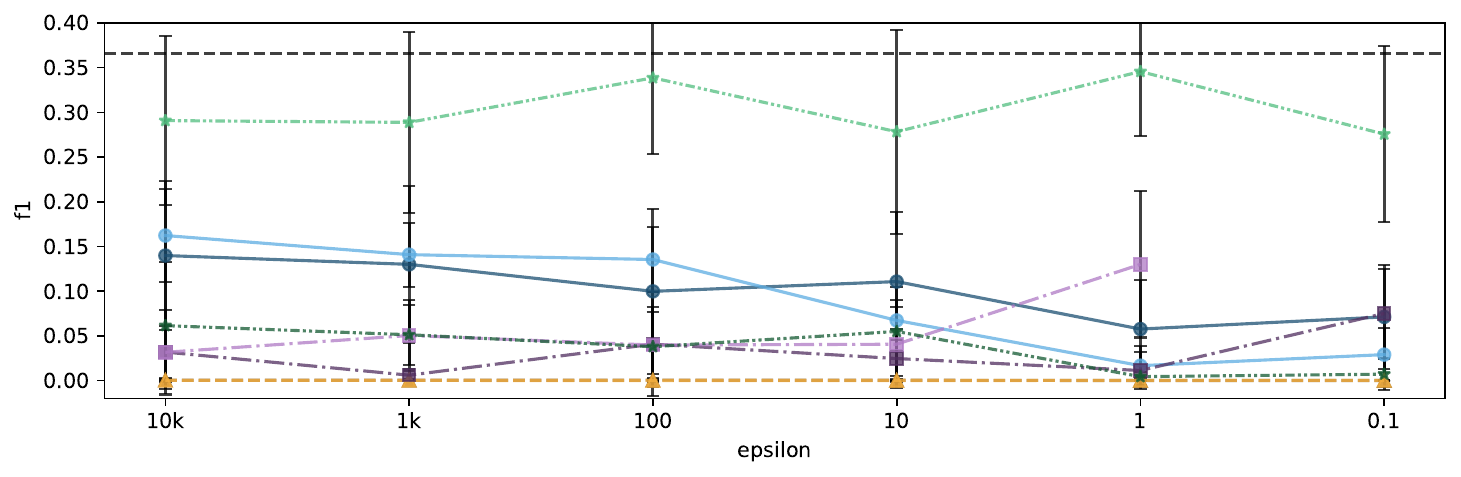}
		\caption{F1}
	\end{subfigure}
	\caption{T4: Accuracy and F1 on local data and using a pre-trained model for different $\epsilon$ levels, \emph{Census}.}
	\label{fig:census_pretrained_acc_f1}
\reduce
\end{figure}

\descr{Increasing Number of Rows.}
We also observe that more data does not always translate to improved quality for all models and evaluations (e.g., the GAN models and MST, apart from marginal similarity).
On the other hand, for some models (MST and DP-WGAN), a minimum data threshold is needed to perform better than random.
Therefore, more research is needed to find a good balance.
One avenue could be to investigate optimal times for early stopping or dataset sampling techniques.
Also, one could build relevant public datasets for the tabular domain and develop pre-trained models; researchers and practitioners could then fine-tune the models on their specific (private) dataset~\cite{harder2022differentially}.
This approach has proved to be very promising in other areas, including NLP~\cite{li2022large, yu2022differentially} and vision~\cite{de2022unlocking, golatkar2022mixed, tramer2021differentially}.

As another proof of concept, in Figure~\ref{fig:census_pretrained_acc_f1}, we report accuracy and F1-scores from classifiers fitted on datasets generated by 1) generative models trained on local data only ($n$ approx. 16k or all individuals with known residence region in \emph{Census}), and 2) ``no-DP'' pre-trained generative models on a larger amount of data ($n$ approx. 180k) and fine-tuned on the local data.
PrivBayes benefits greatly from pre-training; its performance is close to the real data even for $\epsilon=0.1$.
Classifiers trained on MST and PATE-GAN data have satisfactory accuracy but display F1 close to 0.
Pre-training on bigger data alleviates this concern for PATE-GAN; in fact, F1 approaches the real baseline.
However, the effect on MST is negligible.
While it is not surprising that pre-training helps GANs, we observe some benefits for graphical models as well (only for PrivBayes), but we leave exploring this fully to future work. %

\descr{Future Work.}
Overall, our work takes an important step in studying state-of-the-art DP synthetic data generation models and their use for downstream tasks.
However, we only focus on two approaches -- graphical and deep generative models, as motivated in Section~\ref{sec:preliminaries} -- leaving, e.g., query-based approaches~\cite{vietri2020new, aydore2021differentially, liu2021iterative, vietri2022private, mckenna2022aim} to future work.
Also, similarly to previous studies~\cite{tao2022benchmarking, mckenna2022aim, stadler2022synthetic, ganev2022robin}, we re-use the default hyperparameters for all models; future work could try to optimize them further to add another dimension to the empirical comparison of graphical vs. deep generative models.
Finally, we plan to consider other factors of the training data, such as skewness and class/subgroup imbalance.

\subsection*{Acknowledgments}
We are grateful to the ACM CCS Program Committee for their valuable feedback and suggestions, which helped us significantly improve our paper.

\bibliographystyle{ACM-Reference-Format}

%
%

%
%

%
%
%

\appendix

\section{Additional Results and Plots}
\label{app:experiments}

\subsection{M1: Scalability}
\label{app:scalability}
We report summaries of the runtime of the generation step for \emph{Corr~Gauss} with varying $n$ and $d$ in Tables~\ref{tab:scal_gen_rows} and~\ref{tab:scal_gen_cols}.
The results are discussed in Section~\ref{subsec:scalability}.

\begin{table}[h]
	\centering
	\small
	\setlength{\tabcolsep}{4pt}
	\begin{tabular}{l@{}rrrrrrrr}
		\toprule
		\bf DP Model$\downarrow$ \bf $n\hspace{-0.05cm}\rightarrow$ & \bf 250 & \bf 500 & \bf 1k & \bf 4k & \bf 16k & \bf 32k & \bf 64k & \bf 128k \\
		\midrule
		Independent & 0.00 & 0.00 & 0.00 & 0.00 & 0.01 & 0.01 & 0.03 & 0.06 \\
		PrivBayes & 0.00 & 0.00 & 0.00 & 0.00 & 0.01 & 0.01 & 0.03 & 0.05 \\
		MST & 0.01 & 0.01 & 0.01 & 0.01 & 0.02 & 0.02 & 0.04 & 0.07 \\
		DP-WGAN & 0.00 & 0.00 & 0.00 & 0.00 & 0.01 & 0.01 & 0.02 & 0.04 \\
		PATE-GAN & 0.00 & 0.00 & 0.00 & 0.00 & 0.00 & 0.00 & 0.00 & 0.01 \\
		\bottomrule
	\end{tabular}
	\captionof{table}{M1: Runtime (in mins) of the model's generation step for fitted DP generative models, on \emph{Corr~Gauss}, varying $n$ and $d=32$.}
	\label{tab:scal_gen_rows}
\end{table}

\begin{table}[h]
	\centering
	\small
	\setlength{\tabcolsep}{4pt}
	\begin{tabular}{l@{}rrrrrrrr}
		\toprule
		\bf DP Model$\downarrow$ \bf $d\hspace{-0.05cm}\rightarrow$ & \bf 8 & \bf 16 & \bf 32 & \bf 64 & \bf 128 & \bf 256 & \bf 512 & \bf 1,024 \\
		\midrule
		Independent & 0.00 & 0.00 & 0.01 & 0.02 & 0.07 & 0.24 & 0.99 & 3.69 \\
		PrivBayes & 0.00 & 0.00 & 0.01 & 0.02 & 0.03 & 0.07 & & \\
		MST & 0.00 & 0.01 & 0.02 & 0.05 & 0.13 & & & \\
		DP-WGAN & 0.00 & 0.00 & 0.01 & 0.01 & 0.02 & 0.06 & 0.39 & 1.63 \\
		PATE-GAN & 0.00 & 0.00 & 0.00 & 0.00 & 0.01 & 0.05 & 0.30 & 1.57 \\
		\bottomrule
	\end{tabular}
	\captionof{table}{M1: Runtime (in mins) of the model's generation step for fitted DP generative models, on \emph{Corr~Gauss}, varying $d$ and $n=16k$.}
	\label{tab:scal_gen_cols}
\reduce
\end{table}

\begin{figure*}[t!]
	\centering
	\begin{subfigure}{0.51\linewidth}
		 \includegraphics[width=0.99\textwidth]{plots2/gaussians/legend_real.pdf}
	\end{subfigure}\\
	\centering
	\begin{subfigure}{0.495\linewidth}
		 \includegraphics[width=0.99\textwidth]{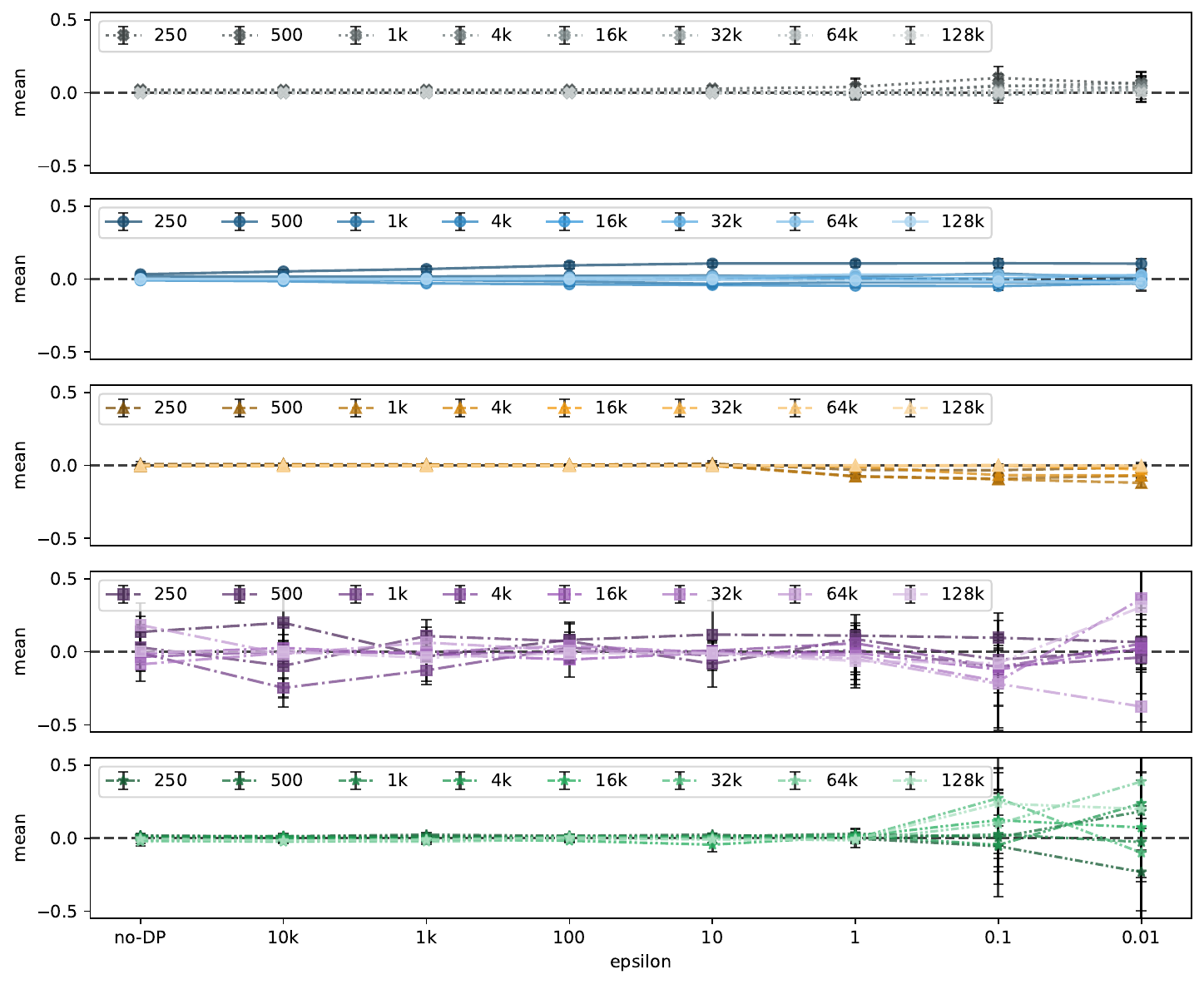}
		\caption{Varying $n$ and $d=32$}
	\end{subfigure}
	\label{fig:eye_mean_a}
	\begin{subfigure}{0.495\linewidth}
		\includegraphics[width=0.99\textwidth]{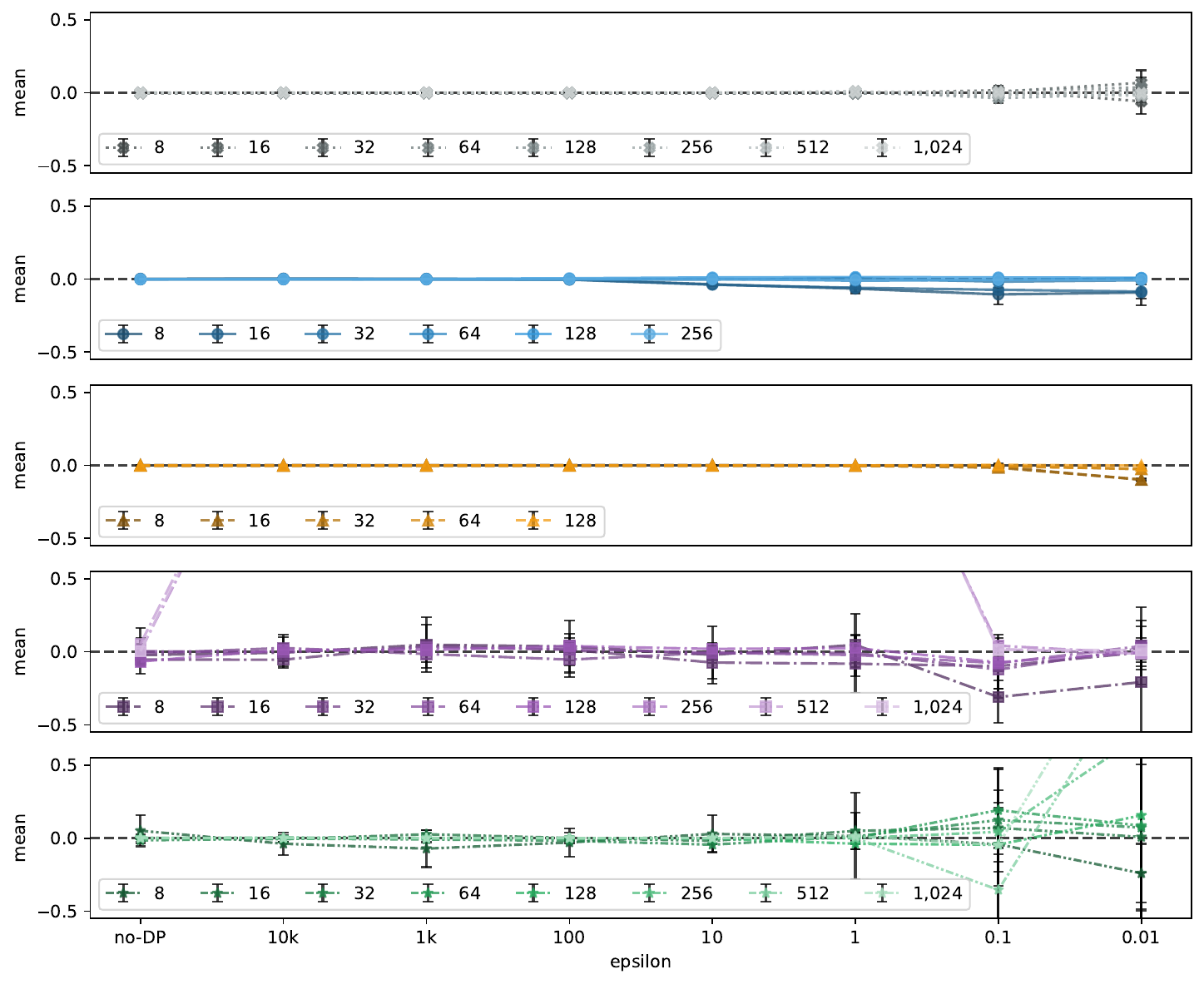}
		\caption{Varying $d$ and $n=16k$}
	\end{subfigure}
	\label{fig:eye_mean_b}
	\caption{T1: Marginal mean for different $\epsilon$ levels, on \emph{Eye~Gauss}, varying $n$ and $d$.}
	\label{fig:eye_mean}
\end{figure*}

\subsection{T1: Statistics}
\label{app:statistics}
\begin{figure*}[t!]
	\centering
	\begin{subfigure}{0.51\linewidth}
		 \includegraphics[width=0.99\textwidth]{plots2/gaussians/legend_real.pdf}
	\end{subfigure}
	\centering
	\begin{subfigure}{0.495\linewidth}
		 \includegraphics[width=0.99\textwidth]{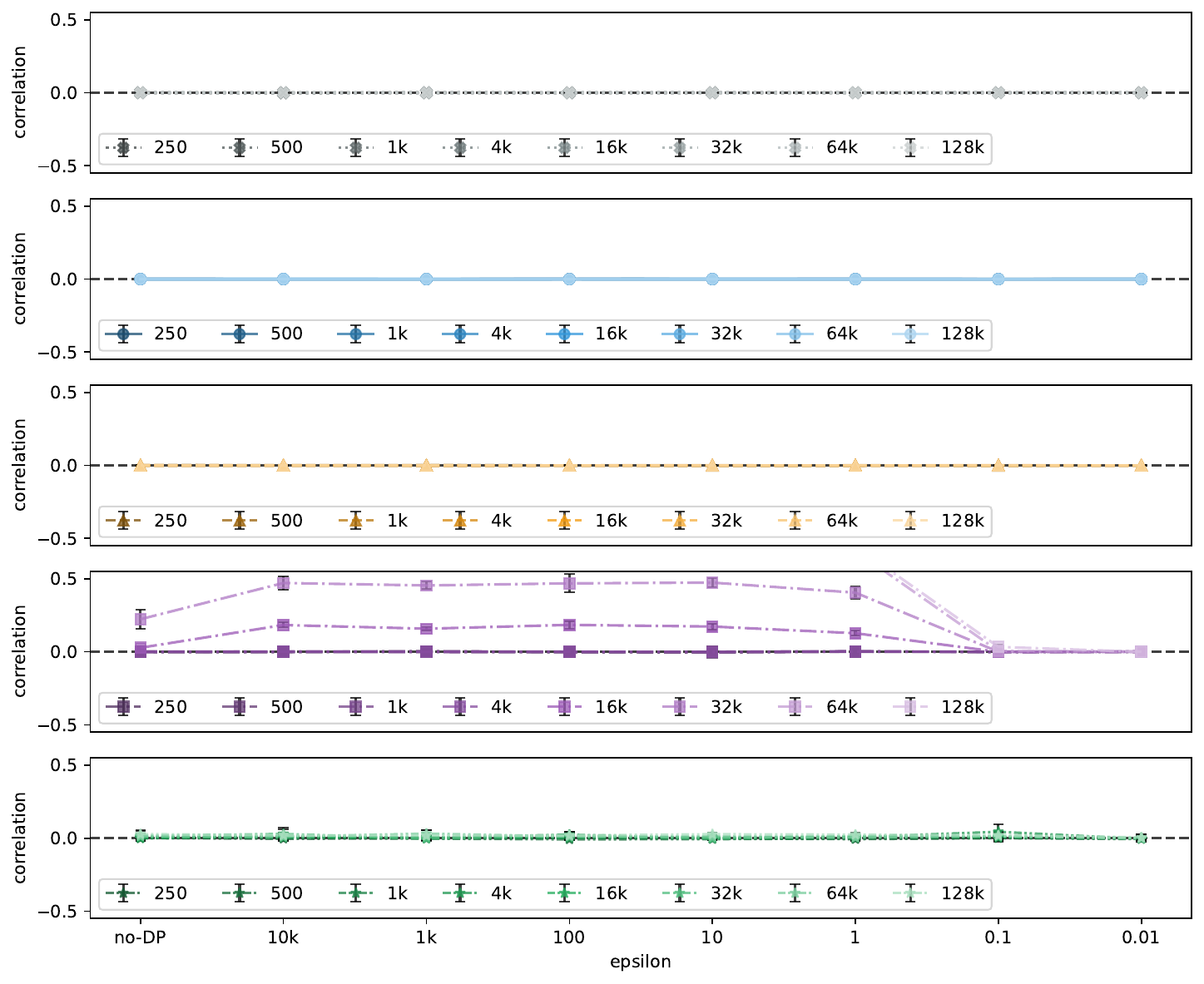}
		\caption{Varying $n$ and $d=32$}
	\end{subfigure}
	\begin{subfigure}{0.495\linewidth}
		\includegraphics[width=0.99\textwidth]{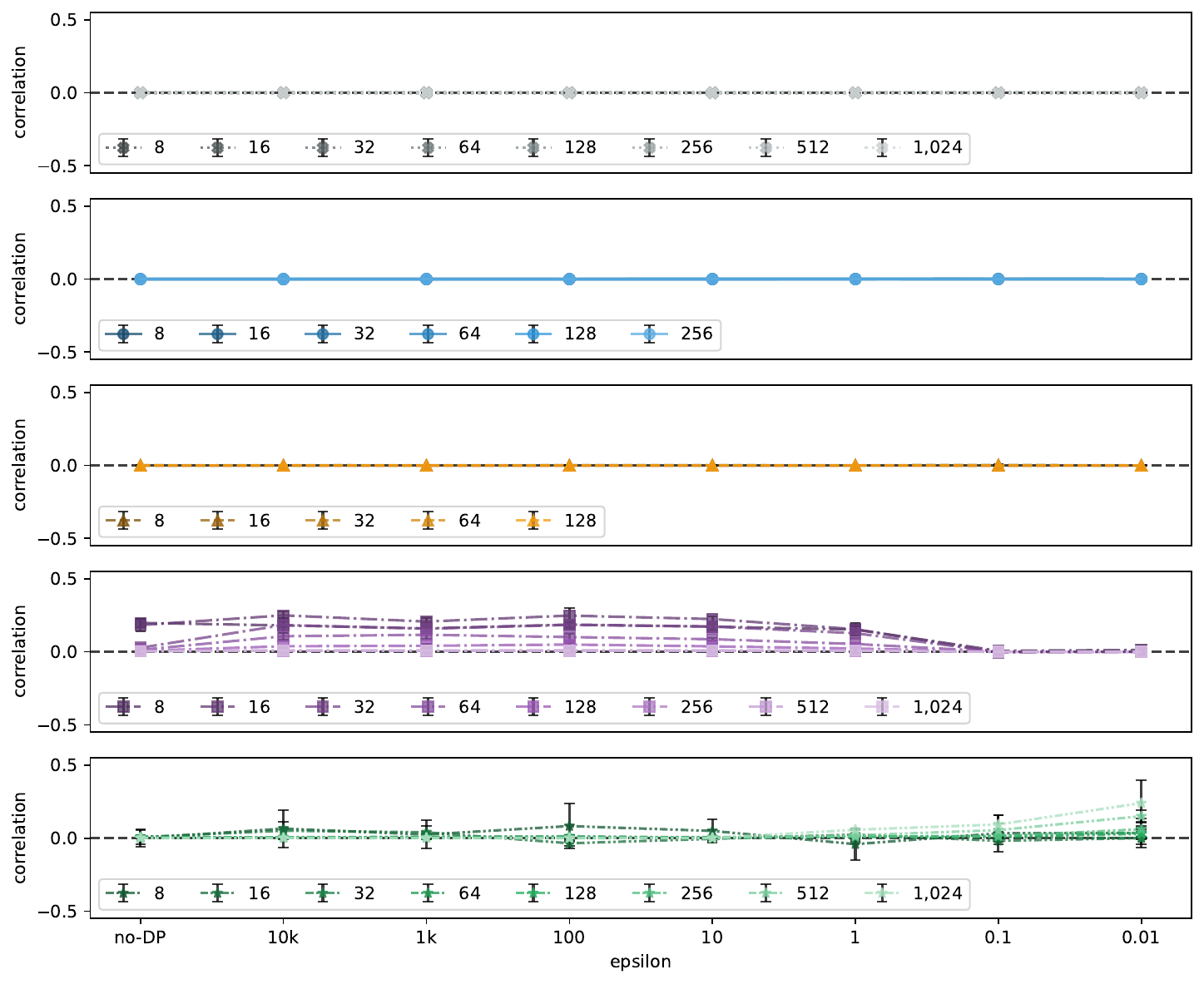}
		\caption{Varying $d$ and $n=16k$}
	\end{subfigure}
	\caption{T1: Other (apart from diagonal) pairwise correlation for different $\epsilon$ levels, on \emph{Eye~Gauss}, varying $n$ and $d$.}
	\label{fig:eye_other_corr}
\end{figure*}

\begin{figure*}[t!]
	\centering
	\begin{subfigure}{0.51\linewidth}
		 \includegraphics[width=0.99\textwidth]{plots2/gaussians/legend_real.pdf}
	\end{subfigure}
	\centering
	\begin{subfigure}{0.495\linewidth}
		 \includegraphics[width=0.99\textwidth]{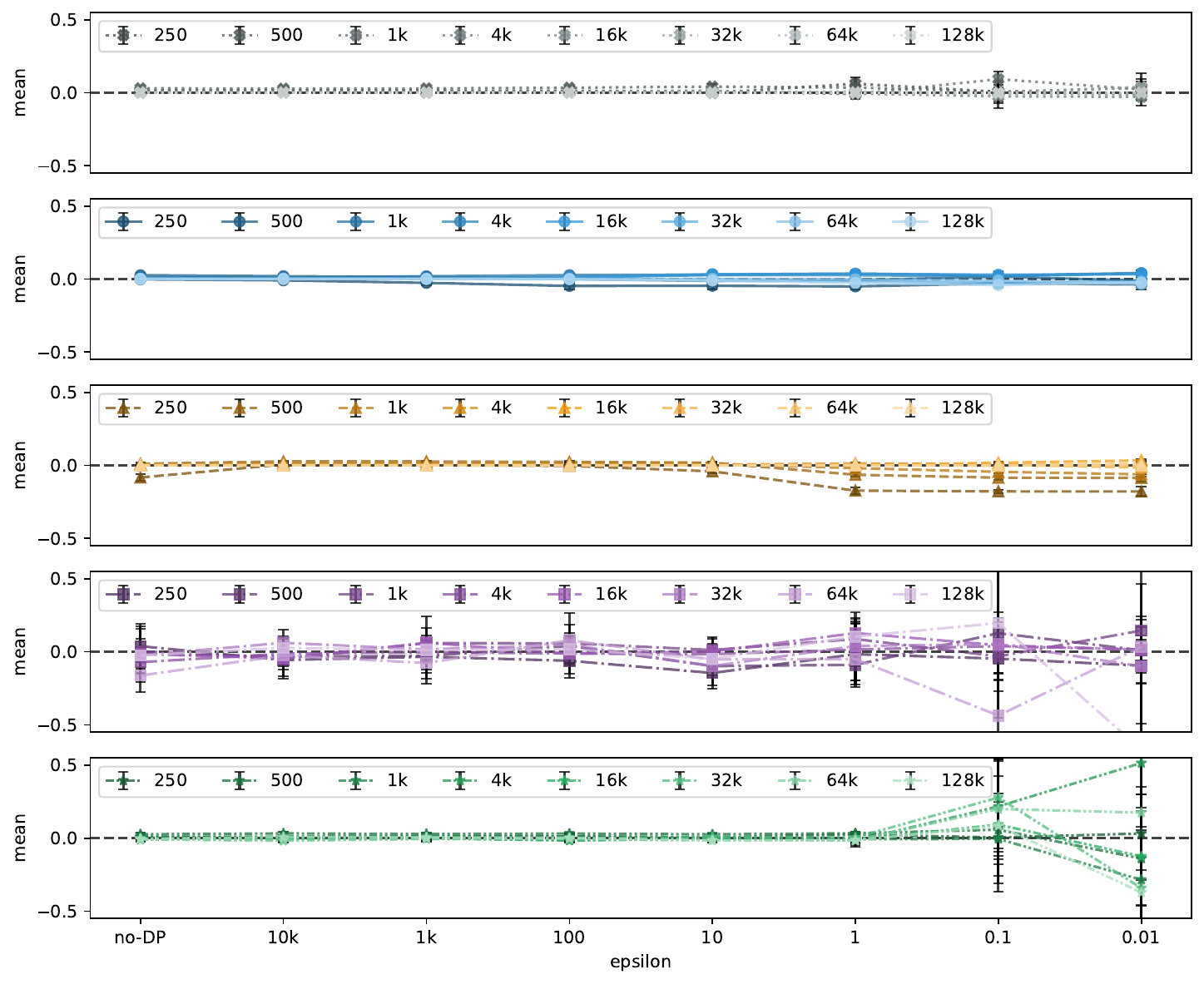}
		\caption{Varying $n$ and $d=32$}
	\end{subfigure}
	\begin{subfigure}{0.495\linewidth}
		\includegraphics[width=0.99\textwidth]{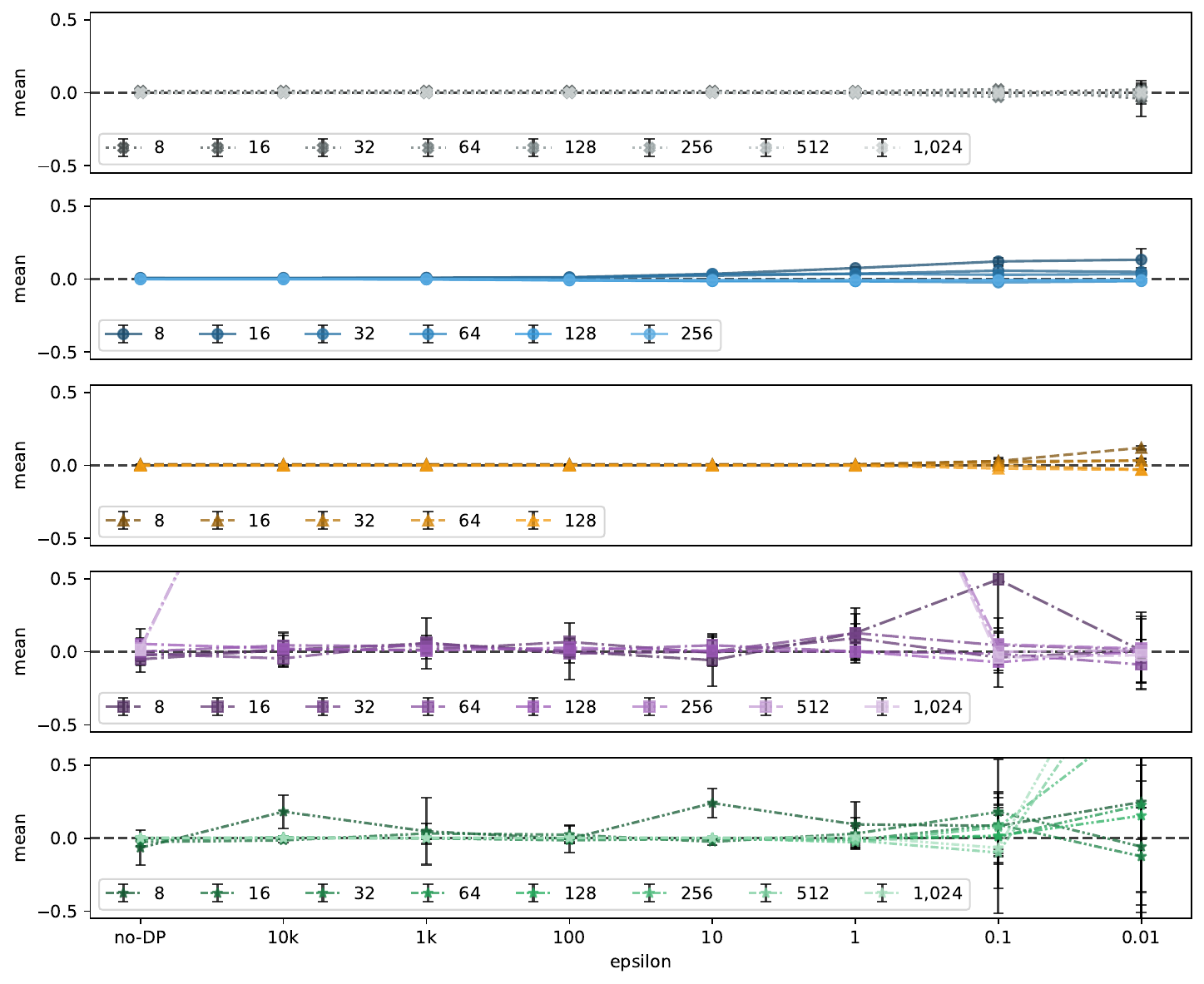}
		\caption{Varying $d$ and $n=16k$}
	\end{subfigure}
	\caption{T1: Marginal mean for different $\epsilon$ levels, on \emph{Corr~Gauss}, varying $n$ and $d$.}
	\label{fig:corr_mean}
\end{figure*}

\begin{figure*}[t!]
	\centering
	\begin{subfigure}{0.51\linewidth}
		 \includegraphics[width=0.99\textwidth]{plots2/gaussians/legend_real.pdf}
	\end{subfigure}
	\centering
	\begin{subfigure}{0.495\linewidth}
		 \includegraphics[width=0.99\textwidth]{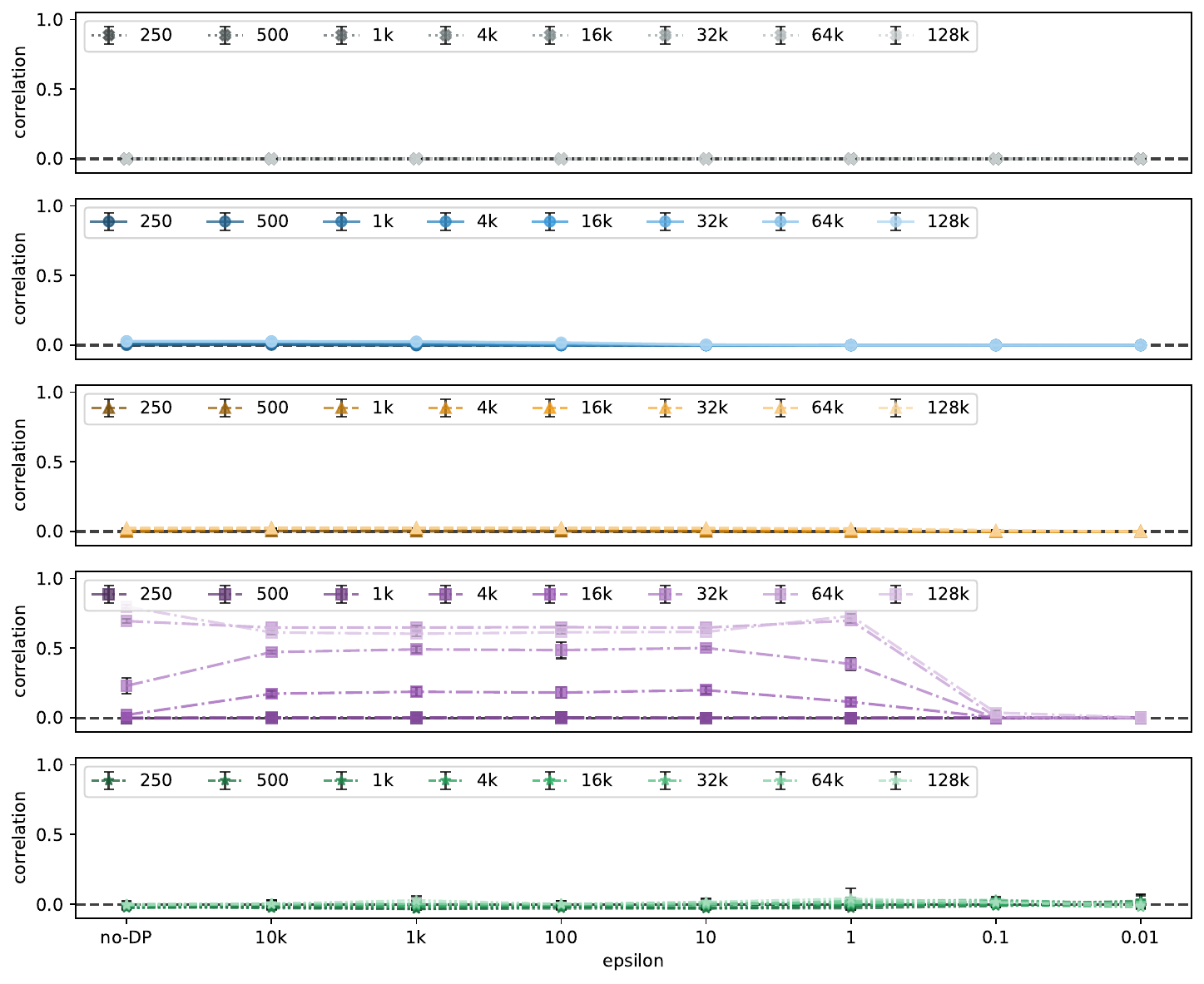}
		\caption{Varying $n$ and $d=32$}
	\end{subfigure}
	\begin{subfigure}{0.495\linewidth}
		\includegraphics[width=0.99\textwidth]{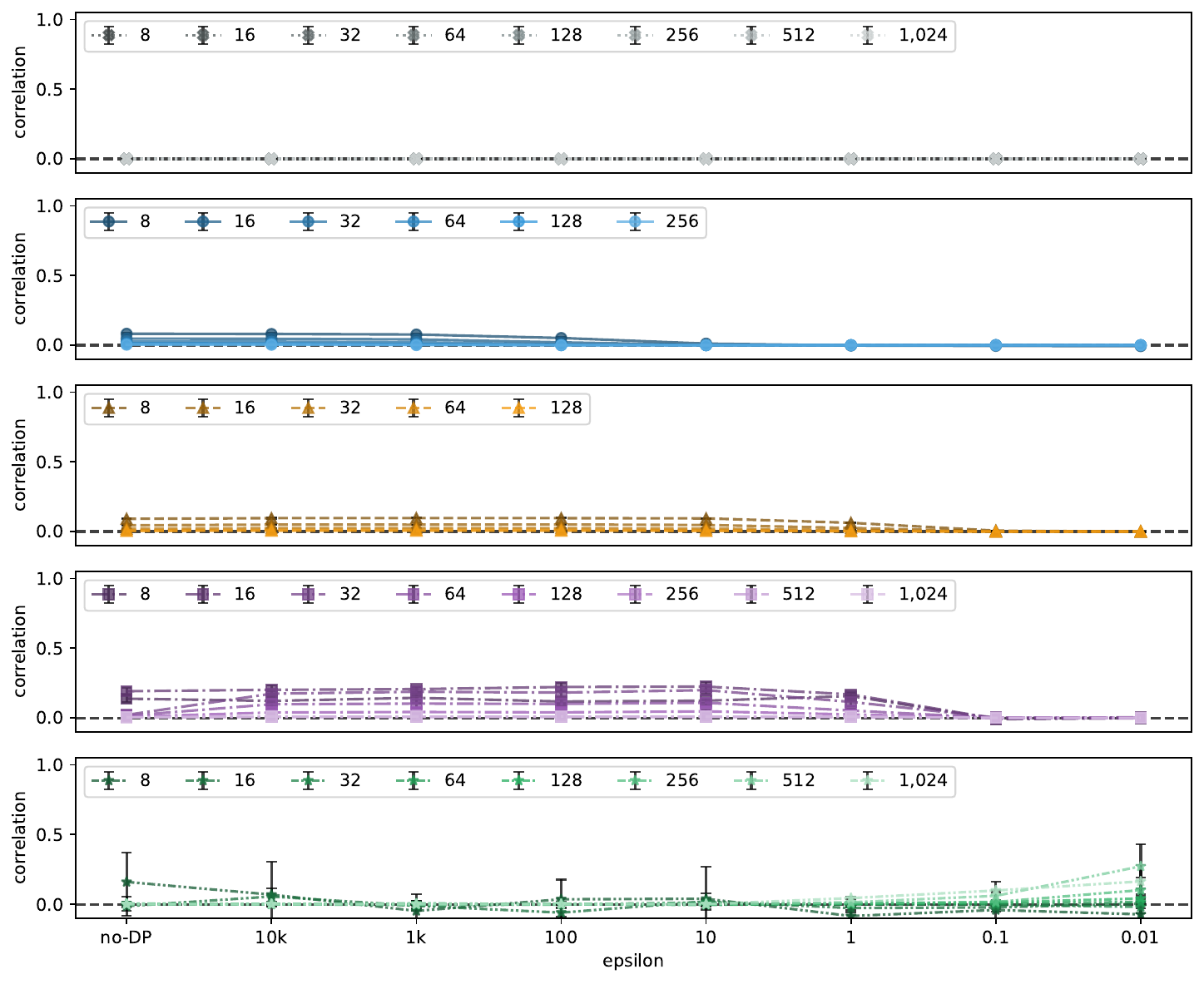}
		\caption{Varying $d$ and $n=16k$}
	\end{subfigure}
	\caption{T1: Other (apart from diagonal and off-diagonal) pairwise correlation for different $\epsilon$ levels, on \emph{Corr~Gauss}, varying $n$ and $d$.}
	\label{fig:corr_other_corr}
\end{figure*}

We visualize the average statistics for \emph{Eye~Gauss} and \emph{Corr~Gauss} with varying dimensions; more precisely, mean and correlation for the former in Figure~\ref{fig:eye_mean},~\ref{fig:eye_other_corr} and mean and correlation for the latter in Figure~\ref{fig:corr_mean},~\ref{fig:corr_other_corr}.
The results are discussed in detail in Section~\ref{subsec:statistics}.

\subsection{T2: Similarity}
\label{app:similarity}
\begin{figure*}[t!]
	\centering
	\begin{subfigure}{0.51\linewidth}
		 \includegraphics[width=0.99\textwidth]{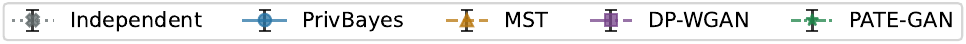}
	\end{subfigure}
	\centering
	\begin{subfigure}{0.495\linewidth}
		 \includegraphics[width=0.99\textwidth]{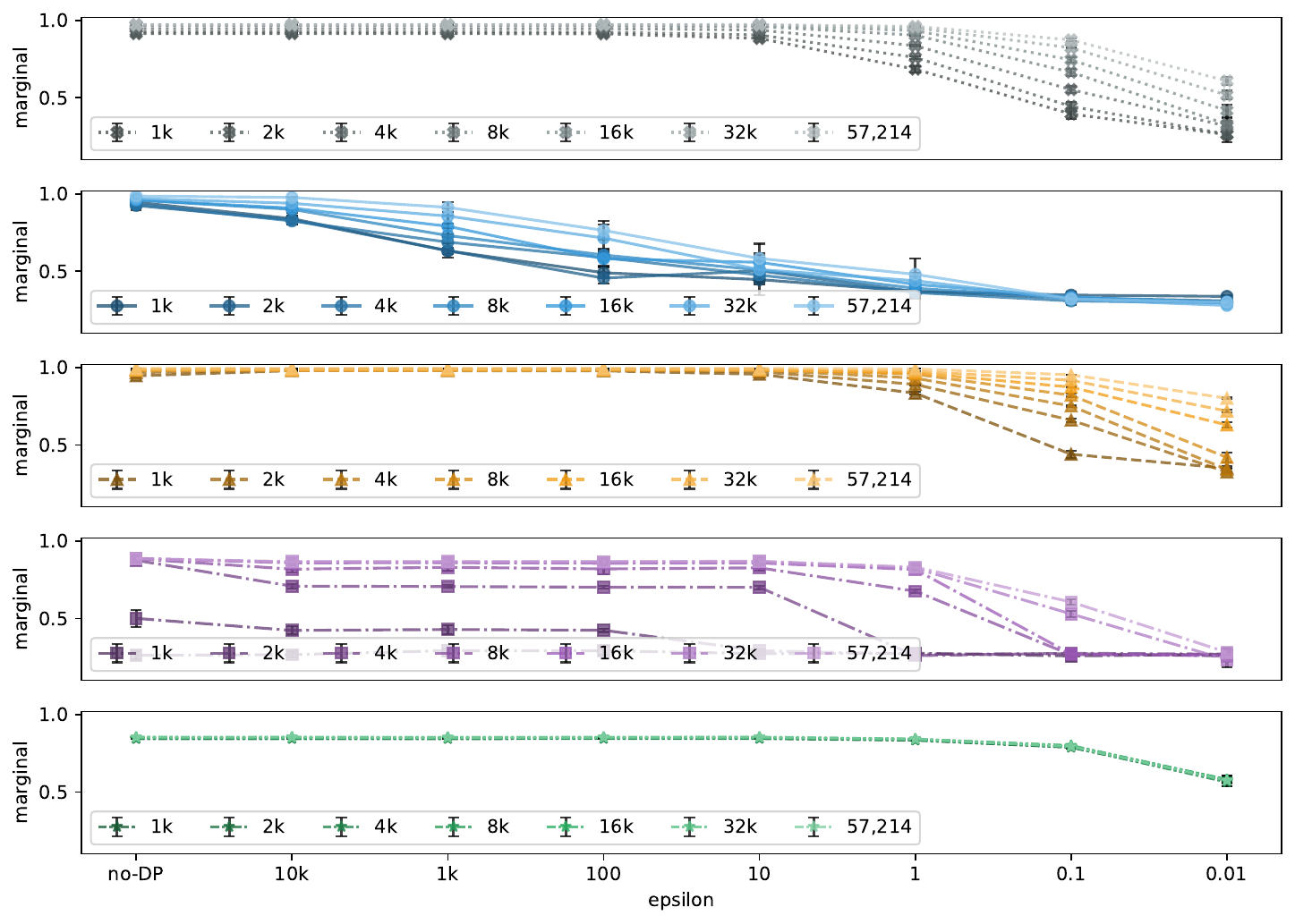}
		\caption{Marginal}
	\end{subfigure}
	\begin{subfigure}{0.495\linewidth}
		\includegraphics[width=0.99\textwidth]{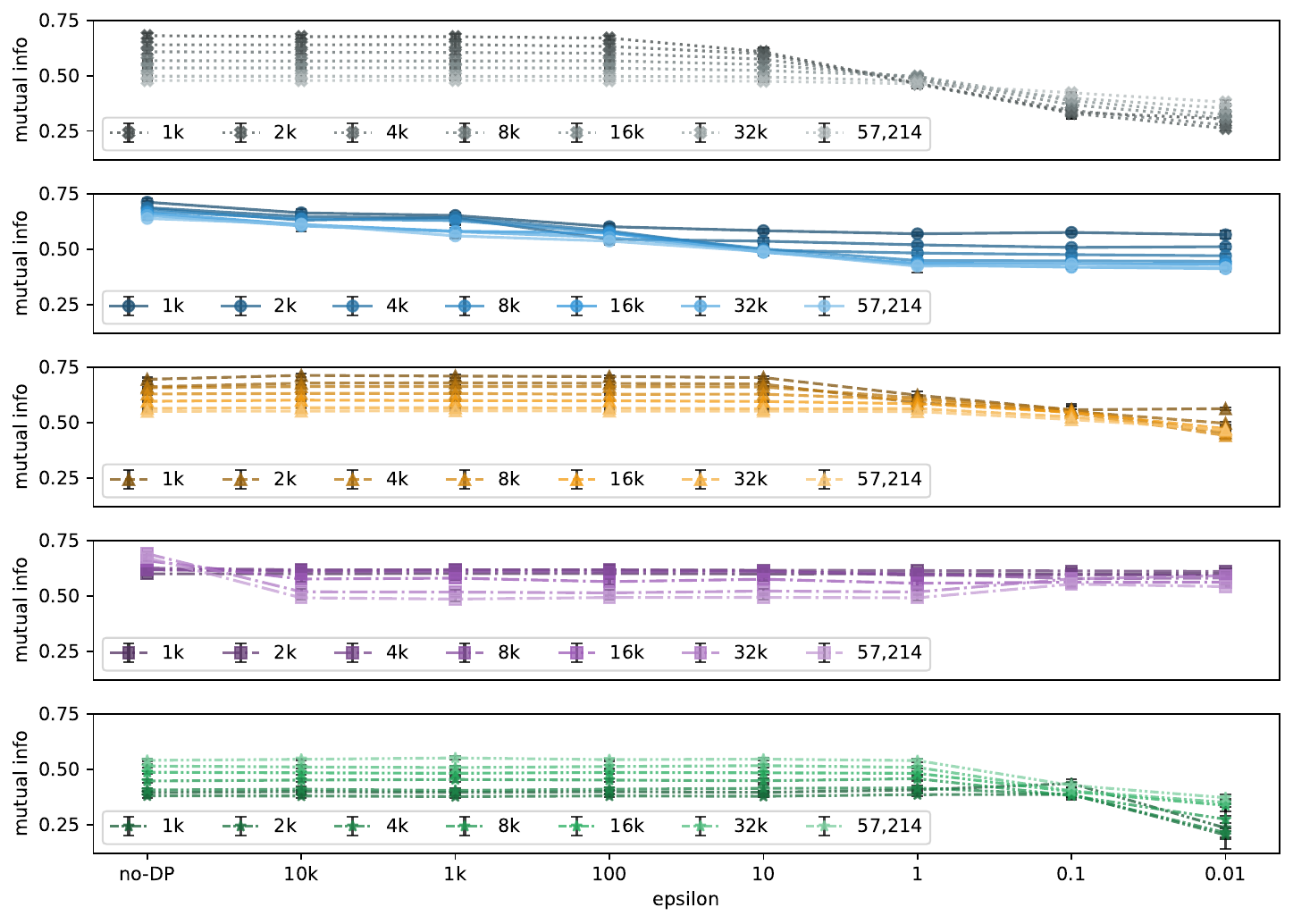}
		\caption{Mutual information}
		\label{fig:diabetes_mi}
	\end{subfigure}
	\caption{T2: Marginal and pairwise mutual information similarity for different $\epsilon$ levels, on \emph{Diabetes}, varying $n$.}
	\label{fig:diabetes_sim_mi}
\end{figure*}

\begin{figure*}[t!]
	\centering
	\begin{subfigure}{0.51\linewidth}
		 \includegraphics[width=0.99\textwidth]{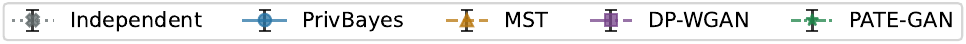}
	\end{subfigure}
	\centering
	\begin{subfigure}{0.495\linewidth}
		 \includegraphics[width=0.99\textwidth]{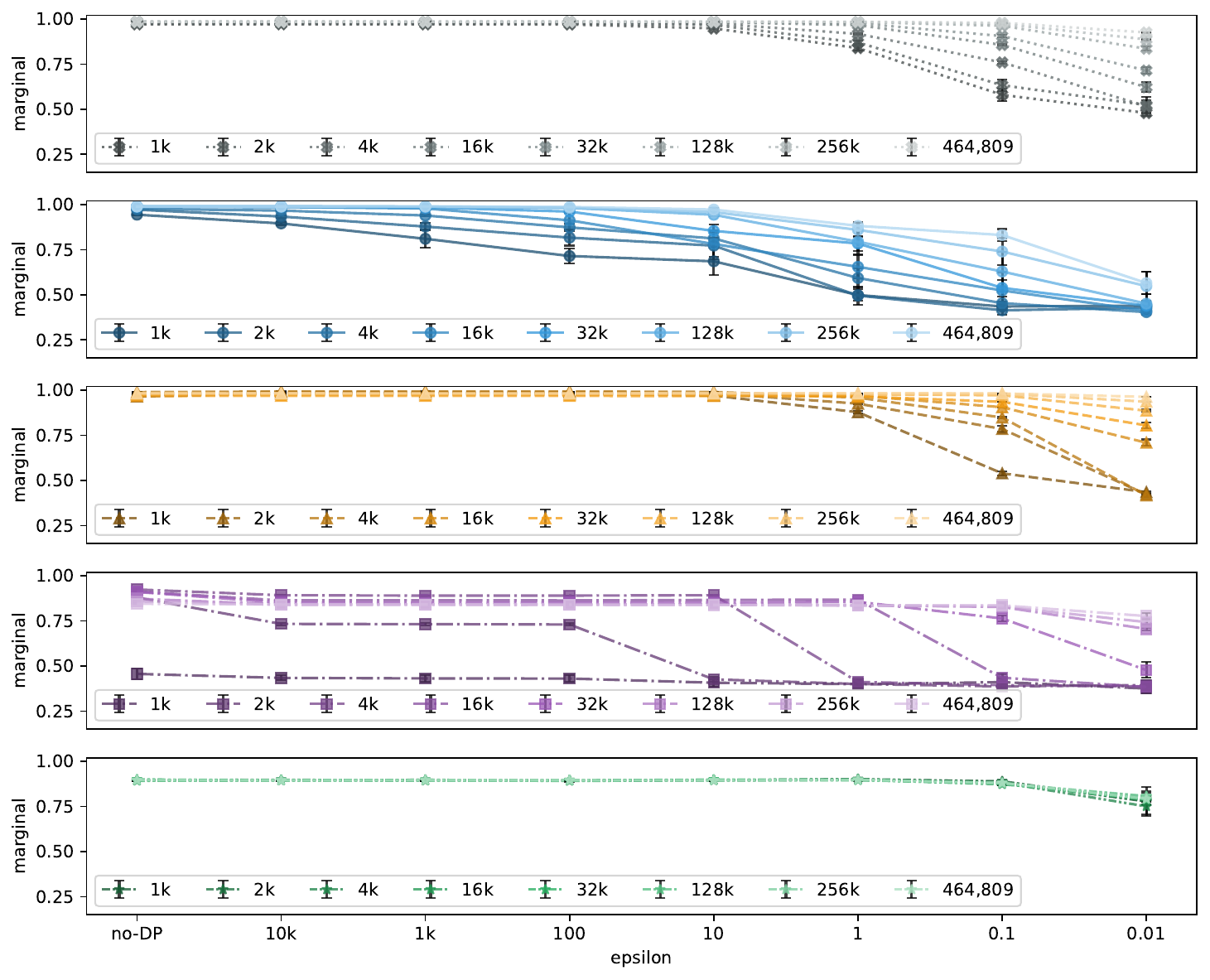}
		\caption{Marginal}
	\end{subfigure}
	\begin{subfigure}{0.495\linewidth}
		\includegraphics[width=0.99\textwidth]{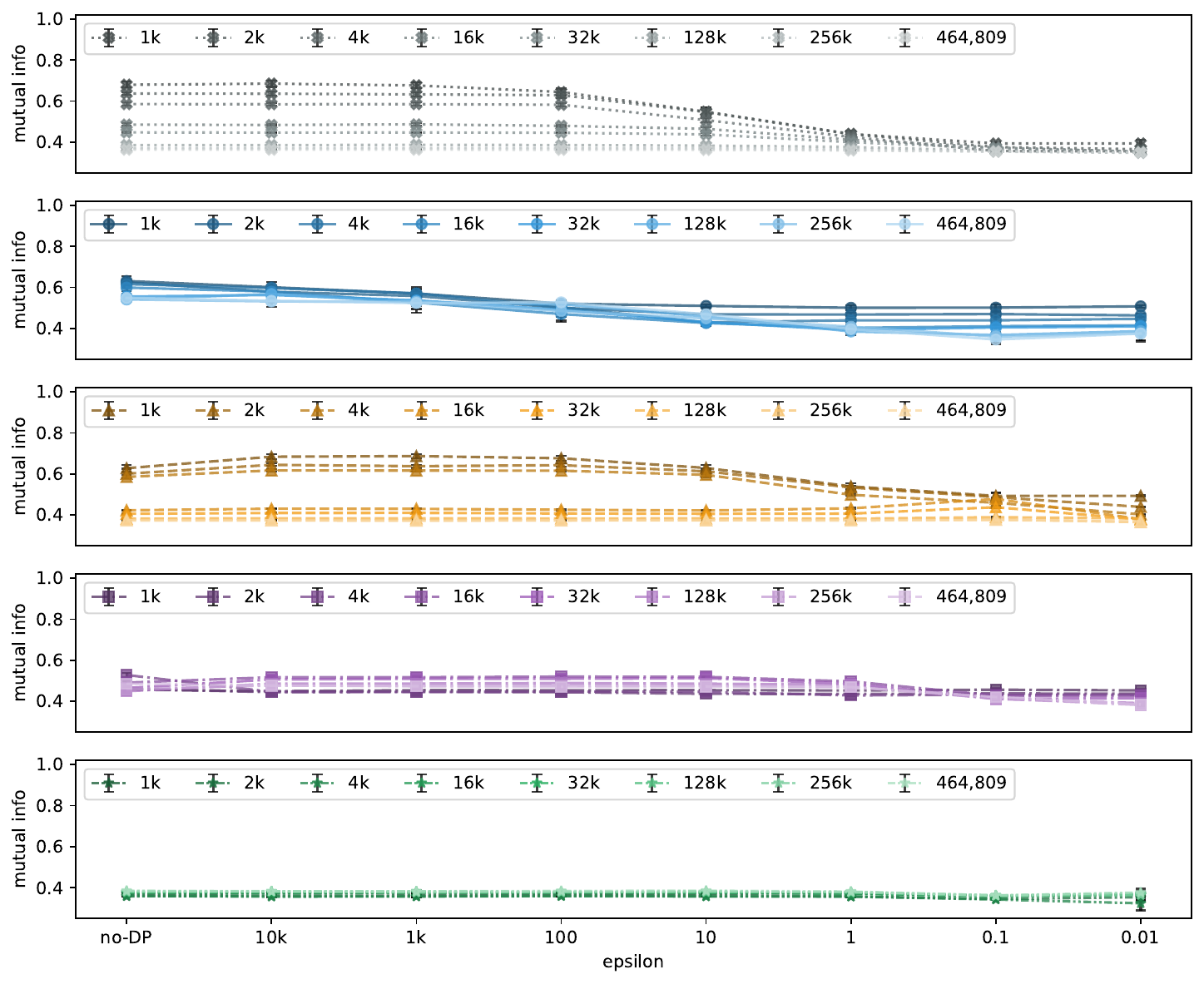}
		\caption{Mutual information}
		\label{fig:covertype_mi}
	\end{subfigure}
	\caption{T2: Marginal and pairwise mutual information similarity for different $\epsilon$ levels, on \emph{Covertype}, varying $n$.}
	\label{fig:covertype_sim_mi}
\end{figure*}

Marginal and pairwise mutual information similarly results for all models with varying $n$ on \emph{Diabetes} and \emph{Covertype} are displayed in Figure~\ref{fig:diabetes_sim_mi} and~\ref{fig:covertype_sim_mi}.
These experiments are discussed in Section~\ref{subsec:similarity}.

\subsection{T3: Clustering}
\label{app:pca}
\begin{figure*}[t!]
	\centering
	\begin{subfigure}{0.0763\linewidth}
		 \includegraphics[width=0.99\textwidth]{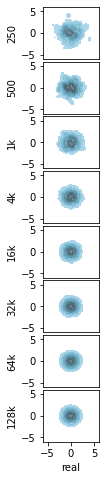}
		\caption{Real}
	\end{subfigure}
	\begin{subfigure}{0.175\linewidth}
		 \includegraphics[width=0.99\textwidth]{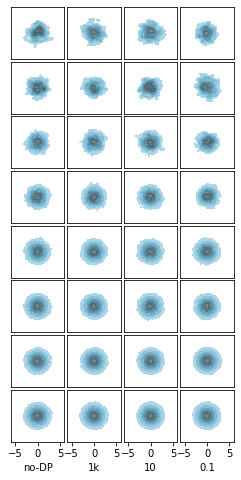}
		\caption{Independent}
	\end{subfigure}
	\begin{subfigure}{0.175\linewidth}
		 \includegraphics[width=0.99\textwidth]{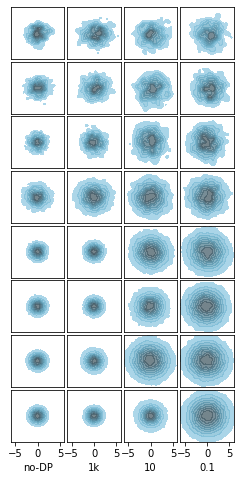}
		\caption{PrivBayes}
		\label{fig:eye_pca_rows_privbayes}
	\end{subfigure}
	\begin{subfigure}{0.175\linewidth}
		 \includegraphics[width=0.99\textwidth]{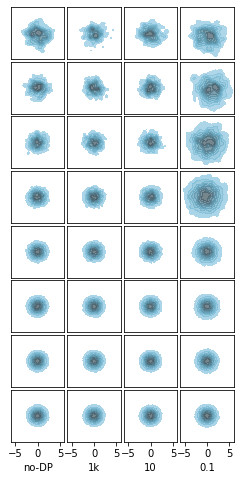}
		\caption{MST}
	\end{subfigure}
	\begin{subfigure}{0.175\linewidth}
		 \includegraphics[width=0.99\textwidth]{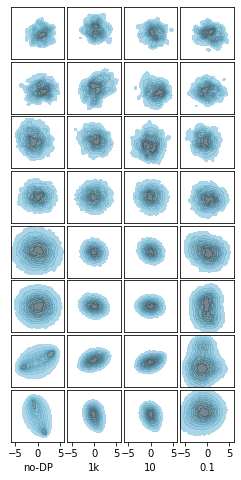}
		\caption{DP-WGAN}
	\end{subfigure}
	\begin{subfigure}{0.175\linewidth}
		 \includegraphics[width=0.99\textwidth]{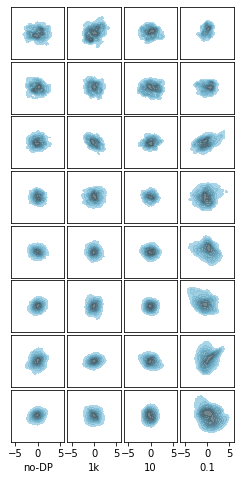}
		\caption{PATE-GAN}
	\end{subfigure}
	\caption{T3: KDE on the first 2 PCA principles for different $\epsilon$ levels, on \emph{Eye~Gauss}, varying $n$.}
	\label{fig:eye_pca_rows}
\reduce
\end{figure*}

\begin{figure*}[t!]
	\centering
	\begin{subfigure}{0.0763\linewidth}
		 \includegraphics[width=0.99\textwidth]{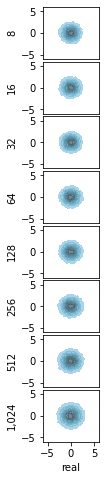}
		\caption{Real}
	\end{subfigure}
	\begin{subfigure}{0.175\linewidth}
		 \includegraphics[width=0.99\textwidth]{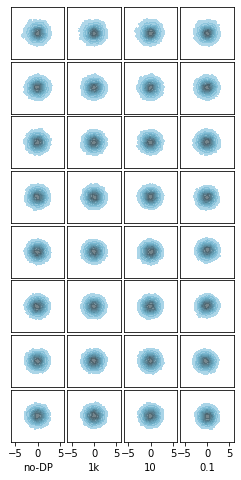}
		\caption{Independent}
	\end{subfigure}
	\begin{subfigure}{0.175\linewidth}
		 \includegraphics[width=0.99\textwidth]{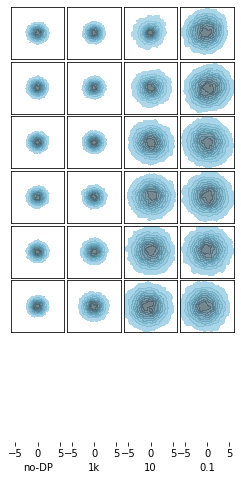}
		\caption{PrivBayes}
	\end{subfigure}
	\begin{subfigure}{0.175\linewidth}
		 \includegraphics[width=0.99\textwidth]{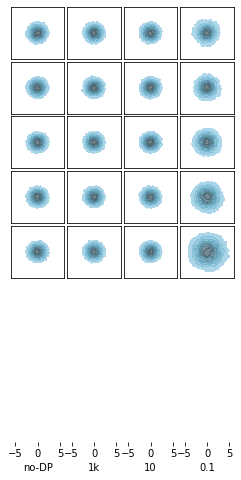}
		\caption{MST}
	\end{subfigure}
	\begin{subfigure}{0.175\linewidth}
		 \includegraphics[width=0.99\textwidth]{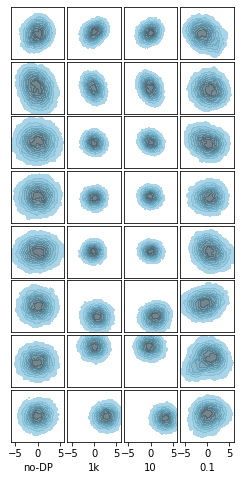}
		\caption{DP-WGAN}
	\end{subfigure}
	\begin{subfigure}{0.175\linewidth}
		 \includegraphics[width=0.99\textwidth]{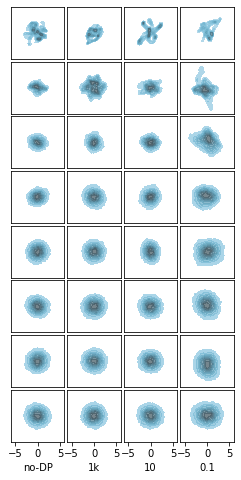}
		\caption{PATE-GAN}
	\end{subfigure}
	\caption{T3: KDE on the first 2 PCA principles for different $\epsilon$ levels, on \emph{Eye~Gauss}, varying $d$.}
	\label{fig:eye_pca_cols}
\reduce
\end{figure*}

\begin{figure*}[t!]
	\centering
	\begin{subfigure}{0.0763\linewidth}
		 \includegraphics[width=0.99\textwidth]{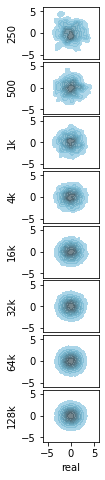}
		\caption{Real}
	\end{subfigure}
	\begin{subfigure}{0.175\linewidth}
		 \includegraphics[width=0.99\textwidth]{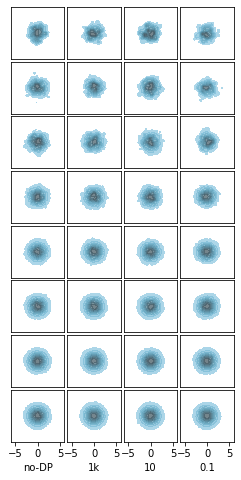}
		\caption{Independent}
	\end{subfigure}
	\begin{subfigure}{0.175\linewidth}
		 \includegraphics[width=0.99\textwidth]{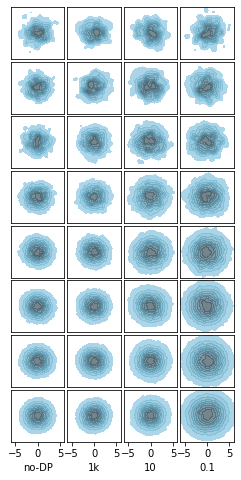}
		\caption{PrivBayes}
	\end{subfigure}
	\begin{subfigure}{0.175\linewidth}
		 \includegraphics[width=0.99\textwidth]{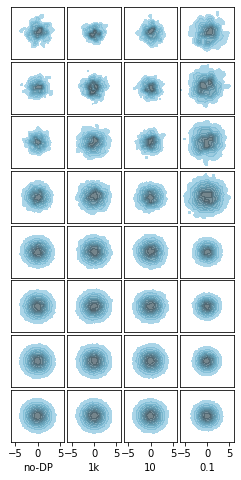}
		\caption{MST}
	\end{subfigure}
	\begin{subfigure}{0.175\linewidth}
		 \includegraphics[width=0.99\textwidth]{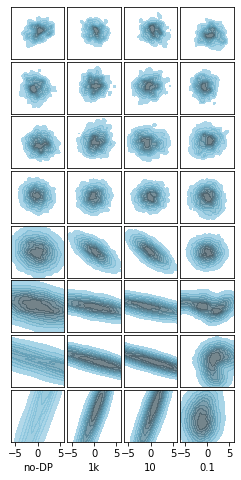}
		\caption{DP-WGAN}
	\end{subfigure}
	\begin{subfigure}{0.175\linewidth}
		 \includegraphics[width=0.99\textwidth]{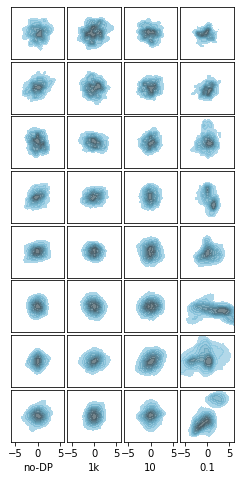}
		\caption{PATE-GAN}
	\end{subfigure}
	\caption{T3: KDE on the first 2 PCA principles for different $\epsilon$ levels, on \emph{Corr~Gauss}, varying $n$.}
	\label{fig:corr_pca_rows}
\reduce
\end{figure*}

\begin{figure*}[t!]
	\centering
	\begin{subfigure}{0.0763\linewidth}
		 \includegraphics[width=0.99\textwidth]{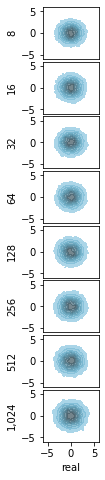}
		\caption{Real}
	\end{subfigure}
	\begin{subfigure}{0.175\linewidth}
		 \includegraphics[width=0.99\textwidth]{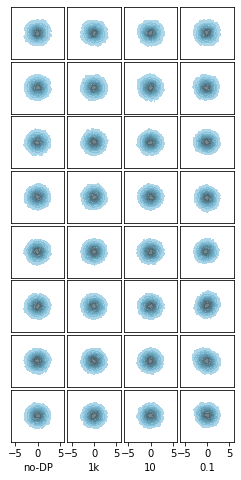}
		\caption{Independent}
	\end{subfigure}
	\begin{subfigure}{0.175\linewidth}
		 \includegraphics[width=0.99\textwidth]{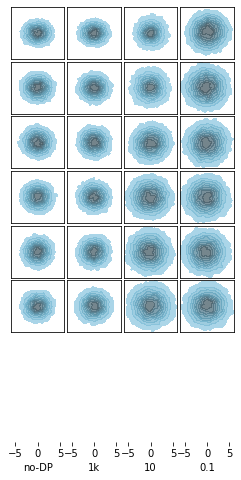}
		\caption{PrivBayes}
	\end{subfigure}
	\begin{subfigure}{0.175\linewidth}
		 \includegraphics[width=0.99\textwidth]{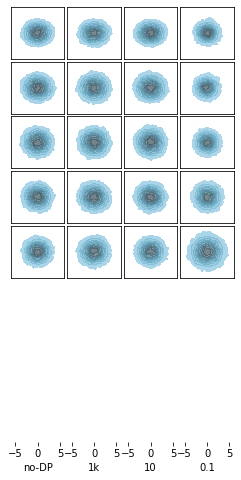}
		\caption{MST}
	\end{subfigure}
	\begin{subfigure}{0.175\linewidth}
		 \includegraphics[width=0.99\textwidth]{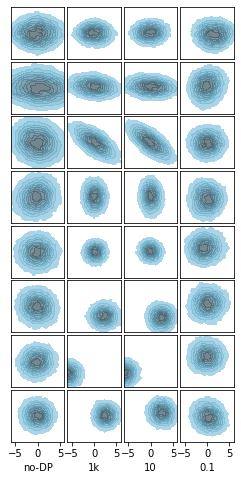}
		\caption{DP-WGAN}
	\end{subfigure}
	\begin{subfigure}{0.175\linewidth}
		 \includegraphics[width=0.99\textwidth]{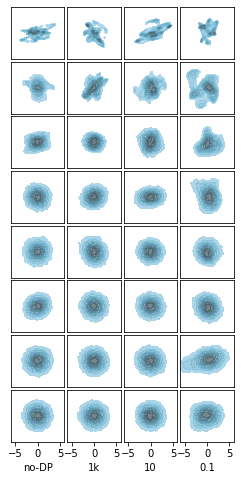}
		\caption{PATE-GAN}
	\end{subfigure}
	\caption{T3: KDE on the first 2 PCA principles for different $\epsilon$ levels, on \emph{Corr~Gauss}, varying $d$.}
	\label{fig:corr_pca_cols}
\reduce
\end{figure*}

\begin{figure*}[t!]
	\centering
	\begin{subfigure}{0.0763\linewidth}
		 \includegraphics[width=0.99\textwidth]{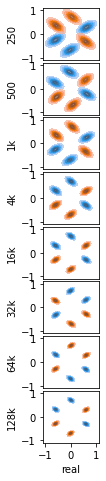}
		\caption{Real}
	\end{subfigure}
	\begin{subfigure}{0.175\linewidth}
		 \includegraphics[width=0.99\textwidth]{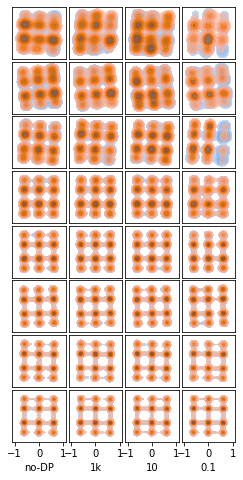}
		\caption{Independent}
	\end{subfigure}
	\begin{subfigure}{0.175\linewidth}
		 \includegraphics[width=0.99\textwidth]{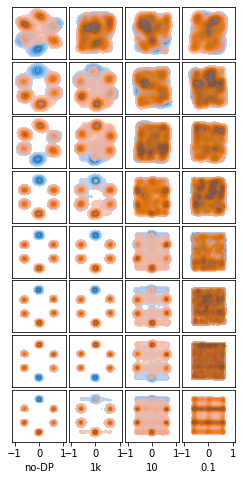}
		\caption{PrivBayes}
	\end{subfigure}
	\begin{subfigure}{0.175\linewidth}
		 \includegraphics[width=0.99\textwidth]{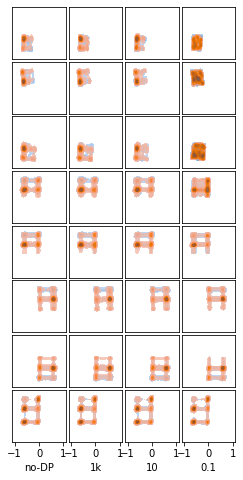}
		\caption{MST}
	\end{subfigure}
	\begin{subfigure}{0.175\linewidth}
		 \includegraphics[width=0.99\textwidth]{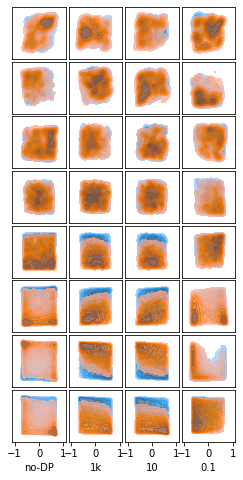}
		\caption{DP-WGAN}
	\end{subfigure}
	\begin{subfigure}{0.175\linewidth}
		 \includegraphics[width=0.99\textwidth]{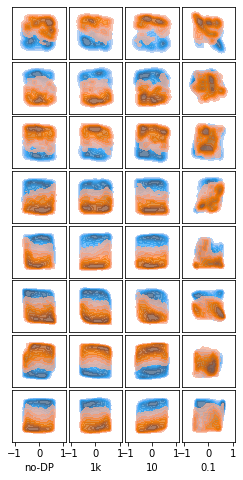}
		\caption{PATE-GAN}
	\end{subfigure}
	\caption{T3: KDE on the first 2 PCA principles for different $\epsilon$ levels, on \emph{Mix~Gauss~Sup}, varying $n$.}
	\label{fig:mix_target_pca_rows}
\reduce
\end{figure*}

\begin{figure*}[t!]
	\centering
	\begin{subfigure}{0.0763\linewidth}
		 \includegraphics[width=0.99\textwidth]{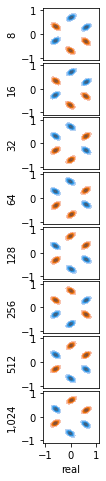}
		\caption{Real}
	\end{subfigure}
	\begin{subfigure}{0.175\linewidth}
		 \includegraphics[width=0.99\textwidth]{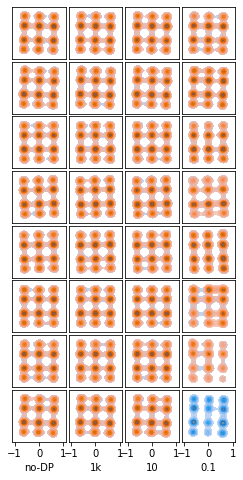}
		\caption{Independent}
	\end{subfigure}
	\begin{subfigure}{0.175\linewidth}
		 \includegraphics[width=0.99\textwidth]{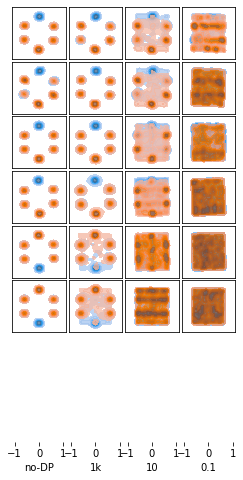}
		\caption{PrivBayes}
	\end{subfigure}
	\begin{subfigure}{0.175\linewidth}
		 \includegraphics[width=0.99\textwidth]{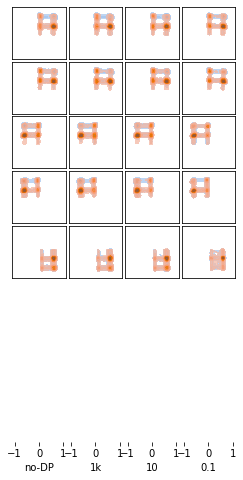}
		\caption{MST}
	\end{subfigure}
	\begin{subfigure}{0.175\linewidth}
		 \includegraphics[width=0.99\textwidth]{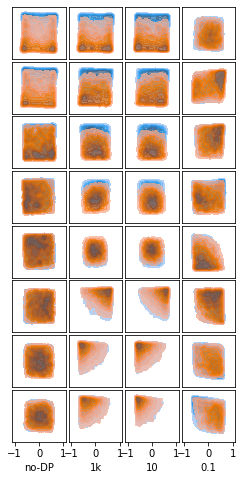}
		\caption{DP-WGAN}
	\end{subfigure}
	\begin{subfigure}{0.175\linewidth}
		 \includegraphics[width=0.99\textwidth]{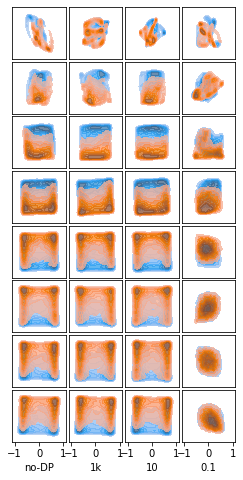}
		\caption{PATE-GAN}
	\end{subfigure}
	\caption{T3: KDE on the first 2 PCA principles for different $\epsilon$ levels, on \emph{Mix~Gauss~Sup}, varying $d$.}
	\label{fig:mix_target_pca_cols}
\reduce
\end{figure*}

\begin{figure*}[t!]
	\centering
	\begin{subfigure}{0.0813\linewidth}
		 \includegraphics[width=0.99\textwidth]{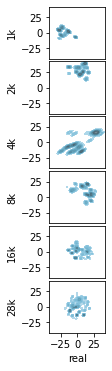}
		\caption{Real}
	\end{subfigure}
	\begin{subfigure}{0.175\linewidth}
		 \includegraphics[width=0.99\textwidth]{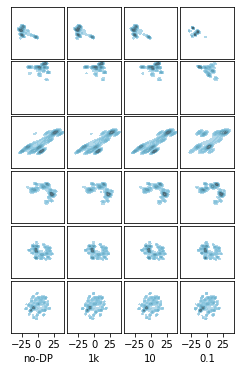}
		\caption{Independent}
	\end{subfigure}
	\begin{subfigure}{0.175\linewidth}
		 \includegraphics[width=0.99\textwidth]{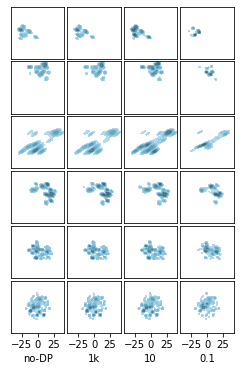}
		\caption{PrivBayes}
	\end{subfigure}
	\begin{subfigure}{0.175\linewidth}
		 \includegraphics[width=0.99\textwidth]{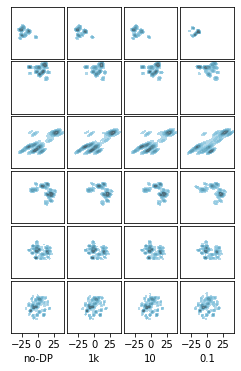}
		\caption{MST}
	\end{subfigure}
	\begin{subfigure}{0.175\linewidth}
		 \includegraphics[width=0.99\textwidth]{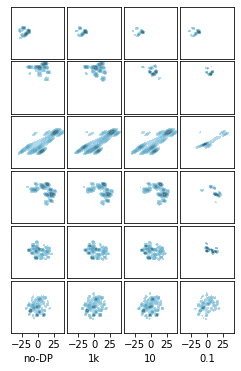}
		\caption{DP-WGAN}
	\end{subfigure}
	\begin{subfigure}{0.175\linewidth}
		 \includegraphics[width=0.99\textwidth]{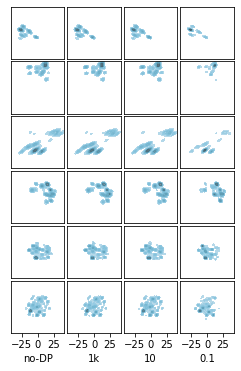}
		\caption{PATE-GAN}
	\end{subfigure}
	\caption{T3: KDE on the first 2 UMAP projections for different $\epsilon$ levels, on \emph{Plants}, varying $n$.}
	\label{fig:plants_umap}
\reduce
\end{figure*}

\begin{figure*}[t!]
	\centering
	\begin{subfigure}{0.51\linewidth}
		 \includegraphics[width=0.99\textwidth]{plots2/gaussians/legend_real.pdf}
	\end{subfigure}
	\centering
	\begin{subfigure}{0.495\linewidth}
		 \includegraphics[width=0.99\textwidth]{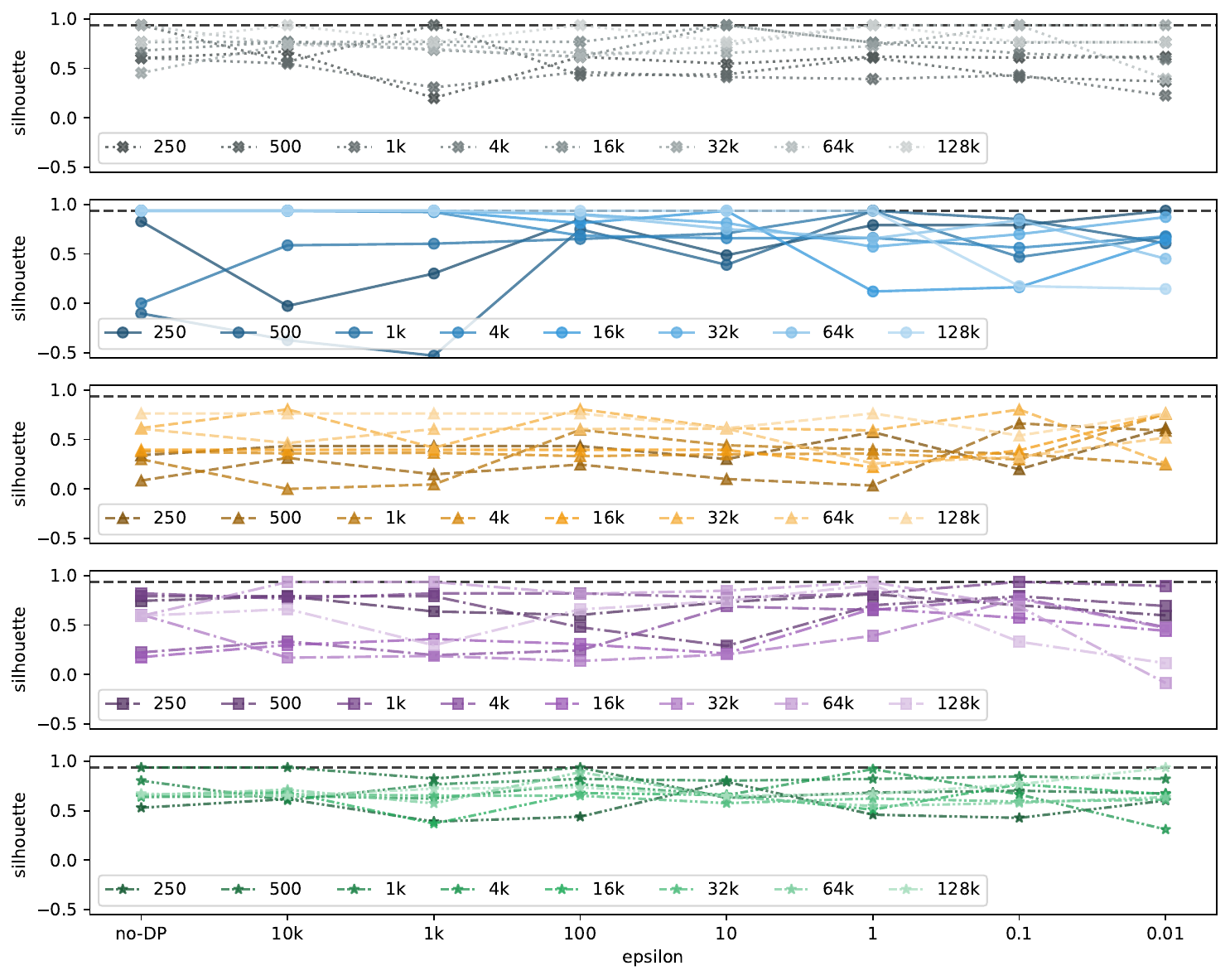}
		\caption{Varying $n$ and $d=32$}
	\end{subfigure}
	\begin{subfigure}{0.495\linewidth}
		\includegraphics[width=0.99\textwidth]{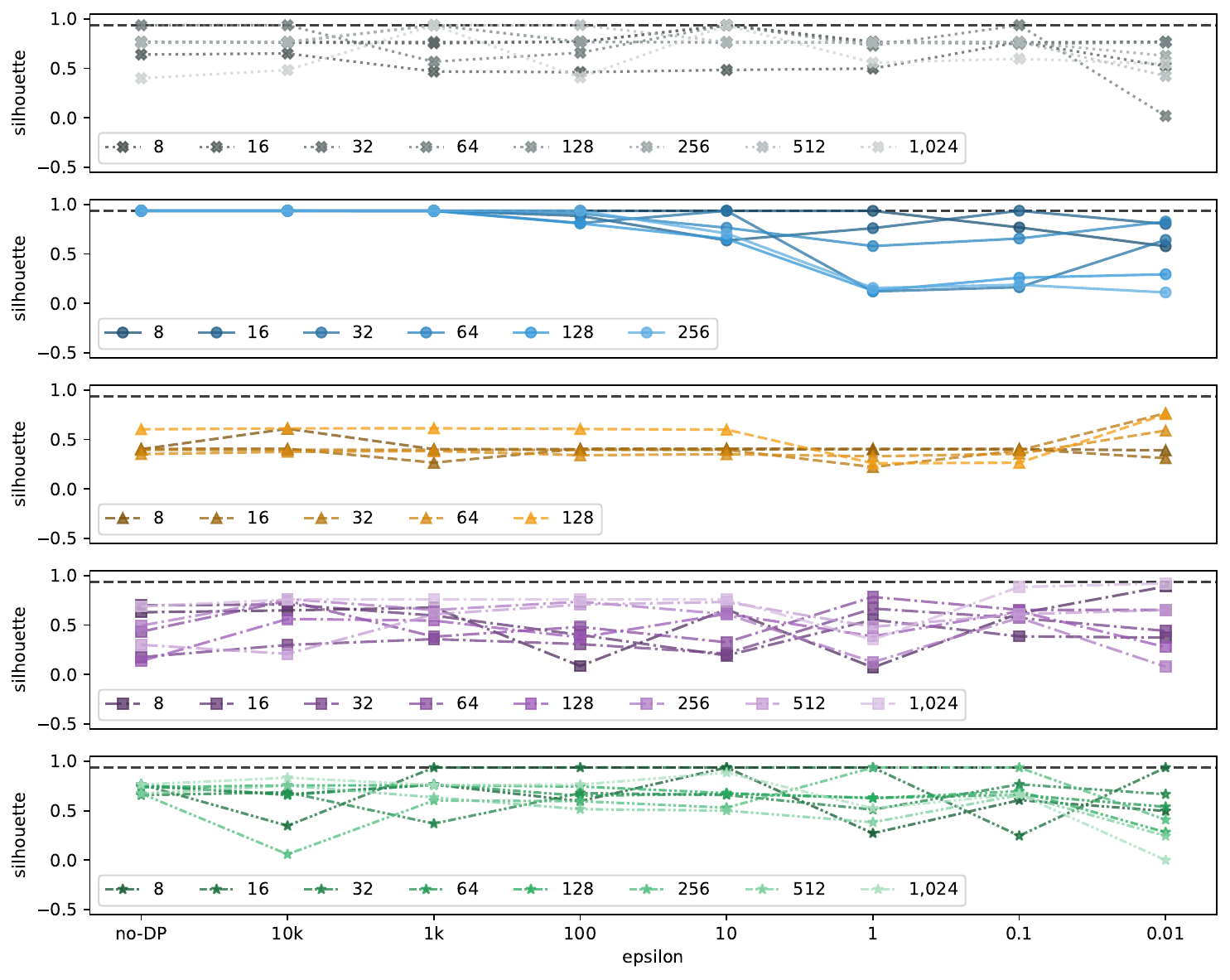}
		\caption{Varying $n$ and $d=32$}
	\end{subfigure}
	\caption{T3: Silhouette score for different $\epsilon$ levels, on \emph{Mix~Gauss~Unsup}, varying $n$ and $d$.}
	\label{fig:mix_sil}
\end{figure*}

\begin{figure*}[t!]
	\centering
	\begin{subfigure}{0.51\linewidth}
		 \includegraphics[width=0.99\textwidth]{plots2/gaussians/legend_real.pdf}
	\end{subfigure}
	\centering
	\begin{subfigure}{0.495\linewidth}
		 \includegraphics[width=0.99\textwidth]{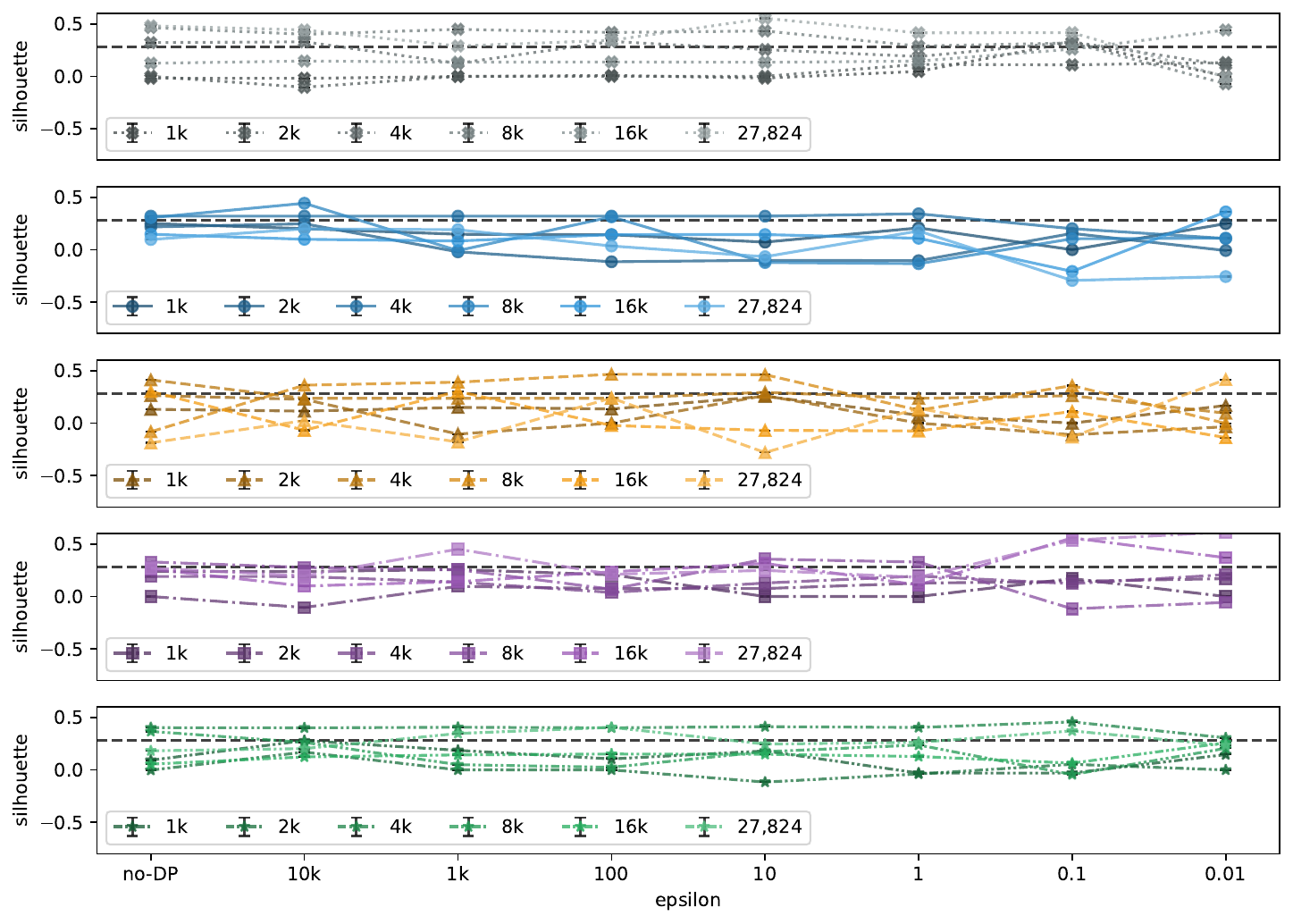}
	\end{subfigure}
	\caption{T3: Silhouette score for different $\epsilon$ levels, on \emph{Plants}, varying $n$.}
	\label{fig:plants_sil}
\end{figure*}

The KDE on the first 2 PCA components for three of the \emph{Gauss} datasets with varying dimensions are plotted in Figure~\ref{fig:eye_pca_rows},~\ref{fig:eye_pca_cols},~\ref{fig:corr_pca_rows},~\ref{fig:corr_pca_cols},~\ref{fig:mix_target_pca_rows}, and~\ref{fig:mix_target_pca_cols} while Figure~\ref{fig:plants_umap} displays the UMAP visualization for \emph{Plants}.
The silhouette scores of \emph{Corr~Gauss} and \emph{Plants} are shown in Figure~\ref{fig:mix_sil} and~\ref{fig:plants_sil}, respectively.
We analyze the results in Section~\ref{subsec:pca}.

\subsection{T4: Classification}
\label{app:classification}
\begin{figure*}[t!]
	\centering
	\begin{subfigure}{0.51\linewidth}
		 \includegraphics[width=0.99\textwidth]{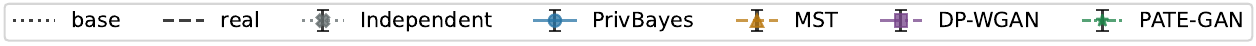}
	\end{subfigure}
	\centering
	\begin{subfigure}{0.495\linewidth}
		 \includegraphics[width=0.99\textwidth]{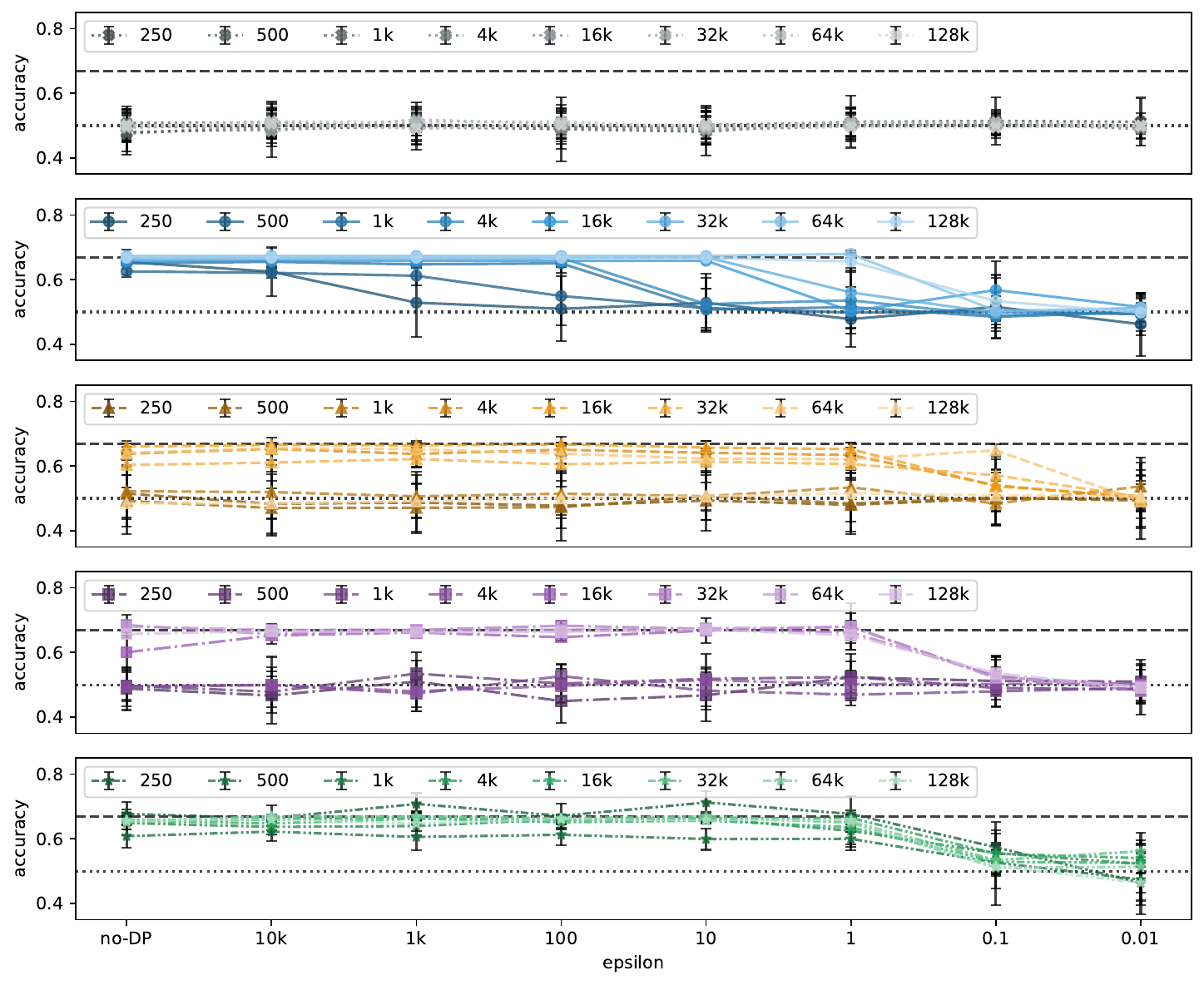}
		\caption{Varying $n$ and $d=32$}
	\end{subfigure}
	\begin{subfigure}{0.495\linewidth}
		\includegraphics[width=0.99\textwidth]{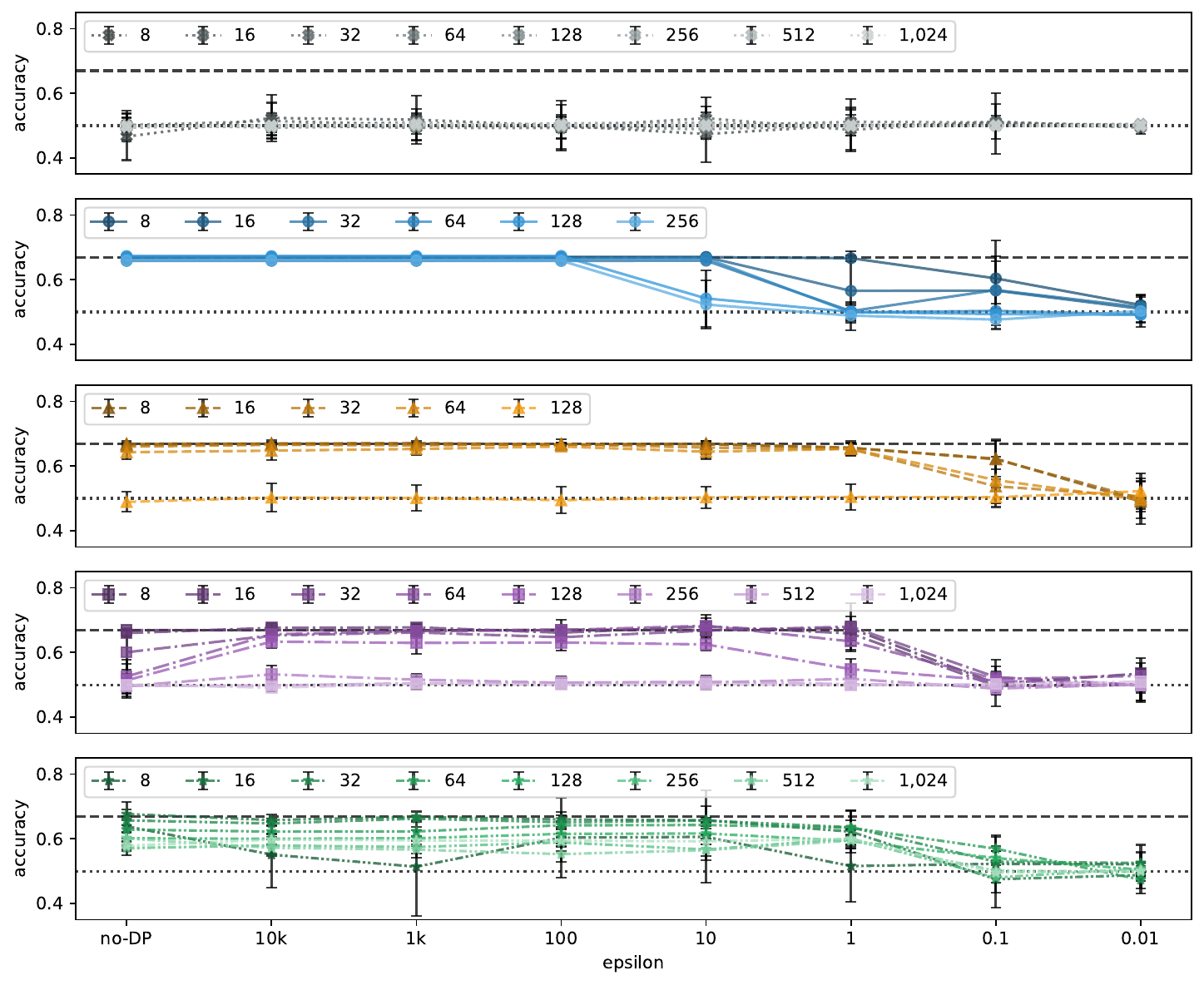}
		\caption{Varying $d$ and $n=16k$}
	\end{subfigure}
	\caption{T4: Accuracy for different $\epsilon$ levels, on \emph{Mix~Gauss~Sup}, varying $n$ and $d$.}
	\label{fig:mix_target_acc}
\end{figure*}

\begin{figure*}[t!]
	\centering
	\begin{subfigure}{0.51\linewidth}
		 \includegraphics[width=0.99\textwidth]{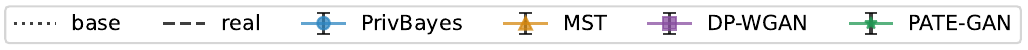}
	\end{subfigure}
	\centering
	\begin{subfigure}{0.495\linewidth}
		 \includegraphics[width=0.99\textwidth]{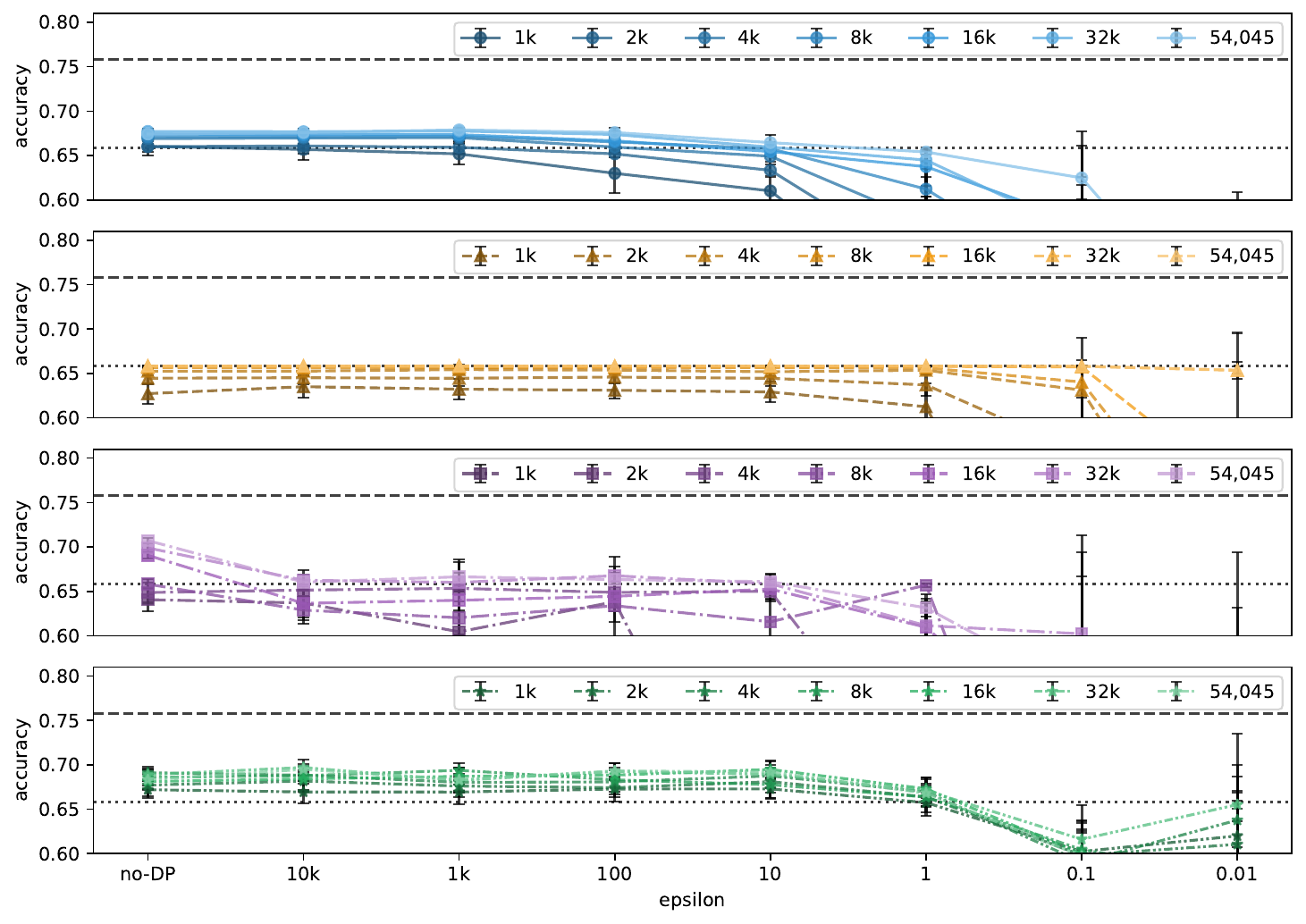}
		\caption{Accuracy}
	\end{subfigure}
	\begin{subfigure}{0.495\linewidth}
		\includegraphics[width=0.99\textwidth]{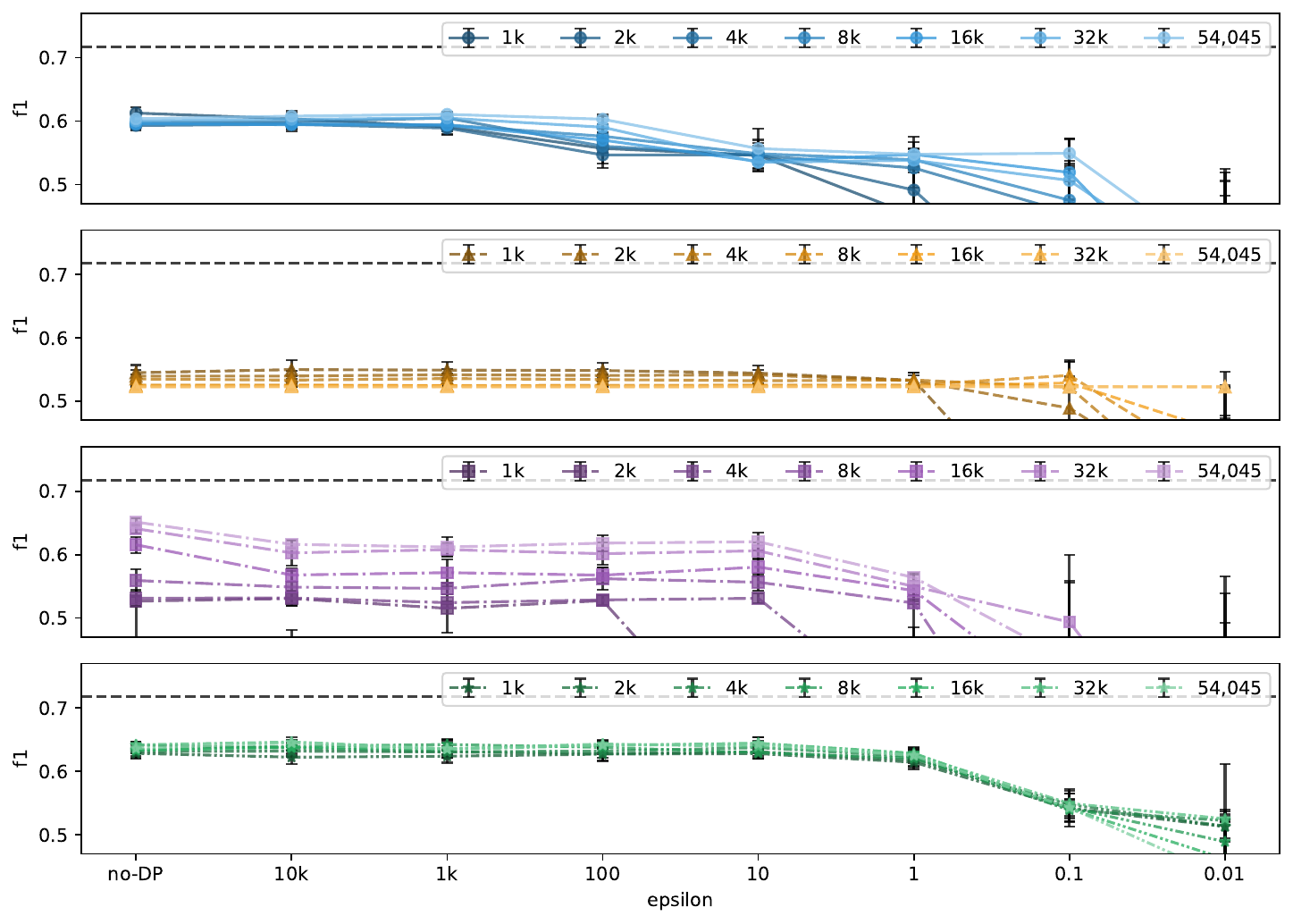}
		\caption{F1}
	\end{subfigure}
	\caption{T4: Accuracy and F1 for different $\epsilon$ levels, on \emph{Connect~4}, varying $n$.}
	\label{fig:connect_4_acc_f1}
\end{figure*}

The accuracy for \emph{Mix~Gauss~Sup} with varying dimensions is plotted in Figure~\ref{fig:mix_target_acc} while the accuracy and F1 for \emph{Connect~4} with increasing $n$ are displayed in Figure~\ref{fig:connect_4_acc_f1}.
We discuss them in Section~\ref{subsec:classification}.

\end{document}